%% file: main.tex
\documentclass[letterpaper]{article} 
\usepackage{aaai2026}  
\usepackage{times}  
\usepackage{helvet}  
\usepackage{courier}  
\usepackage[hyphens]{url}  
\usepackage{graphicx} 
\usepackage{natbib}  
\usepackage{caption} 
\usepackage{algorithm}
\usepackage{algorithmic}

\usepackage{newfloat}
\usepackage{listings}
\DeclareCaptionStyle{ruled}{labelfont=normalfont,labelsep=colon,strut=off} 
\floatstyle{ruled}
\newfloat{listing}{tb}{lst}{}
\floatname{listing}{Listing}
\usepackage{multirow}
\usepackage{bm}
\usepackage{soul}
\usepackage{xcolor}
\usepackage{mathtools}
\usepackage{pifont}
\usepackage{subcaption}
\usepackage{amsmath}
\usepackage{cleveref}
\crefname{appendix}{Appendix}{Appendices}
\Crefname{appendix}{Appendix}{Appendices}
\usepackage{amsthm}
\usepackage{booktabs}
\usepackage{amsfonts}
\usepackage{dsfont}

\newtheorem{lemma}{Lemma} 

\definecolor{BrickRed}{rgb}{0.6,0,0}
\definecolor{RoyalBlue}{rgb}{0,0,0.8}
\definecolor{Tdgreen}{rgb}{0,0.4,0.7}
\definecolor{cadmiumgreen}{rgb}{0.0, 0.42, 0.24}

\newcommand{\ours}{PosteL}

\newcommand{\tabnum}[2]{{#1{\footnotesize±}{\scriptsize#2}}}
\newcommand{\besttabnum}[2]{{\textbf{#1}{\footnotesize\textbf{±}}{\scriptsize\textbf{#2}}}}

\newcommand{\set}[1]{\mathcal{#1}}
\renewcommand{\vec}[1]{\boldsymbol{#1}}
\newcommand{\oname}[1]{\operatorname{#1}}
\newcommand{\nei}[2]{\set{N}^{#1}_{#2}}

\newcommand{\dw}[1]{\textcolor{red}{#1}}

\title{Posterior Label Smoothing for Node Classification}
\author{
    Jaeseung Heo\textsuperscript{\rm 1}, MoonJeong Park\textsuperscript{\rm 1}, Dongwoo Kim\textsuperscript{\rm 1, \rm 2}\\
}
\affiliations{
    \textsuperscript{\rm 1}Graduate School of Artificial Intelligence, POSTECH, South Korea \\
    \textsuperscript{\rm 2}Department of Computer Science \& Engineering, POSTECH, South Korea \\
    jsheo12304@postech.ac.kr, mjeongp@postech.ac.kr, dongwookim@postech.ac.kr \\
    
}

\usepackage{bibentry}

\begin{document}

\maketitle
\input{tex/0.abstract}

\begin{links}
    \link{Code}{https://github.com/ml-postech/PosteL}
\end{links}

\input{tex/1.introduction}

\input{tex/2.related_work}
\input{tex/3.method}
\input{tex/4.analysis}

\input{tex/5.experiments}

\input{tex/6.conclusion}

\clearpage
\section*{Acknowledgements}
We are grateful to Sangwoo Seo for providing insightful comments on this work. This work was supported by the National Research Foundation of Korea (NRF) grant funded by the Korea government(MSIT) (RS-2024-00337955; RS-2023-00217286) and Institute of Information \& communications Technology Planning \& Evaluation (IITP) grant funded by the Korea government(MSIT) (RS-2024-00457882, National AI Research Lab Project; RS-2019-II191906, Artificial Intelligence Graduate School Program(POSTECH)).

\bibliography{aaai2026}

\appendix
\input{tex/appendix}



\end{document}

%% file: tex/0.abstract.tex
\begin{abstract}
Label smoothing is a widely studied regularization technique in machine learning.
However, its potential for node classification in graph-structured data, spanning homophilic to heterophilic graphs, remains largely unexplored. 
We introduce \emph{posterior label smoothing}, a novel method for transductive node classification that derives soft labels from a posterior distribution conditioned on neighborhood labels. 
The likelihood and prior distributions are estimated from the global statistics of the graph structure, allowing our approach to adapt naturally to various graph properties.
We evaluate our method on 10 benchmark datasets using eight baseline models, demonstrating consistent improvements in classification accuracy.
The following analysis demonstrates that soft labels mitigate overfitting during training, leading to better generalization performance, and that pseudo-labeling effectively refines the global label statistics of the graph.
\end{abstract}

%% file: tex/1.introduction.tex
\section{Introduction}

Soft label, which contains class-wise probabilities, has demonstrated remarkable success in training neural networks across various domains, including computer vision and natural language processing \citep{ lukov2022teaching, muller2019does, Szegedy_2016_CVPR, vasudeva2024geols,lsnlp_NIPS2017_3f5ee243, lsvision_21}. One of the popular approaches to obtain a soft label is label smoothing, which introduces uniform noise into the ground-truth labels.
Despite its simplicity, this technique effectively regularizes the output distribution and enhances generalization \citep{pereyra2017regularizing}. Knowledge distillation~\citep{kd_hinton_15} is another effective option, which trains a teacher model with a given one-hot label and utilizes its output as a soft-label to train the student model.

One of the most convincing explanations of why knowledge distillation works is that soft label enables the learning of \emph{``dark knowledge''} included in instances~\citep{allen-zhu2023towards, kd_hinton_15}.
Since additional information captured by the teacher model that one-hot labels cannot convey is encoded as a soft label, the student model learns richer features.

Considering the graph dataset, the relation between nodes that the graph already contains can be helpful for node classification. 
As the quote says, \emph{``You can tell a person by the company they keep,''} our idea is to encode the neighbor's information into the soft label.
More specifically, we utilize the posterior distribution, i.e., the probability of the node label given its neighbor nodes' labels.
This principle naturally generalizes both homophilic and heterophilic settings: in homophilic graphs, a target node is likely to have the same label as its neighbors, whereas in heterophilic graphs, the target node is likely to have different labels, which is supported by our theoretical analysis.

Existing approaches that generate soft labels in the graph domain build up the method based on a more specified assumption that nodes tend to share the same label with their neighboring nodes.
Based on this assumption, they construct soft labels by naively aggregating the labels of neighboring nodes~\citep{wang2021structure,zhou2023adaptive}. 
This approach aligns well with homophilic graphs, where nodes of the same class are likely to be connected, leading to improved generalization. However, it conflicts with the nature of heterophilic graphs, where edges frequently connect nodes with different labels.

Based on this intuition, we propose \textbf{Poste}rior \textbf{L}abel Smoothing (\ours{}), a novel method that derives the soft label as the posterior distribution. 
The likelihood is approximated by the product of conditional label distributions over the node’s neighborhood. 
To estimate the prior and conditional distributions, we count label occurrences at nodes and label co-occurrences across edges, thereby constructing global statistics that capture the label dependencies encoded in the graph structure. 
The resulting soft label, therefore, encapsulates rich information from both the local neighborhood structure and the global label distribution. 

Since \ours{} needs the information of label co-occurrences and global statistics of the graph, accurate information would be important for the success of our approach. 
However, we can only access the labels of train nodes, while the labels of test nodes remain unknown. 
The lack of information can result in weakening the efficacy of our method.
To address this issue, we propose an iterative pseudo-labeling procedure that utilizes pseudo-labels to re-obtain soft labels. 
Specifically, neighbor nodes' information is updated and recalculated by the prior and likelihood, which are also re-estimated by pseudo labels.

We apply our smoothing method to eight baseline neural network models, including a multi-layer perceptron and variants of graph neural networks, and test their performances on 10 graph benchmarks, including five homophilic and five heterophilic graphs. Across 80 model–dataset combinations, the soft label approach with iterative pseudo-labeling improves classification accuracy in 76 cases. 

In summary, we make the following contributions:
\begin{itemize}
\item We propose a novel \emph{posterior label smoothing} (\ours{}) method that leverages local neighborhood structure and global adjacency statistics to derive soft labels.
\item We prove that, under mild conditions, \ours{} reflects the structural properties of the graph, particularly by preventing the soft label from being similar to the neighborhood labels in heterophilic graphs.
\item We comprehensively evaluate \ours{} on eight baseline neural network models and 10 graph datasets, achieving accuracy improvements in 76 of 80 model–dataset combinations.
\end{itemize}

\if\else
Under a Markov assumption between labels, the posterior distribution of a label can be formulated as a conditional probability of the label given neighboring labels. The posterior is more robust than the naive aggregation against characteristics of graphs, as the conditional distribution accounts for whether the same or different labels are more observable given neighboring labels. 
The following theoretical analysis shows that \ours{} assigns a soft label according to the graph structure; for homophilic graphs, the soft label is dominated by the neighborhood labels, whereas for heterophilic graphs, the soft label depends on the adjacent label distribution of the graph.
\fi

%% file: tex/2.related_work.tex
\section{Related Work}


\subsection{Node Classification}
Various works leverage graph structures in different ways to perform node classification.
Early approaches such as GCN~\citep{kipf2016semi}, GraphSAGE~\citep{sage2017_5dd9db5e}, and GAT~\citep{velivckovic2017graph} aggregate neighbor representations under the homophilic assumption. 
For tackling class imbalance on homophilic graphs, GraphSMOTE~\citep{zhao2021graphsmote}, ImGAGN~\citep{qu2021imgagn}, and GraphENS~\citep{park2022graphens} have been proposed.
Meanwhile, H\textsubscript{2}GCN~\citep{beyond_58ae23d8} and U-GCN~\citep{jin2021universal} enhance performance on heterophilic graphs by aggregating representations from multi-hop neighbors. Other research focuses on adaptively learning the graph structure itself. For instance, GPR-GNN~\citep{chien2020adaptive} and CPGNN~\citep{zhu2021graph} determine which nodes to aggregate, while ChebNet~\citep{defferrard2016convolutional}, APPNP~\citep{gasteiger2018predict}, and BernNet~\citep{he2021bernnet} focus on learning appropriate filters from graph signals.

\subsection{Classification with Soft Labels}
\citet{kd_hinton_15} demonstrate that training a small student model with soft labels derived from a large teacher model’s predictions outperforms training with one-hot labels. This approach, known as knowledge distillation (KD), has proven effective for both model compression and performance improvement~\citep{jiao-etal-2020-tinybert,vision_kd_2019,recommendation_kd_18}.

Alternatively, simpler methods for generating soft labels exist. Label smoothing~\citep{Szegedy_2016_CVPR} adds uniform noise to one-hot labels, and its benefits have been widely explored. 
For instance, \citet{muller2019does} show that the label smoothing improves model calibration, while \citet{lukasik2020does} connect the label smoothing to label-correction techniques and demonstrate its utility in addressing label noise. 
The label smoothing is popular in computer vision~\citep{lukov2022teaching,vasudeva2024geols,lsvision_21} and NLP~\citep{guo2021label,song2020learning,lsnlp_NIPS2017_3f5ee243}, yet it remains relatively underexplored in the graph domain.

To our knowledge, only two studies specifically propose label smoothing techniques for node classification. SALS~\citep{wang2021structure} smooths a node’s label to match those of its neighbors, and ALS~\citep{zhou2023adaptive} aggregates neighborhood labels with adaptive refinements. However, neither work focuses on heterophilic graphs, where nodes often connect to dissimilar neighbors. Meanwhile, other studies have proposed smoothing the prediction output based on the graph structure~\citep{xie2023label,zhang2021node}, but their motivations differ substantially from the label smoothing approach investigated in this paper (e.g., they adjust output logits rather than training labels).

%% file: tex/3.method.tex
\section{Method}
\input{figure/figure_main_figure.tex}
In this section, we present our label smoothing approach for node classification and propose a new training strategy that iteratively refines soft labels via pseudo-labels obtained after training.

\subsection{Posterior Label Smoothing}
\label{subsec:posterior_node_relabeling}
Let $\set{G}=(\set{V},\set{E}, \vec{X})$ be a graph, where $\set{V}$ is a set of nodes, and $\set{E}$ is a set of edges, and $\vec{X} \in \mathbb{R}^{\lvert\set{V}\rvert \times d}$ is the $d$-dimensional node feature matrix. We consider a transductive node classification scenario in which we observe the graph structure for all nodes, including test nodes, but only the labels of nodes in the training set. For each node $i$ in a training set, we have a label $y_i \in [K]$, where $K$ is the total number of classes. Let $\vec{e}_i \in \{0,1\}^{K}$ be a one-hot encoding of $y_i$, i.e., {${e}_{ik} = 1$} if $y_i = k$ and {$\sum_k{e}_{ik} = 1$}.

\if\else
Let \(\hat{Y}_i\) be a random variable representing the soft label of node \(i\).
we assume the neighborhood’s labels are conditionally independent given $\hat{Y}_i$, i.e., 
\begin{equation}
    \label{eqn:factorization}
    {P}(\{Y_j\}_{j\in\set{N}(i)}\mid\hat{Y}_i) = \prod_{j \in \set{N}(i)} {P}(Y_j | \hat{Y}_i).
\end{equation}
\dw{We empirically verify the conditional independence assumption in \cref{sec:analysis}.}
We also define a prior distribution $P(\hat{Y}_i)$. By Bayes' rule, the posterior of $\hat{Y}_i$ given its one-hop neighbors is
\begin{equation}
    \label{eqn:bayes}
    P(\hat{Y}_i = k \mid \{Y_j = y_j\}_{j\in\set{N}(i)}) = \frac{{P}(\{Y_j = y_j\}_{j\in\set{N}(i)}|\hat{Y}_i=k){P}(\hat{Y}_i=k)}{\sum_{\ell=1}^K {P}(\{Y_j = y_j\}_{j\in\set{N}(i)}|\hat{Y}_i=\ell){P}(\hat{Y}_i=\ell)}\;.
\end{equation}
\fi

We propose an effective relabeling method to allocate a new soft label to each node based on the \emph{local neighborhood structure} and \emph{global label statistics}. 
Let \(\hat{Y}_i\) be a random variable representing the soft label of node \(i\). Given the one-hop neighborhood \(\mathcal{N}(i) = \{j \mid (i,j) \in \mathcal{E}\}\), we compute the posterior distribution of \(\hat{Y}_i\) conditioned on the labels of its neighbors using Bayes' rule:
\begin{multline}
    \label{eqn:bayes}
    P(\hat{Y}_i = k \mid \{Y_j = y_j\}_{j\in\set{N}(i)}) \\
    = \frac{{P}(\{Y_j = y_j\}_{j\in\set{N}(i)}|\hat{Y}_i=k){P}(\hat{Y}_i=k)}{\sum_{\ell=1}^K {P}(\{Y_j = y_j\}_{j\in\set{N}(i)}|\hat{Y}_i=\ell){P}(\hat{Y}_i=\ell)}\;.
\end{multline}
To obtain the likelihood ${P}(\{Y_j = y_j\}_{j\in\set{N}(i)}|\hat{Y}_i=k)$, we assume that the labels of the neighboring nodes are conditionally independent given $\hat{Y}_i$, i.e., 
\begin{equation}
\label{eqn:factorization}
P(\{Y_j\}_{j\in\mathcal{N}(i)} \mid \hat{Y}_i) = \prod_{j \in \mathcal{N}(i)} P(Y_j \mid \hat{Y}_i)\;.
\end{equation}
We empirically verify the conditional independence assumption in \Cref{sec:analysis}.

There are multiple ways to model the individual conditionals in the factorized form of \Cref{eqn:factorization}. In this work, we use the global statistics between adjacent nodes to estimate the conditional. Specifically, we define
\begin{multline}
    \label{eqn:edgewise_likelihood}
    P(Y_j=m|\hat{Y}_i=n)\\
    \coloneq \frac{\lvert \{(u,v) \mid y_v=m, y_u=n, (u,v)\in\set{E}\} \rvert}{\lvert \{(u,v) \mid y_u=n, (u,v)\in\set{E}\} \rvert}\;.
\end{multline}
We also estimate the prior distribution from global label frequencies. Concretely, we set
$P(\hat{Y}_i=m) \coloneq {\lvert \{ u \mid y_u = m\} \rvert}/{\lvert \set{V} \rvert}$. 
In \Cref{appendix:additional_exp}, we investigate alternative designs for the likelihood and compare their performances. 
\Cref{fig:main} presents an example of obtaining the posterior distribution on a toy graph.

The posterior distribution serves as a soft label for model training. However, to prevent the posterior from becoming overly confident, we incorporate a small amount of uniform noise, $\epsilon$. Additionally, because the most probable label from the posterior may not always align with the ground-truth, e.g., due to label noise or limited local information, we interpolate the posterior with the one-hot label.
To this end, we obtain the target label $\hat{\vec{e}}_i$ used for actual training as
\begin{equation}
\label{eqn:targetlabel}
\hat{\vec{e}}_{i} = \alpha \tilde{\vec{e}}_{i} + (1-\alpha) \vec{e}_{i}\;,    
\end{equation}
where $\tilde{{{e}}}_{ik} \propto P(\hat{
Y}_i = k \mid \{Y_j = y_j\}_{j\in\set{N}(i)}) + \beta \epsilon $, and $\alpha$ and $\beta$ are hyperparameters controlling the weights of interpolation and uniform noise.
By enforcing $\alpha < 1/2$, we can keep the most probable label of the target label the same as the ground-truth label, but we find that this condition is not necessary for empirical experiments. We refer to our method as \ours{} (\textbf{Poste}rior \textbf{L}abel smoothing). 
The detailed algorithm of \ours{} is shown in \Cref{alg:appendix_pls} in \Cref{appendix:alg}.


\subsection{Iterative Pseudo-labeling}
\label{subsec:ILR}
Posterior relabeling derives a node's soft label by leveraging the labels of its neighbors. However, its effectiveness can be limited by certain graph properties, particularly sparsity and label noise. For instance, if a node has no labeled neighbors, the likelihood term becomes uniform, making the posterior depend solely on the prior; if only a few neighbors are labeled and those labels are noisy, the posterior can become skewed. These challenges are more pronounced in sparse graphs: in the Cornell dataset, for example, 26.35\% of nodes have no labeled neighbors, making posterior relabeling especially difficult.

To address these limitations, we propose updating the likelihoods and priors using pseudo-labels generated for the validation and test nodes. Specifically, we first train a graph neural network with the target labels obtained from \Cref{eqn:targetlabel} and then use its predictions on the validation and test nodes to obtain pseudo-labels. We assign each unlabeled node the most probable class from the model's output. Next, we update the likelihood and prior based on these pseudo-labels, while retaining the ground-truth labels for training nodes, to recalibrate both the posterior smoothing and the resulting soft labels.

We repeat this cycle of training and re-calibration until we achieve the best validation loss, aiming to maximize node classification performance. Intuitively, if posterior label smoothing improves predictive accuracy through better likelihood and prior estimation, then the resulting pseudo-labels should, in turn, further refine these distributions, provided that the pseudo-labels contain minimal errors. 
The detailed algorithms for the training process involving iterative pseudo-labeling are presented in Algorithm 2 in Appendix A.

%% file: figure/figure_main_figure.tex
\begin{figure*}[t]
    \centering
    \includegraphics[width=\linewidth]{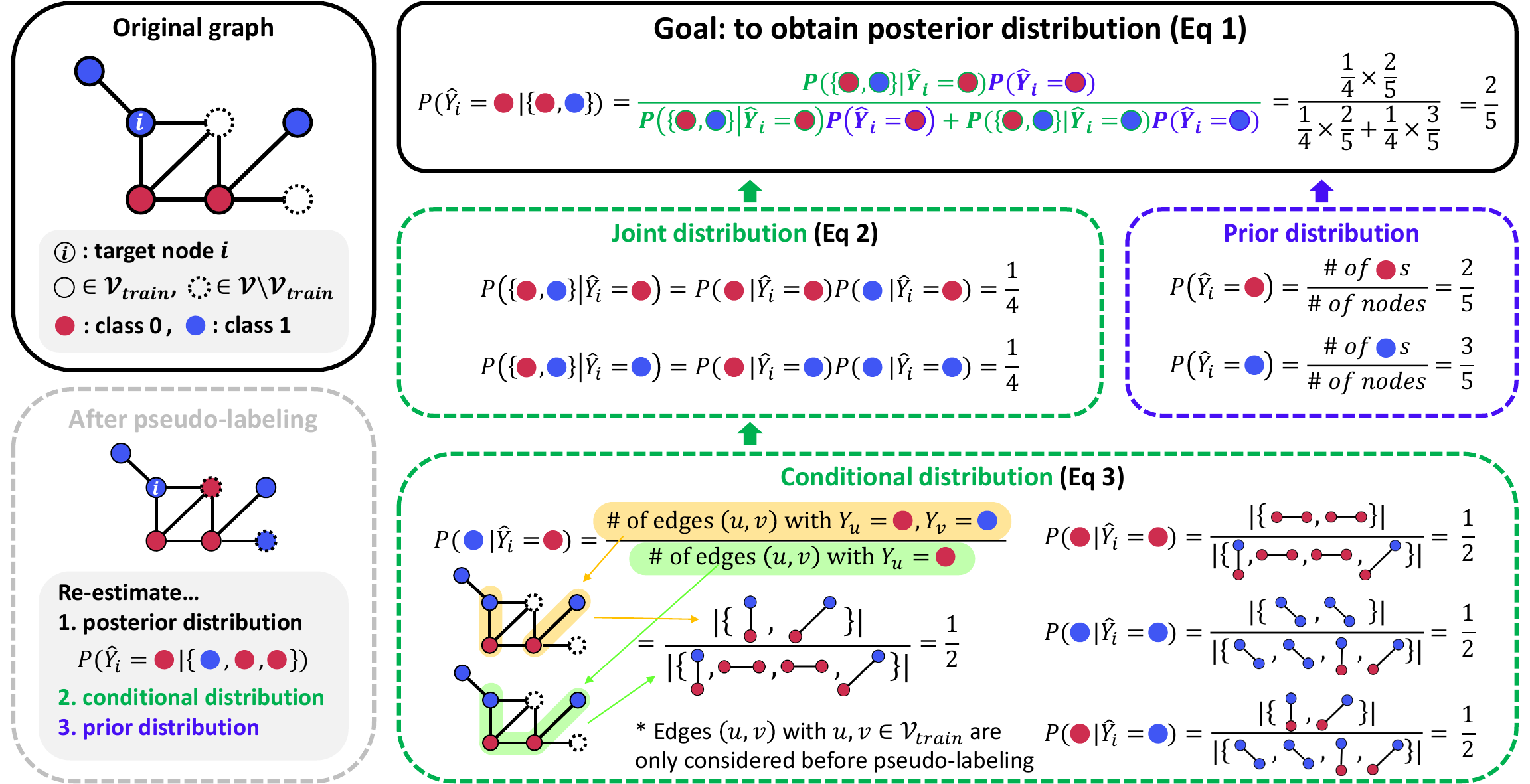}
    \caption{Overall illustration of posterior label smoothing. To relabel the node label, we compute the posterior distribution of the label given neighborhood labels. The likelihood and prior distributions are estimated from global statistics. The statistics are updated through the pseudo-labels after training, resulting in an iterative algorithm.}
    \label{fig:main}
\end{figure*}

%% file: tex/4.analysis.tex
\section{{Theoretical }Analysis of \ours{}}
We analyze how \ours{} behaves under different graph homophily and heterophily conditions in a binary classification setting. Specifically, we demonstrate how \ours{} (i) adapts label assignments based on the neighborhood label distribution and (ii) remains robust across both homophilic and heterophilic graphs. While our focus here is on binary classification for clarity, a similar argument extends to multi-class scenarios as well.

Recall from \Cref{eqn:edgewise_likelihood} that \ours{} captures the adjacency relationship via empirical edge statistics. In the binary setting, let $\set{N}_k(i)$ denote the set of neighbors of node $i$ with label $k \in \{0, 1\}$. Further, we define the \emph{class homophily} $c_k$ for each label $k$ as 
\begin{equation}
c_k:=\frac{\lvert\{(i,j)\mid(i,j)\in\mathcal{E},y_i=k,y_j=k\}\rvert}{\lvert\{(i,j)\mid(i,j)\in\mathcal{E},y_i=k\}\rvert},
\end{equation}
which measures how likely two adjacent nodes are both labeled $k$ among all edges that include a node labeled $k$.
Thus, $c_k > 0.5$ indicates that nodes labeled $k$ tend to be adjacent to others with label $k$, i.e., \emph{homophilic}, whereas $c_k < 0.5$ indicates they tend to connect to nodes with the opposite label, i.e., \emph{heterophilic}.

The following lemma states the condition under which the posterior of label $k$ is higher than $1-k$.
\begin{lemma}[Homophilic graph]
\label{lemma:homo}
    Suppose that the classes are balanced, i.e., $P(\hat{Y} = 0) = P(\hat{Y} = 1)$ and the graph is homophilic, i.e., $c_k > 1 - c_{1-k}$.
    Then, for any node $i$ with neighbors $\mathcal{N}(i)$, the posterior probability satisfies, $$P(\hat{Y}_i = k | \{Y_j=y_j\}_{j\in\set{N}(i)}) > 0.5$$
    if and only if
     $$\lvert\mathcal{N}_k(i)\rvert > \lvert\mathcal{N}_{1-k}(i)\rvert \cdot \frac{\log c_{1-k} - \log (1 - c_k)}{\log c_k - \log (1 - c_{1-k})}.$$
\end{lemma}
Intuitively, \Cref{lemma:homo} states that if the graph is sufficiently homophilic, having more neighbors with label $k$ than with label $1-k$ pushes the posterior probability for $k$ above 0.5.

A similar statement holds for heterophilic graphs.
\begin{lemma}[Heterophilic graph]
\label{lemma:hetero}
    Under the same assumptions used in \Cref{lemma:homo}, but now with a heterophilic condition, i.e., $c_k < 1 - c_{1-k}$,
    we have, $$P(\hat{Y}_i = k | \{Y_j=y_j\}_{j\in\set{N}(i)}) > 0.5$$
    if and only if
     $$\lvert\mathcal{N}_k(i)\rvert < \lvert\mathcal{N}_{1-k}(i)\rvert \cdot \frac{\log c_{1-k} - \log (1 - c_k)}{\log c_k - \log (1 - c_{1-k})}.$$
\end{lemma}
\Cref{lemma:hetero} indicates that in a heterophilic graph, having fewer neighbors of label $k$ than of label $1-k$ can make the posterior favor $k$. Detailed proofs are provided in Appendix B.
These two lemmas highlight the key difference between \ours{} and the neighborhood aggregation method in \citet{wang2021structure}. In their method, naively aggregating neighborhood labels for smoothing on heterophilic graphs results in the soft label being dominated by the majority neighborhood label. This \emph{majority dominance} contradicts the inherent heterophilic property, where nodes are more likely to connect to dissimilar labels. 
{In contrast, \Cref{lemma:hetero} demonstrates that \ours{} assigns a lower probability to the majority neighborhood label in heterophilic graphs, thereby mitigating majority dominance, while \Cref{lemma:homo} shows that \ours{} effectively retains similarity to the majority neighbor label in homophilic graphs. Moreover, nodes in heterophilic graphs tend to connect with nodes that have different labels, which implies that $|\mathcal{N}_{y_i}(i)|<|\mathcal{N}_{1-y_i}(i)|$. In this case, \ours{} preserves the ground-truth label as the most probable label. A similar effect is also found in homophilic graphs.}

\input{figure/figure_compare_sals}
Additionally, we extend the previous results to accommodate the degrees of nodes with heterophilic graphs as follows.

\input{figure/main_table_mini}

\begin{lemma}[Same degree]
    \label{lemma:same_degree}
    With a balanced heterophilic graph where $0 < c_k < 0.5$, if two nodes $n$ and $m$ have the same degree $d$ and $|\mathcal{N}_k(n)| > |\mathcal{N}_k(m)|$, then
    $$P(\hat{Y}_{n} = k \mid \{Y_j\}_{j\in\set{N}(n)}) < P(\hat{Y}_{m} = k \mid \{Y_j\}_{j\in\set{N}(m)}).$$
\end{lemma}
\Cref{lemma:same_degree} shows that for nodes with the same degree, the posterior probability for a label $k$ decreases as more neighbors share that label.

\begin{lemma}[Different degree]
    \label{lemma:diff_degree}
    With a balanced heterophilic graph where $0 < c_k < 0.5$, if there are two nodes $n$ and $m$ with degrees $d_n$ and $d_m$ such that $d_n > d_m$ and $|\mathcal{N}_k(n)| = |\mathcal{N}_k(m)|$, implying that $|\mathcal{N}_{1-k}(n)| > |\mathcal{N}_{1-k}(m)|$, then
    $$P(\hat{Y}_{n} = k \mid \{Y_j\}_{j\in\set{N}(n)}) > P(\hat{Y}_{m} = k \mid \{Y_j\}_{j\in\set{N}(m)}).$$
\end{lemma}

\Cref{lemma:diff_degree} highlights that for nodes with different degrees but the same number of neighbors labeled $k$, the posterior probability of $k$ increases for nodes with more neighbors overall. This captures how higher connectivity amplifies the effect of dissimilar neighbors in heterophilic settings.

In \Cref{fig:compare_sals}, we illustrate the difference between SALS~\citep{wang2021structure} and \ours{} with different levels of homophily. 
The first column shows three examples of a target node (represented as \texttt{T}) with different local neighborhood structures. The second and third columns show how SALS and PosteL make soft labels with homophilic and heterophilic graphs, respectively. 
We visualize \Cref{lemma:same_degree} through the heterophilic parts of the first and second graphs (rows) and \Cref{lemma:diff_degree} through the first and third graphs (rows).
As the results indicate, both methods perform similarly in homophilic graphs but not in heterophilic ones. We note that the behavior of ALS~\citep{zhou2023adaptive} is challenging to analyze due to the presence of the learnable component in their method. Except for the learnable part, the basic aggregation method of ALS is similar to that of SALS.

%% file: figure/figure_compare_sals.tex
\begin{figure}[t]
    \centering
    \includegraphics[width=\linewidth]{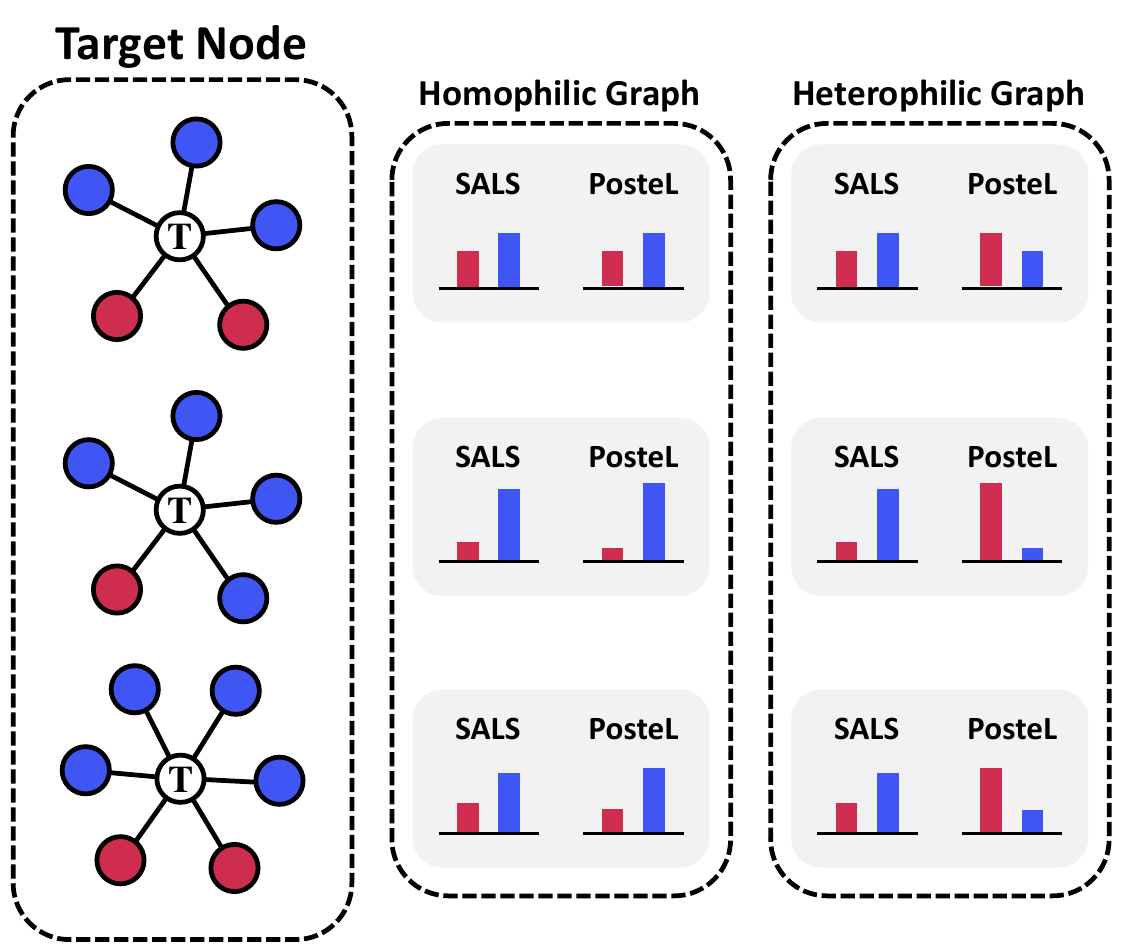}
    \caption{Toy example illustrating the difference between \ours{} and SALS~\citep{wang2021structure}. The leftmost column shows three examples of a target node (represented as \texttt{T}) with different local neighborhood structures. The second and third columns show how SALS and PosteL create soft labels with homophilic and heterophilic graphs, respectively.}
    \label{fig:compare_sals}
\end{figure}


%% file: figure/main_table_mini.tex
\begin{table*}[t!]
    \centering
    \resizebox{\linewidth}{!}{
    \begin{tabular}{lcccccccccc} 
         \toprule &\multicolumn{5}{c}{Homophilic} & \multicolumn{5}{c}{Heterophilic} \\
         \cmidrule(lr){2-6}\cmidrule(lr){7-11}
         & Cora & CiteSeer & PubMed & Computers & Photo & Chameleon & Actor & Squirrel & Texas & Cornell \\
         \midrule
         GCN & \tabnum{87.14}{1.01} & \tabnum{79.86}{0.67} & \tabnum{86.74}{0.27} & \tabnum{83.32}{0.33}& \tabnum{88.26}{0.73} & \tabnum{59.61}{2.21} & \tabnum{33.23}{1.16} & \tabnum{46.78}{0.87} & \tabnum{77.38}{3.28}& \tabnum{65.90}{4.43} \\
         +LS & \tabnum{87.77}{0.97} & \tabnum{81.06}{0.59} & \tabnum{87.73}{0.24} & \tabnum{89.08}{0.30} & \tabnum{94.05}{0.26} & \tabnum{64.81}{1.53} & \tabnum{33.81}{0.75} & \tabnum{49.53}{1.10} & \tabnum{77.87}{3.11} & \tabnum{67.87}{3.77} \\
         +KD & \tabnum{87.90}{0.90} & \tabnum{80.97}{0.56} & \tabnum{87.03}{0.29} & \tabnum{88.56}{0.36} & \tabnum{93.64}{0.31} & \tabnum{64.49}{1.38} & \tabnum{33.33}{0.78} & \tabnum{49.38}{0.64} & \tabnum{78.03}{2.62} & \tabnum{63.61}{5.57} \\
         +SALS & \tabnum{88.10}{1.08} & \tabnum{80.52}{0.85} & \tabnum{87.23}{0.13} & \tabnum{88.88}{0.54} & \tabnum{93.80}{0.31} & \tabnum{63.00}{1.75} & \tabnum{33.24}{0.92} & \tabnum{49.16}{0.77} & \tabnum{70.00}{3.93} & \tabnum{58.36}{7.54} \\
         +ALS & \tabnum{88.10}{0.85} & \tabnum{81.02}{0.52} & \tabnum{87.30}{0.30} & \tabnum{89.18}{0.36} & \tabnum{93.88}{0.27} & \tabnum{64.11}{1.29} & \tabnum{34.05}{0.49} & \tabnum{47.44}{0.76} & \tabnum{77.38}{2.13} & \tabnum{71.64}{3.28} \\
         +\ours{} & \besttabnum{88.56}{0.90} & \besttabnum{82.10}{0.50} & \besttabnum{88.00}{0.25} & \besttabnum{89.30}{0.23}& \besttabnum{94.08}{0.35} & \besttabnum{65.80}{1.23} & \besttabnum{35.16}{0.43} & \besttabnum{52.76}{0.64}  & \besttabnum{80.82}{2.79}& \besttabnum{80.33}{1.80} \\
         $\Delta$ & $+1.42(\textcolor{red}{\uparrow})$ & $+2.24(\textcolor{red}{\uparrow})$ & $+1.26(\textcolor{red}{\uparrow})$ & $+5.98(\textcolor{red}{\uparrow})$ & $+5.82(\textcolor{red}{\uparrow})$ & $+6.19(\textcolor{red}{\uparrow})$ & $+1.93(\textcolor{red}{\uparrow})$ & $+5.98(\textcolor{red}{\uparrow})$ & $+3.44(\textcolor{red}{\uparrow})$ & $+14.43(\textcolor{red}{\uparrow})$\\ 
         \midrule
         GAT & \tabnum{88.03}{0.79} & \tabnum{80.52}{0.71} & \tabnum{87.04}{0.24} & \tabnum{83.32}{0.39}& \tabnum{90.94}{0.68} & \tabnum{63.13}{1.93} & \tabnum{33.93}{2.47} & \tabnum{44.49}{0.88} & \besttabnum{80.82}{2.13}& \tabnum{78.21}{2.95}\\
         +LS & \tabnum{88.69}{0.99} & \tabnum{81.27}{0.86} & \tabnum{86.33}{0.32} & \tabnum{88.95}{0.31} & \tabnum{94.06}{0.39} & \tabnum{65.16}{1.49} & \tabnum{34.55}{1.15} & \tabnum{45.94}{1.60} & \tabnum{78.69}{4.10} & \tabnum{74.10}{4.10} \\
         +KD & \tabnum{87.47}{0.94} & \tabnum{80.79}{0.60} & \tabnum{86.54}{0.31} & \tabnum{88.99}{0.46} & \tabnum{93.76}{0.31} & \tabnum{65.14}{1.47} & \tabnum{35.13}{1.36} & \tabnum{43.86}{0.85} & \tabnum{79.02}{2.46} & \tabnum{73.44}{2.46} \\
         +SALS & \tabnum{88.64}{0.94} & \tabnum{81.23}{0.59} & \tabnum{86.49}{0.25} & \tabnum{88.75}{0.36} & \tabnum{93.74}{0.37} & \tabnum{62.76}{1.42} & \tabnum{33.91}{1.41} & \tabnum{42.29}{0.94} & \tabnum{74.92}{4.43} & \tabnum{65.57}{10.00} \\
         +ALS & \tabnum{88.60}{0.92} & \tabnum{81.09}{0.68} & \tabnum{87.06}{0.24} & \tabnum{89.57}{0.35} & \tabnum{94.16}{0.36} & \tabnum{66.15}{1.25} & \tabnum{34.05}{0.52} & \tabnum{46.85}{1.45} & \tabnum{78.03}{3.11} & \tabnum{75.08}{3.77} \\
         +\ours{} & \besttabnum{89.21}{1.08} & \besttabnum{82.13}{0.64} & \besttabnum{87.08}{0.19} & \besttabnum{89.60}{0.29}& \besttabnum{94.31}{0.31} & \besttabnum{66.28}{1.14} & \besttabnum{35.92}{0.72} & \besttabnum{49.38}{1.05} & \tabnum{80.33}{2.62}& \besttabnum{80.33}{1.81} \\
         $\Delta$ & $+1.18(\textcolor{red}{\uparrow})$ & $+1.61(\textcolor{red}{\uparrow})$ & $+0.04(\textcolor{red}{\uparrow})$ & $+6.28(\textcolor{red}{\uparrow})$ & $+3.37(\textcolor{red}{\uparrow})$ & $+3.15(\textcolor{red}{\uparrow})$ & $+1.99(\textcolor{red}{\uparrow})$ & $+4.89(\textcolor{red}{\uparrow})$ & $-0.49(\textcolor{blue}{\downarrow})$ & $+2.12(\textcolor{red}{\uparrow})$\\ 
         \midrule
         BernNet & \tabnum{88.52}{0.95} & \tabnum{80.09}{0.79} & \tabnum{88.48}{0.41} & \tabnum{87.64}{0.44}& \tabnum{93.63}{0.35} & \tabnum{68.29}{1.58} & \besttabnum{41.79}{1.01} & \tabnum{51.35}{0.73} & \tabnum{93.12}{0.65}& \tabnum{92.13}{1.64} \\
         +LS & \tabnum{88.80}{0.92} & \tabnum{80.37}{1.05} & \tabnum{87.40}{0.27} & \tabnum{88.32}{0.38} & \tabnum{93.70}{0.21} & \tabnum{69.58}{0.94} & \tabnum{39.60}{0.53} & \tabnum{52.39}{0.60} & \tabnum{91.80}{1.80} & \tabnum{90.49}{1.48} \\
         +KD & \tabnum{87.78}{0.99} & \tabnum{81.20}{0.86} & \tabnum{87.59}{0.41} & \tabnum{87.35}{0.40} & \tabnum{93.96}{0.40} & \tabnum{67.75}{1.42} & \tabnum{41.04}{0.89} & \tabnum{51.25}{0.83} & \tabnum{93.61}{1.31} & \tabnum{90.33}{2.30} \\
         +SALS & \tabnum{88.77}{0.85} & \tabnum{81.20}{0.61} & \tabnum{88.61}{0.35} & \tabnum{88.87}{0.33} & \tabnum{94.22}{0.43} & \tabnum{64.62}{0.85} & \tabnum{40.15}{1.07} & \tabnum{46.19}{0.78} & \tabnum{85.90}{4.10} & \tabnum{88.03}{3.12} \\
         +ALS & \tabnum{89.13}{0.79} & \tabnum{81.17}{0.67} & \besttabnum{89.19}{0.46} & \tabnum{89.52}{0.30} & \besttabnum{94.54}{0.32} & \tabnum{67.92}{1.07} & \tabnum{40.51}{0.61} & \tabnum{51.83}{1.31} & \tabnum{93.77}{1.31} & \tabnum{92.79}{1.48} \\
         +\ours{} & \besttabnum{89.39}{0.92} & \besttabnum{82.46}{0.67} & \tabnum{89.07}{0.29} & \besttabnum{89.56}{0.35}& \besttabnum{94.54}{0.36} & \besttabnum{69.65}{0.83} & \tabnum{40.40}{0.67} & \besttabnum{53.11}{0.87} & \besttabnum{93.93}{1.15}& \besttabnum{92.95}{1.80} \\
         $\Delta$ & $+0.87(\textcolor{red}{\uparrow})$ & $+2.37(\textcolor{red}{\uparrow})$ & $+0.59(\textcolor{red}{\uparrow})$ & $+1.92(\textcolor{red}{\uparrow})$ & $+0.91(\textcolor{red}{\uparrow})$ & $+1.36(\textcolor{red}{\uparrow})$ & $-1.39(\textcolor{blue}{\downarrow})$ & $+1.76(\textcolor{red}{\uparrow})$ & $+0.81(\textcolor{red}{\uparrow})$ & $+0.82(\textcolor{red}{\uparrow})$\\ 
         \bottomrule
    \end{tabular}
    }
    \caption{Classification accuracy on 10 node classification datasets. $\Delta$ represents the performance improvement achieved by \ours{} compared to the backbone model trained with the ground-truth label. All results of the backbone model trained with the ground-truth label are sourced from \citet{he2021bernnet}.}
    \label{tab:main_mini}
\end{table*}

%% file: tex/5.experiments.tex
\section{Experiments}
\label{sec:exp}

The experimental section is composed of two parts. First, we evaluate the performance of our method for node classification through various datasets and models. Second, we provide a comprehensive analysis highlighting the importance of each design choice.

\input{tex/experiments/1.node_classification}
\input{tex/experiments/2.empirical_analysis}

\input{tex/experiments/3.PosteL_with_limited_label}

%% file: tex/experiments/1.node_classification.tex
\subsection{Node Classification}
\label{subsec:node_classification}

In this section, we evaluate the improvements in node classification performance achieved by our method across a range of datasets and backbone models. We aim to demonstrate the robustness and consistent effectiveness of our approach across graphs with varying structural and label characteristics.

\paragraph{Datasets}
We assess the performance of our method across 10 node classification datasets. To examine the effect of our method on diverse types of graphs, we conduct experiments on both homophilic and heterophilic graphs. For the homophilic setting, we evaluate our method on five datasets: Cora, CiteSeer, and PubMed, which are citation networks where nodes represent documents and edges correspond to citation links~\citep{sen2008collective, yang2016revisiting}, as well as the Amazon co-purchase graphs Computers and Photo~\citep{mcauley2015image}, where nodes represent products and edges indicate frequent co-purchases. For the heterophilic setting, we use five datasets: Chameleon and Squirrel, which are Wikipedia networks where nodes represent pages and edges correspond to mutual links~\citep{rozemberczki2021multi}; Actor, a co-occurrence network where nodes represent actors and edges indicate co-appearances on the same Wikipedia pages~\citep{tang2009social}; and Texas and Cornell, which are webpage graphs where nodes represent web pages and edges denote hyperlinks~\citep{pei2020geom}. 
Detailed statistics of each dataset are illustrated in Appendix C.

\paragraph{Experimental Setup and Baselines}
We evaluate the performance of \ours{} across various backbone models, including a multi-layer perceptron (MLP) without graph structure, and seven widely used graph neural networks: GCN~\citep{kipf2016semi}, GAT~\citep{velivckovic2017graph}, APPNP~\citep{gasteiger2018predict}, ChebNet~\citep{defferrard2016convolutional}, GPR-GNN~\citep{chien2020adaptive}, BernNet~\citep{he2021bernnet}, and OrderedGNN~\citep{song2023ordered}.

We follow the experimental setup and backbone implementations of \citet{he2021bernnet}. Specifically, we use fixed 10 sets of train, validation, and test splits with ratios of 60\%/20\%/20\%, respectively, and measure the accuracy at the lowest validation loss. Each model is trained for 1,000 epochs, with early stopping applied if the validation loss does not improve over the last 200 epochs. 
Details of the experimental setup, including the hyperparameter search spaces and additional implementation specifics, are provided in Appendix D.

We compare our method with two domain-agnostic soft labeling methods, including label smoothing ({LS})~\citep{Szegedy_2016_CVPR} and knowledge distillation (KD)~\citep{kd_hinton_15}, as well as two label smoothing methods designed for node classification: SALS~\citep{wang2021structure} and ALS~\citep{zhou2023adaptive}. 

\input{figure/figure_joint_vs_product}

\paragraph{Results}
\Cref{tab:main_mini} reports the classification accuracy and 95\% confidence intervals for each of the three models across ten datasets. 
Complete results, including the performance of APPNP, ChebNet, MLP, GPR-GNN, and OrderedGNN, are presented in Table 9 of Appendix G.
Our method outperforms baseline methods in 76 out of 80 experimental settings. In 41 of these cases, the performance improvements exceed the 95\% confidence interval, highlighting the robustness of our approach. On the Cornell dataset, using the GCN backbone, \ours{} achieves a substantial improvement of 14.43\%.

Compared to other soft labeling methods, \ours{} consistently achieves superior performance. In particular, our method outperforms SALS and ALS, which are label smoothing methods specifically tailored for node classification, on both homophilic and heterophilic datasets. The improvements are especially significant on heterophilic datasets, indicating that the heterophily-aware label assignment strategy of \ours{} effectively enhances classification performance in heterophilic graph settings.

Nevertheless, on the Actor dataset, \ours{} exhibits relatively weaker performance compared to other datasets. 
We provide an analysis of the reasons behind this weaker performance in Appendix E.
In summary, on the Actor dataset, \ours{} behaves similarly to uniform label smoothing due to nearly identical conditional distributions across different labels.

%% file: figure/figure_joint_vs_product.tex
\begin{figure}[t]
    \centering

    \begin{subfigure}{\linewidth}
    \includegraphics[width=\linewidth]{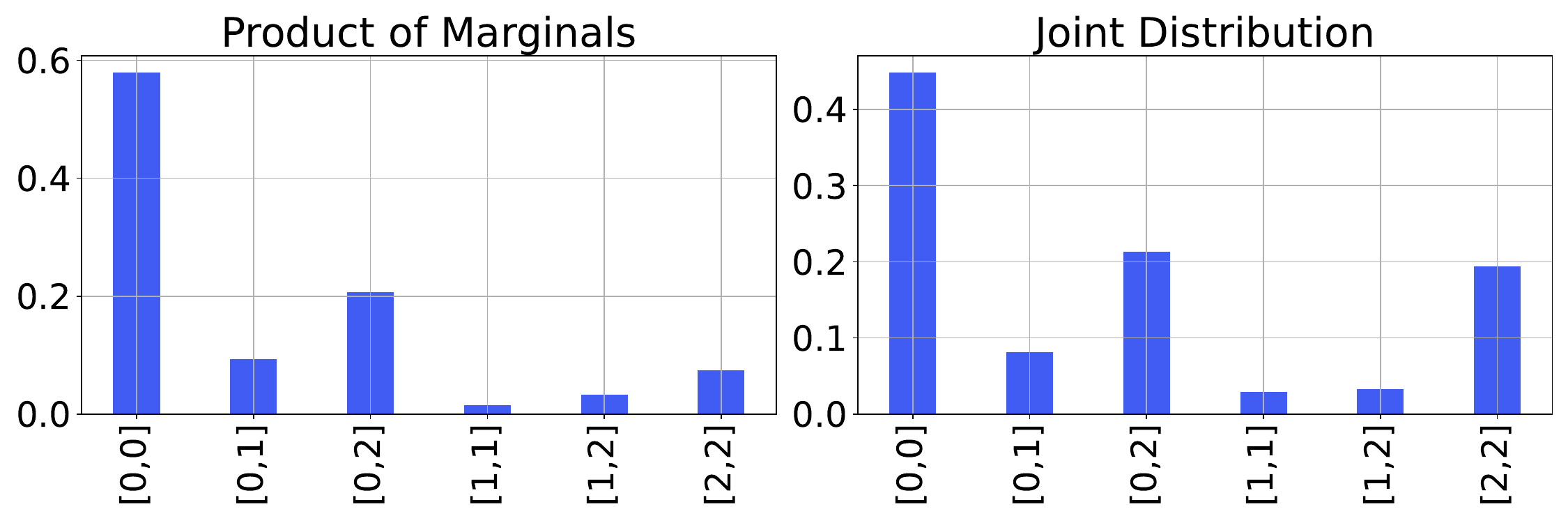}
    \caption{PubMed}
    \end{subfigure}
    \begin{subfigure}{\linewidth}
    \includegraphics[width=\linewidth]{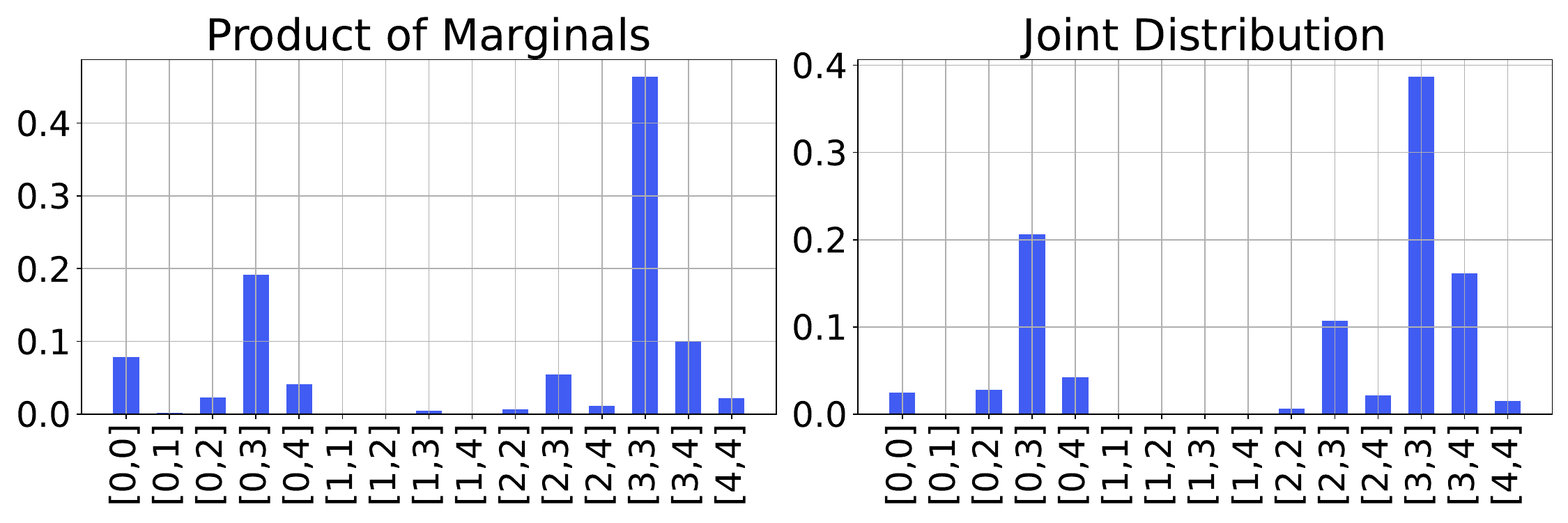}
    \caption{Cornell}
    \end{subfigure}
    
    \caption{Estimated likelihood via product of marginals $P(Y_j|Y_i=0,j\in\mathcal{N}(i))\times P(Y_k|Y_i=0,k\in\mathcal{N}(i))$ and empirical joint distribution $P(Y_j, Y_k|Y_i=0,j,k\in\mathcal{N}(i))$.}
    \label{fig:joint_vs_marginal}
\end{figure}

%% file: tex/experiments/2.empirical_analysis.tex
\subsection{Empirical Analysis}
\label{sec:analysis}

In this section, we analyze the main experimental results from multiple perspectives, including validation of the conditional independence assumption, analysis of loss curves, and computational complexity. 
Further analyses provided in Appendix G include hyperparameter sensitivity, ablation studies, scalability to large-scale graphs, and comparisons of likelihood model design choices.

\paragraph{Empirical Validation of the Conditional Independence in \Cref{eqn:factorization}}

In \Cref{eqn:factorization}, we approximate the joint conditional distribution of neighborhood labels using the product of individual conditional distributions. Although this factorization is exact when the neighborhood labels are conditionally independent given the central node's label, this assumption is often violated in real-world datasets. To empirically validate our approximation, we compare the true joint distribution $P(Y_j=n, Y_k=m \vert Y_i=l)$ to the product of marginals $P(Y_j=n \vert Y_i=l) \times P(Y_k=m \vert Y_i=l)$. \Cref{fig:joint_vs_marginal} illustrates these distributions for the case $l=0$. We observe that the product of marginals closely approximates the joint distribution, supporting the validity of our approximation.

\input{figure/figure_loss_curve_short}
\input{figure/figure_conditional_estimation_short}

\paragraph{Loss Curves Analysis}

We examine how soft labels affect GNN training dynamics by plotting the loss curves of GCN on the Squirrel dataset. 
\Cref{fig:gt_vs_ours} compares the training, validation, and test losses when using ground-truth (GT) labels and \ours{} labels.
With \ours{}, the gap between training loss and validation/test losses is noticeably smaller, indicating reduced overfitting.
While the model trained with ground-truth labels begins to overfit after 50 epochs, \ours{} remains stable through 200 epochs.

We hypothesize that correctly predicting \ours{} labels, which encode local neighborhood information, enhances the model's understanding of the graph structure and thereby improves generalization. Similar context-prediction strategies have been used as pretraining methods in previous studies~\citep{hu2019strategies, rong2020self}.
Loss curves for homophilic datasets are provided in Figure 9 and heterophilic in Figure 10 in Appendix G, showing consistent patterns across datasets.

\paragraph{Effect of Iterative Pseudo-labeling}
We analyze the impact of iterative pseudo-labeling by examining the loss curves across iterations. \Cref{fig:iteration_comparison} shows the loss curves on the Cornell dataset, where test losses consistently decrease with each iteration. In this example, the model achieves its best performance after four iterations. On average, the best performance is observed at 1.13 iterations. 
The average number of iterations used to report the results in Table 9 is detailed in Appendix F.

\paragraph{Complexity Analysis}
The computational complexity of the posterior calculation is $O(|\mathcal{E}|K)$, which is negligible compared to the time complexity of a single pass through an $L$-layer GCN with fixed hidden dimension $h$, $O(L|\mathcal{E}|h+L|\mathcal{V}|h^2)$, since the posterior is computed once before training. The training time scales linearly with the number of pseudo-labeling iterations, but experiments show that an average of 1.13 iterations is sufficient, making our approach practical. 
A detailed complexity analysis can be found in Appendix F.

%% file: figure/figure_loss_curve_short.tex
\begin{figure}[t!]
    \centering
    \begin{subfigure}{\linewidth}
        \includegraphics[width=0.49\linewidth]{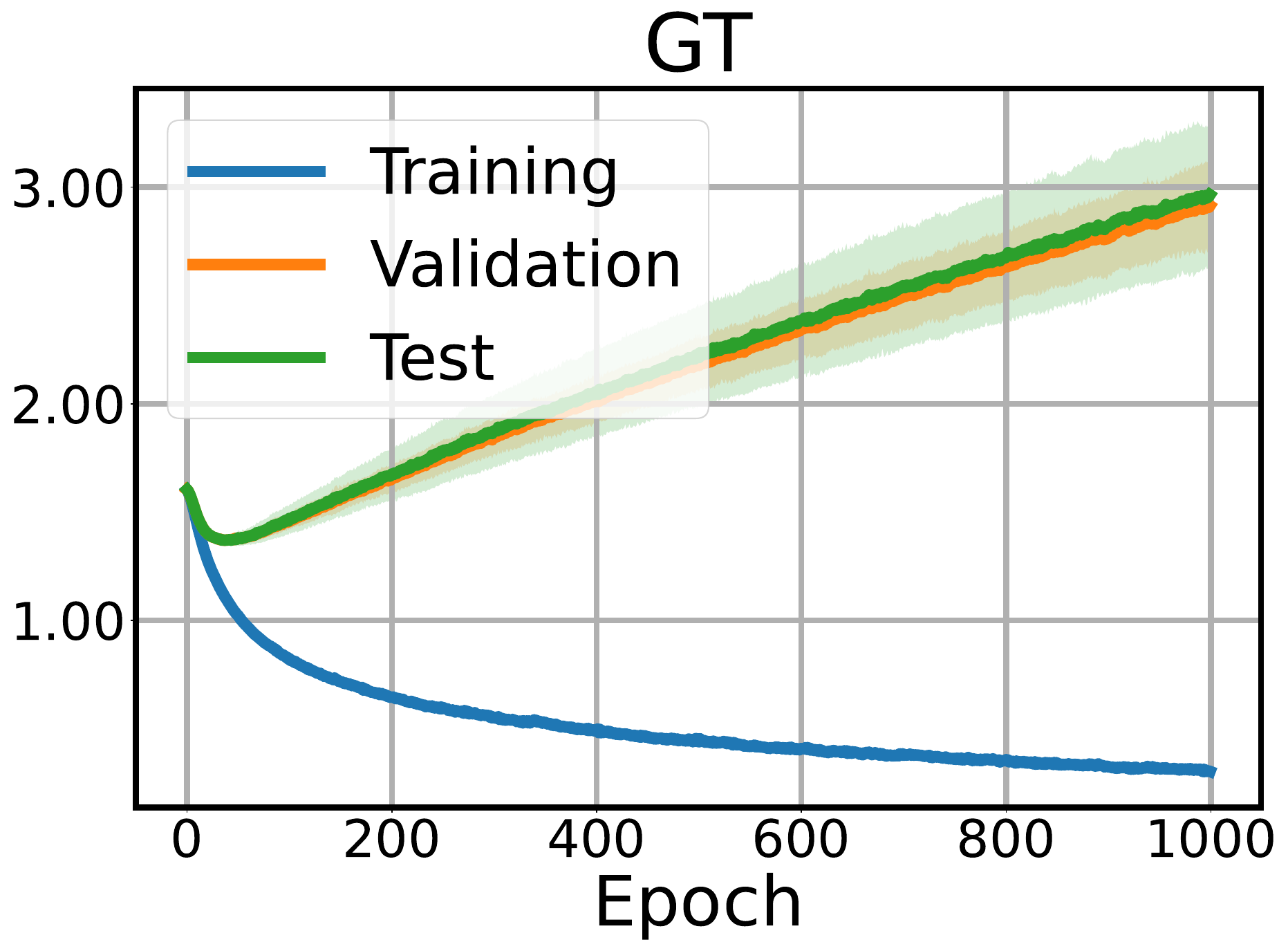}
        \includegraphics[width=0.49\linewidth]{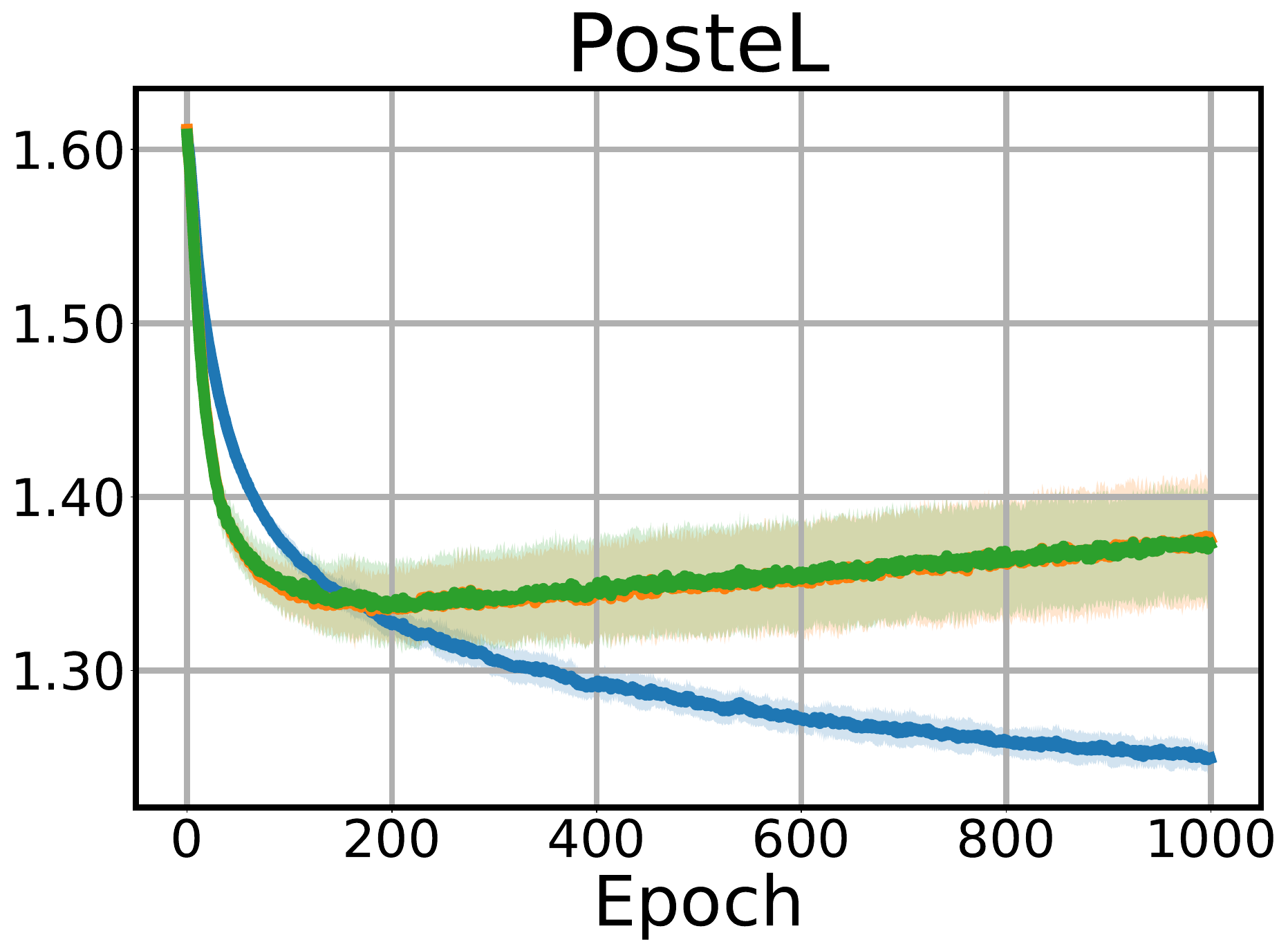}
        \caption{Ground-truth vs. \ours{} labels}
        \label{fig:gt_vs_ours}
    \end{subfigure}
    \begin{subfigure}{\linewidth}
        \centering
        \includegraphics[width=0.49\linewidth]{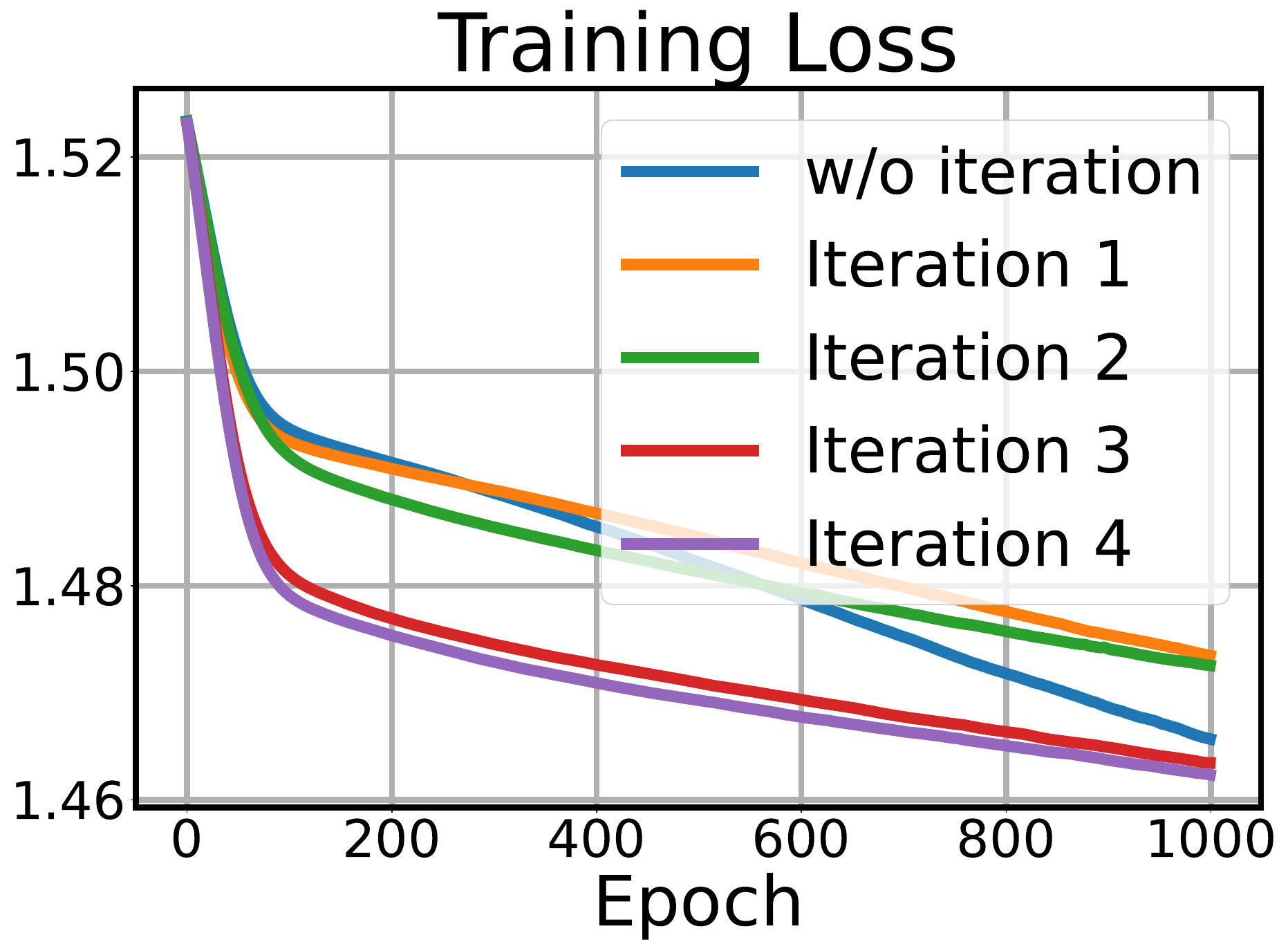}
        \includegraphics[width=0.49\linewidth]{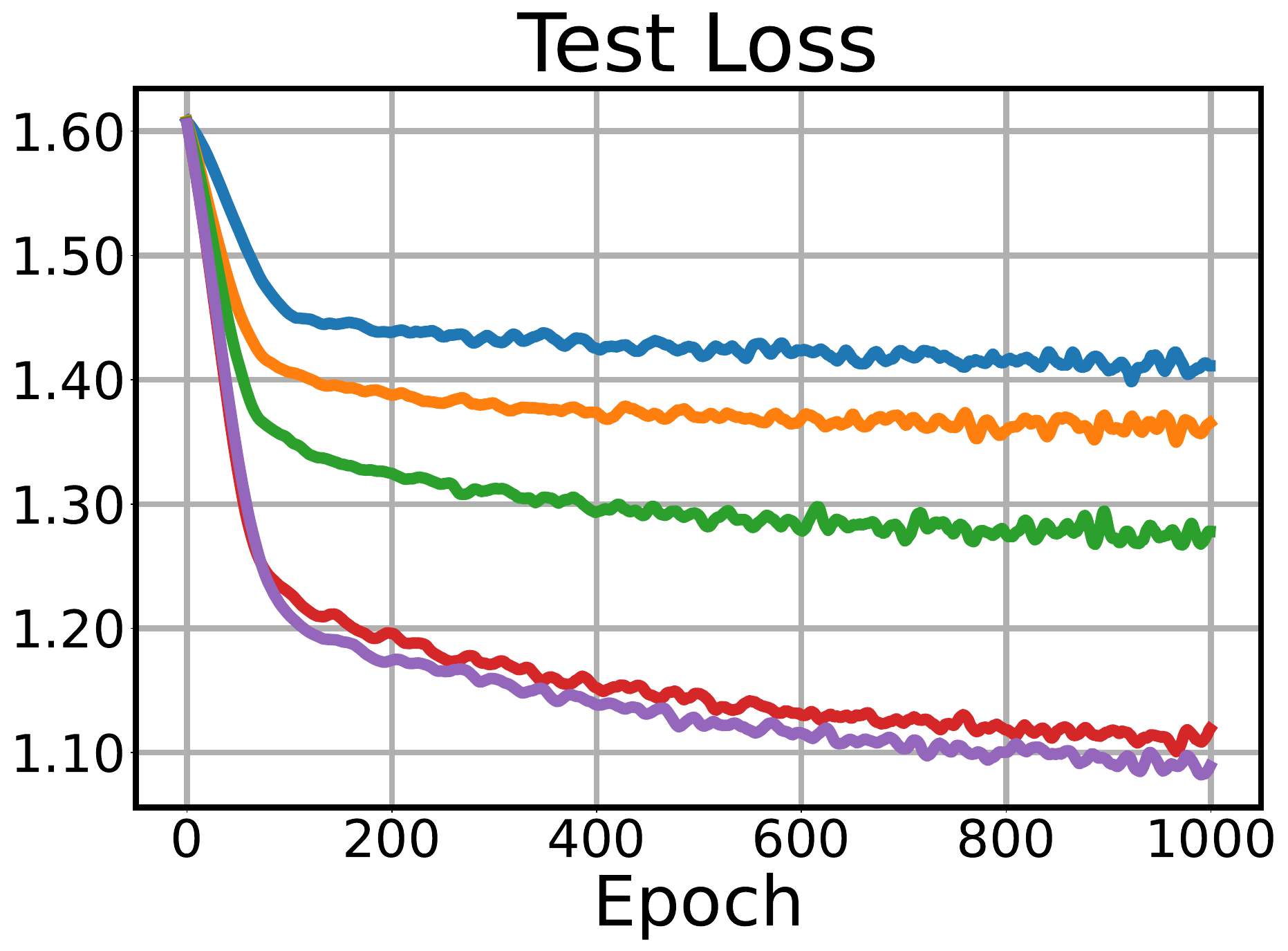}
        \caption{Loss curves across iterations}
        \label{fig:iteration_comparison}
    \end{subfigure}
    \label{fig:loss_curve_short}
    \caption{Loss curve comparisons: (a) using ground-truth (GT) labels versus \ours{} labels on the Squirrel dataset; (b) across iterations of iterative pseudo-labeling on the Cornell dataset.}
\end{figure}

%% file: figure/figure_conditional_estimation_short.tex
\begin{figure}[t]
    \centering
    
    \begin{subfigure}{\linewidth}
        \includegraphics[width=\linewidth]{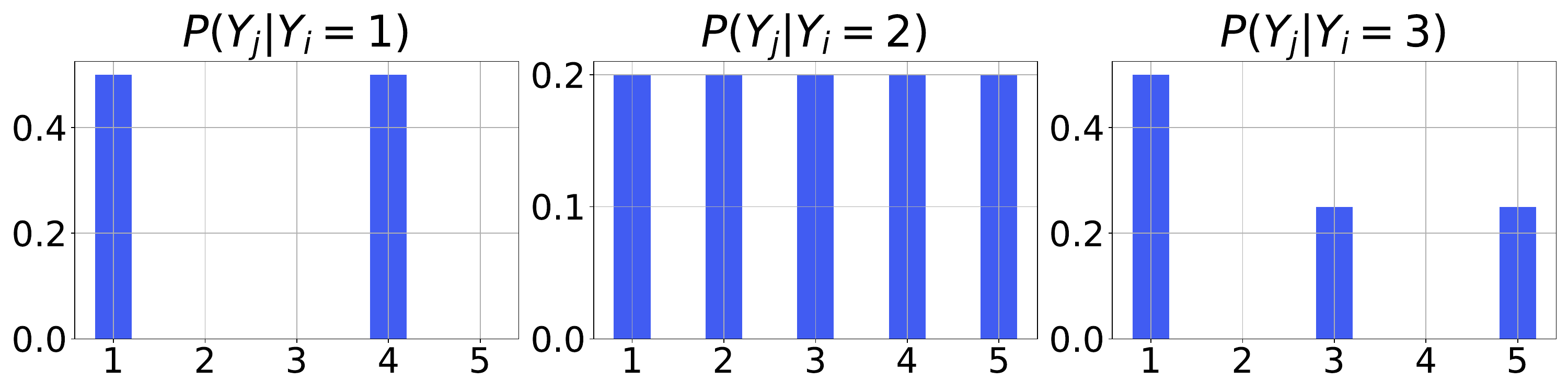}
        \caption{Training labels}
    \end{subfigure}
    
    \begin{subfigure}{\linewidth}
        \includegraphics[width=\linewidth]{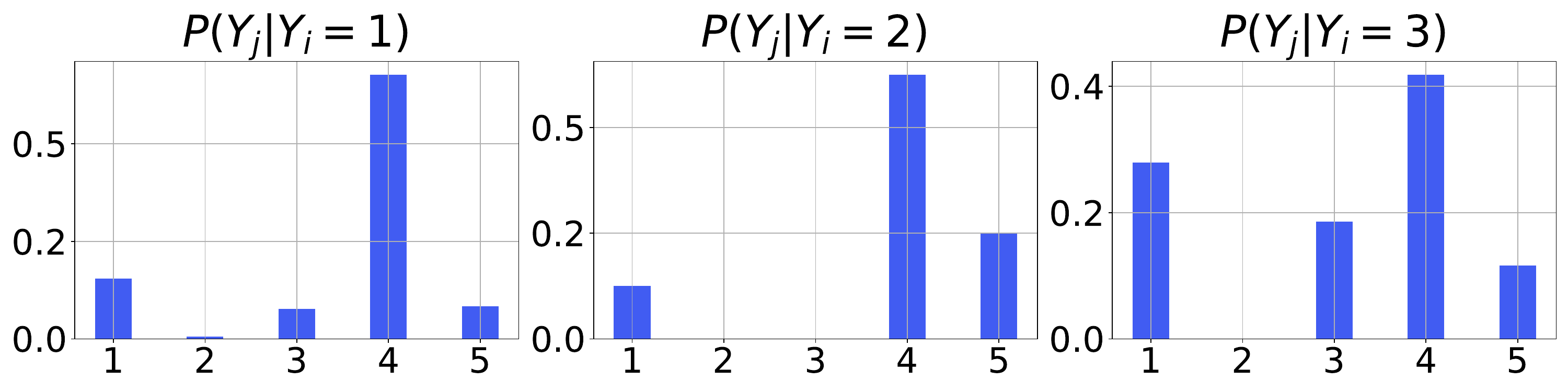}
        \caption{Training labels with pseudo-labels}
    \end{subfigure}
    
    \begin{subfigure}{\linewidth}
        \includegraphics[width=\linewidth]{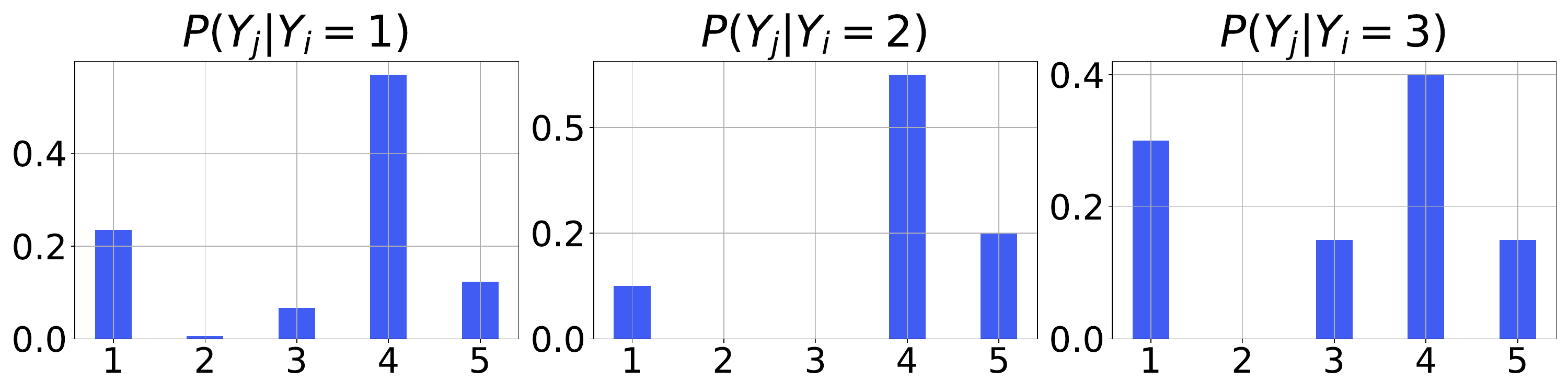}
        \caption{Ground-truth labels (Oracle)}
    \end{subfigure}
    
    \caption{Estimated conditional distributions obtained from (a) training labels only, (b) training labels combined with pseudo-labels, and (c) all ground-truth labels.}
    \label{fig:conditional_estimation_short}
\end{figure}

%% file: tex/experiments/3.PosteL_with_limited_label.tex
\subsection{Posterior Estimation with Limited Labels}
\label{sec:limited_training_labels}

Our method estimates posterior probabilities from training set statistics. However, when training labels are limited, these estimated distributions may substantially deviate from the oracle distributions, potentially leading to inaccurate posterior probabilities. To examine this issue, we evaluate the quality of the estimated distributions using only 10\% of the training data described in \Cref{subsec:node_classification}.

\Cref{fig:conditional_estimation_short} compares the conditional distributions on the Cornell dataset estimated using (1) training labels only, (2) training labels combined with pseudo-labels for validation and test nodes, and (3) all ground-truth labels. The conditional distributions estimated from limited training data show substantial deviation from the oracle distributions derived from all labels. In contrast, incorporating pseudo-labels reduces this discrepancy, yielding conditional distributions that closely match the oracle. 
We provide the same analysis on the other datasets in Appendix G.


\input{figure/table_sparse_label}

\Cref{tab:sparse_label} reports the classification accuracy of GCNs trained on 10\% of the training data. Despite the limited availability of training labels, \ours{} consistently enhances predictive accuracy. Particularly in the Texas and Cornell datasets, where pseudo-labeling substantially improves conditional distribution estimation, iterative pseudo-labeling achieves greater improvements compared to other datasets. This highlights the importance of refining conditional distributions to estimate posterior probabilities accurately.

%% file: figure/table_sparse_label.tex
\begin{table}[!t]
    \centering
    \resizebox{\linewidth}{!}{
    \begin{tabular}{lccccccc} 
         \toprule & IPL & Cora & CiteSeer & Texas & Cornell \\
         \midrule
         GCN & \textendash & \tabnum{80.66}{0.89} & \tabnum{73.52}{1.43} & \tabnum{67.05}{14.92} & \tabnum{58.36}{19.19}\\
         +\ours{} & \ding{55} & \tabnum{81.59}{1.23} & \tabnum{74.97}{1.62} & \tabnum{69.67}{14.76} & \tabnum{64.59}{15.25}\\
         +\ours{} & \checkmark & \besttabnum{82.33}{1.28} & \besttabnum{76.15}{1.05} & \besttabnum{71.48}{13.93} & \besttabnum{67.54}{16.40}\\
         \bottomrule
    \end{tabular}
    }
    \caption{Accuracy of the model trained with a limited training labels. The IPL column indicates whether iterative pseudo-labeling was applied: \ding{55} without IPL and \checkmark with IPL.}
    \label{tab:sparse_label}
\end{table}

\if\else
\begin{table}[!t]
    \centering
    \resizebox{.99\linewidth}{!}{
    \begin{tabular}{lcccccccccc} 
         \toprule & Cora & CiteSeer & Computers & Photo & Chameleon & Actor & Texas & Cornell \\
         \midrule
         GCN &  \tabnum{80.66}{0.89} & \tabnum{73.52}{1.43} & \tabnum{84.47}{0.99} & \tabnum{92.38}{0.41}  & \tabnum{45.01}{3.52} & \tabnum{24.62}{5.83} & \tabnum{67.05}{14.92} & \tabnum{58.36}{19.19}\\
         +\ours{} (w/o IPL) & \tabnum{81.59}{1.23} & \tabnum{74.97}{1.62} & \besttabnum{85.56}{0.57} & \tabnum{92.62}{0.37}  & \tabnum{51.05}{2.30} & \tabnum{30.08}{2.65} & \tabnum{69.67}{14.76} & \tabnum{64.59}{15.25}\\
         +\ours{} & \besttabnum{82.33}{1.28} & \besttabnum{76.15}{1.05} & \tabnum{85.50}{0.50} & \besttabnum{92.99}{0.31}  & \besttabnum{51.49}{2.28} & \besttabnum{31.25}{2.59} & \besttabnum{71.48}{13.93} & \besttabnum{67.54}{16.40}\\
         \bottomrule
    \end{tabular}
    }
    \caption{
    Accuracy of the model trained on a limited number of labeled nodes.
    }
    \label{tab:sparse_label}
\end{table}
\fi

%% file: tex/6.conclusion.tex
\section{Conclusion}

We introduce a novel posterior label smoothing method for node classification on graphs. By combining local neighborhoods with global label statistics, \ours{} improves model generalization. Extensive experiments on multiple datasets and models confirm its effectiveness, demonstrating significant performance gains over baseline methods.



%% file: tex/appendix.tex
\appendix
\newpage

\input{tex/appendix/1.algorithms}
\input{tex/appendix/2.lemma_proof}
\input{tex/appendix/4.dataset_statistics}
\input{tex/appendix/5.experimental_setup}
\input{tex/appendix/5.5.conditional_label_distribution_influence}
\input{tex/appendix/7.complexity_analysis}
\input{tex/appendix/8.additianal_experiments}


%% file: tex/appendix/1.algorithms.tex
\newpage
\section{Algorithms Related to Iterative Pseudo-labeling}
\label[appendix]{appendix:alg}

\Cref{alg:appendix_pls} and \Cref{alg:appendix_ipl} present the detailed algorithms for \ours{} using pseudo-labels and the training process involving iterative pseudo-labeling.

\input{figure/alg_appendix_pls}
\input{figure/alg_appendix_ipl}

%% file: figure/alg_appendix_pls.tex
\begin{algorithm}
\caption{Posterior label smoothing using pseudo-labels}\label{alg:appendix_pls}
\begin{algorithmic}
\REQUIRE The set of training nodes $\set{V}_{\oname{train}}$ and the set of nodes with pseudo-label $\set{V}_{\oname{pseudo}}$; the number of classes $K$; one-hot encoding of node labels $\{\vec{e}_i\}_{i\in\set{V}_{\oname{train}}\cup\set{V}_{\oname{pseudo}}}$; and the hyperparameters $\alpha$ and $\beta$.
\ENSURE The set of soft labels $\{\hat{\vec{e}}_i\}_{i\in\set{V}_{\oname{train}}}$.
\STATE Initialize the set of labeled nodes: 
$$\set{V}_{\oname{labeled}}=\set{V}_{\oname{train}}\cup\set{V}_{\oname{pseudo}.}$$
\STATE Estimate prior distribution for $m\in[K]$: 
$$P(\hat{Y}_i = m)=\sum_{u\in \set{V}_{\oname{labeled}}}{e}_{um}/\lvert\set{V}_{\oname{labeled}}\rvert.$$
\STATE Define the set of labeled neighbors for each node $u$: 
$$\set{N}_{\oname{labeled}}(u) = \mathcal{N}(u) \cap \set{V}_{\oname{labeled}}.$$
\STATE Estimate the empirical conditional for $n,m\in[K]$: 
\begin{multline}
P(Y_j = m| \hat{Y}_i = n)\\
\propto\sum\nolimits_{u:u\in\set{V}_{\oname{labeled}},y_u=n}\sum\nolimits_{v\in\set{N}_{\oname{labeled}}(u)}{e}_{vm}. \notag
\end{multline}
\FOR{each \( i \in \set{V}_{\oname{train}} \)}
    \STATE Approximate the likelihood: 
    \begin{multline}
    P(\{Y_j = y_j\}_{j\in\set{N}_{\oname{labeled}}(i)}|\hat{Y}_i=k) \\
    \approx \prod\nolimits_{j \in \set{N}_{\oname{labeled}}(i)} P(Y_j=y_j | \hat{Y}_i = k).
    \end{multline}
    \STATE Compute posterior distribution using \Cref{eqn:bayes}: 
    $$P(\hat{Y}_i = k \mid \{Y_j = y_j\}_{j\in\set{N}_{\oname{labeled}}(i)}).$$ 
    \STATE Add uniform noise: 
    $$\tilde{e}_{ik}\propto P(\hat{Y}_i = k \mid \{Y_j = y_j\}_{j\in\set{N}_{\oname{labeled}}(i)}) + \beta \epsilon.$$
    \STATE Obtain the soft label: $\hat{\vec{e}}_{i} = \alpha \tilde{\vec{e}}_{i} + (1-\alpha) \vec{e}_{i}$.
\ENDFOR
\end{algorithmic}
\end{algorithm}

%% file: figure/alg_appendix_ipl.tex
\begin{algorithm}
    \caption{Training algorithm with iterative pseudo-labeling}
    \label{alg:appendix_ipl}
    \begin{algorithmic}
        \REQUIRE The input graph $\set{G}=(\set{V},\set{E}, \vec{X})$; the set of training nodes $\set{V}_{\oname{train}}$ and test nodes $\set{V}_{\oname{test}}$, where $\set{V}_{\oname{train}} \cup \set{V}_{\oname{test}} = \set{V}$; one-hot encoded training labels $\{\vec{e}_i\}_{i \in \set{V}_{\oname{train}}}$; and \ours{}, as described in \cref{alg:appendix_pls}, along with its parameters $K, \alpha$, and $\beta$.
        \ENSURE Trained GNN model $f$ with pseudo-labeled nodes.

        \STATE Initialize the pseudo-labeled node set: $\set{V}_{\oname{pseudo}} = \emptyset$.
        \STATE Initialize pseudo-labels: $\{ \vec{e}_i \}_{i \in \set{V}_{\oname{pseudo}}} = \emptyset$.

        \WHILE {validation loss is decreasing}
            \STATE Apply posterior label smoothing: 
            \begin{multline}
            \{ \hat{\vec{e}}_i \}_{i \in \set{V}_{\oname{train}}}
            = \oname{PosteL}(\set{V}_{\oname{train}}, \set{V}_{\oname{pseudo}}, \\
            \{ \vec{e}_i \}_{i \in \set{V}_{\oname{training}}\cup\set{V}_{\oname{pseudo}}}, K, \alpha, \beta). \notag
            \end{multline}
            \STATE Train the GNN model $f$ to predict soft labels for the training nodes $\{ \hat{\vec{e}}_i \}_{i \in \set{V}_{\oname{train}}}$.
            \STATE Obtain pseudo-labels $\{ \bar{y}_i \}_{i \in \set{V}_{\oname{test}}}$ and their one-hot encodings $\{ \bar{\vec{e}}_i \}_{i \in \set{V}_{\oname{test}}}$ for test nodes: 
            \[\{ \bar{y}_i \}_{i \in \set{V}_{\oname{test}}} = \{\arg\max f(\set{G})_i\}_{i \in \set{V}_{\oname{test}}}.\]
            \STATE Update the pseudo-labeled node set: $\set{V}_{\oname{pseudo}} = \set{V}_{\oname{test}}$.
            \STATE Update pseudo-labels: 
            $\{ \vec{e}_i \}_{i \in \set{V}_{\oname{pseudo}}} = \{ \bar{\vec{e}}_i \}_{i \in \set{V}_{\oname{test}}}$.
        \ENDWHILE
    \end{algorithmic}
\end{algorithm}

%% file: tex/appendix/2.lemma_proof.tex
\section{The Proof of the Lemmas}
\label[appendix]{appendix:lemma_proof}

\begin{lemma}[Homophilic graph]
\label{lemma:appendix_homo}
    Suppose that the classes are balanced, i.e., $P(\hat{Y} = 0) = P(\hat{Y} = 1)$ and the graph is homophilic, i.e., $c_k > 1 - c_{1-k}$.
    Then, for any node $i$ with neighbors $\mathcal{N}(i)$, the posterior probability satisfies, $$P(\hat{Y}_i = k | \{Y_j=y_j\}_{j\in\set{N}(i)}) > 0.5$$
    if and only if
     $$\lvert\mathcal{N}_k(i)\rvert > \lvert\mathcal{N}_{1-k}(i)\rvert \cdot \frac{\log c_{1-k} - \log (1 - c_k)}{\log c_k - \log (1 - c_{1-k})}.$$
\end{lemma}

\begin{lemma}[Heterophilic graph]
\label{lemma:appendix_hetero}
    Under the same assumptions used in \Cref{lemma:appendix_homo}, but now with a heterophilic condition, i.e., $c_k < 1 - c_{1-k}$,
    we have, $$P(\hat{Y}_i = k | \{Y_j=y_j\}_{j\in\set{N}(i)}) > 0.5$$
    if and only if
     $$\lvert\mathcal{N}_k(i)\rvert < \lvert\mathcal{N}_{1-k}(i)\rvert \cdot \frac{\log c_{1-k} - \log (1 - c_k)}{\log c_k - \log (1 - c_{1-k})}.$$
\end{lemma}

\begin{proof}
    In binary classification, the conditional probabilities can be expressed in terms of class homophily $c_k$ as follows:
    \begin{align}
    P(Y_j=k|\hat{Y}_i=k) &= c_k, \notag\\
    P(Y_j=1-k|\hat{Y}_i=k) &= 1 - c_k.\notag
    \end{align}
    
    By substituting these conditional probabilities into \Cref{eqn:bayes}, the posterior probability of $\hat{Y}_i$ is given by:
    \begin{multline}
    \label{eqn:posterior_class_homophily}
    P(\hat{Y}_i=k|\{Y_j = y_j\}_{j\in\mathcal{N}(i)})  \\
    = \frac{c_k^{|\nei{k}{i}|}(1-c_k)^{|\nei{1-k}{i}|}}{c_k^{|\nei{k}{i}|}(1-c_k)^{|\nei{1-k}{i}|}+c_{1-k}^{|\nei{1-k}{i}|}(1-c_{1-k})^{|\nei{k}{i}|}},
    \end{multline}
    \begin{multline}
    P(\hat{Y}_i=1-k|\{Y_j = y_j\}_{j\in\mathcal{N}(i)}) \\
    = \frac{c_{1-k}^{|\nei{1-k}{i}|}(1-c_{1-k})^{|\nei{k}{i}|}}{c_k^{|\nei{k}{i}|}(1-c_k)^{|\nei{1-k}{i}|}+c_{1-k}^{|\nei{1-k}{i}|}(1-c_{1-k})^{|\nei{k}{i}|}},\notag
    \end{multline}
    where $\nei{k}{i}=\{y_j = k | j \in \mathcal{N}(i)\}$ and $\nei{1-k}{i}=\{y_j = 1-k | j \in \mathcal{N}(i)\}$.

    The condition under which the posterior probability of the soft label $\hat{Y}_i$ for $k$ is higher than that for $1-k$ is given by the inequality:
    \begin{equation}
    c_k^{|\nei{k}{i}|}(1-c_k)^{|\nei{1-k}{i}|} > c_{1-k}^{|\nei{1-k}{i}|}(1-c_{1-k})^{|\nei{k}{i}|}.\notag
    \end{equation}

    Taking the logarithm of both sides, the inequality expands as follows:
    \begin{multline}
    |\nei{k}{i}|\log c_k + |\nei{1-k}{i}|  \log (1 - c_k) \\
    > |\nei{1-k}{i}|  \log c_{1-k} + |\nei{k}{i}|  \log (1 - c_{1-k}). \notag
    \end{multline}

    Rearranging terms yields:
    \begin{multline}
    |\nei{k}{i}|  \left( \log c_k - \log (1 - c_{1-k}) \right) \\
    >|\nei{1-k}{i}|  \left( \log c_{1-k} - \log (1 - c_k) \right). \notag
    \end{multline}

    Finally, dividing through by $\log c_k - \log (1-c_{1-k})$, assuming it is nonzero, we obtain the condition:
    \begin{equation}
    \label{eqn:cases}
    |\nei{k}{i}| 
    \begin{cases} 
    > |\nei{1-k}{i}| \cdot \frac{\log c_{1-k} - \log (1 - c_k)}{\log c_k - \log (1 - c_{1-k})}, & \text{if } c_k > 1 - c_{1-k}, \\
    < |\nei{1-k}{i}| \cdot \frac{\log c_{1-k} - \log (1 - c_k)}{\log c_k - \log (1 - c_{1-k})}, & \text{if } c_k < 1 - c_{1-k}.
    \end{cases}
    \end{equation}

    Thus, the condition in \Cref{lemma:appendix_homo} holds in the first case of \Cref{eqn:cases}, while the condition in \Cref{lemma:appendix_hetero} holds in the second case of \Cref{eqn:cases}.
\end{proof}

\begin{lemma}[Same degree]
    \label{lemma:appendix_same_degree}
    With a balanced heterophilic graph where $0 < c_k < 0.5$, if two nodes $n$ and $m$ have the same degree $d$ and $|\nei{k}{n}| > |\nei{k}{m}|$, then
    $$P(\hat{Y}_{n} = k \mid \{Y_j\}_{j\in\set{N}(n)}) < P(\hat{Y}_{m} = k \mid \{Y_j\}_{j\in\set{N}(m)}).$$
\end{lemma}

\begin{proof}
    For binary node classification, let $d_i$ denote the degree of node $i$. We can express $|\nei{1-k}{i}|$ as $d_i - |\nei{k}{i}|$. Substituting into \Cref{eqn:posterior_class_homophily} gives:
    \begin{multline}
    \label{eqn:posterior_degree}
    P(\hat{Y}_{i} = k \mid \{Y_j=y_j\}_{j\in\set{N}(i)})\\
    =\frac{c_k^{|\nei{k}{i}|}(1 - c_k)^{d_i-|\nei{k}{i}|}}{c_k^{|\nei{k}{i}|}(1 - c_k)^{d_i-|\nei{k}{i}|} + c_{1-k}^{d_i-|\nei{k}{i}|}(1 - c_{1-k})^{|\nei{k}{i}|}}.
    \end{multline}

    
    By substituting \Cref{eqn:posterior_degree} into the inequality we aim to prove and simplifying the denominator, we obtain:
    \begin{multline}
        c_k^{|\nei{k}{n}|}(1-c_k)^{d-|\nei{k}{n}|}c_{1-k}^{d-|\nei{k}{m}|}(1-c_{1-k})^{|\nei{k}{m}|} \\
        <c_k^{|\nei{k}{m}|}(1-c_k)^{d-|\nei{k}{m}|}c_{1-k}^{d-|\nei{k}{n}|}(1-c_{1-k})^{|\nei{k}{n}|}.\notag
    \end{multline}

    Simplifying this expression leads to the following inequality:
    \begin{multline}
        (c_kc_{1-k})^{|\nei{k}{n}|-|\nei{k}{m}|}\\
        <((1-c_k)(1-c_{1-k}))^{|\nei{k}{n}|-|\nei{k}{m}|}.\notag
    \end{multline}

    Since $c_k < 1 - c_k$ and $c_{1-k} < 1 - c_{1-k}$, it follows that $(1 - c_k)(1 - c_{1-k}) > c_kc_{1-k}$. Additionally, since $|\nei{k}{n}| > |\nei{k}{m}|$, the inequality holds.
\end{proof}

\begin{lemma}[Different degree]
    \label{lemma:appendix_diff_degree}
    With a balanced heterophilic graph where $0 < c_k < 0.5$, if there are two nodes $n$ and $m$ with degrees $d_n$ and $d_m$ such that $d_n > d_m$ and $|\nei{k}{n}| = |\nei{k}{m}|$, implying that $|\mathcal{N}_{1-k}(n)| > |\mathcal{N}_{1-k}(m)|$, then
    $$P(\hat{Y}_{n} = k \mid \{Y_j\}_{j\in\set{N}(n)}) > P(\hat{Y}_{m} = k \mid \{Y_j\}_{j\in\set{N}(m)}).$$
\end{lemma}

\begin{proof}
    Using \Cref{eqn:posterior_degree}, we need to show:
    \begin{multline}
        \frac{c_k^{|\nei{k}{n}|}(1-c_k)^{d_n-|\nei{k}{n}|}}{c_k^{|\nei{k}{n}|}(1-c_k)^{d_n-|\nei{k}{n}|}+c_{1-k}^{d_n-|\nei{k}{n}|}(1-c_{1-k})^{|\nei{k}{n}|}} \\
        > \frac{c_k^{|\nei{k}{m}|}(1-c_k)^{d_m-|\nei{k}{m}|}}{c_k^{|\nei{k}{m}|}(1-c_k)^{d_m-|\nei{k}{m}|}+c_{1-k}^{d_m-|\nei{k}{m}|}(1-c_{1-k})^{|\nei{k}{m}|}}.\notag
    \end{multline}

    Given that $|\nei{k}{n}| = |\nei{k}{m}|$, we expand the inequality:
    \begin{multline}
        \frac{(1-c_k)^{d_n}}{c_k^{|\nei{k}{n}|}(1-c_k)^{d_n-|\nei{k}{n}|}+c_{1-k}^{d_n-|\nei{k}{n}|}(1-c_{1-k})^{|\nei{k}{n}|}} \\
        > \frac{(1-c_k)^{d_m}}{c_k^{|\nei{k}{n}|}(1-c_k)^{d_m-|\nei{k}{n}|}+c_{1-k}^{d_m-|\nei{k}{n}|}(1-c_{1-k})^{|\nei{k}{n}|}}.\notag
    \end{multline}

    By eliminating the denominators:
    \begin{align}
        c_k^{|\nei{k}{n}|}(1-c_k)&^{d_n+d_m-|\nei{k}{n}|} \notag\\
        +c_{1-k}&^{d_m-|\nei{k}{n}|}(1-c_k)^{d_n}(1-c_{1-k})^{|\nei{k}{n}|} \notag\\
        > &c_k^{|\nei{k}{n}|}(1-c_k)^{d_n+d_m-|\nei{k}{n}|} \notag\\
        &+ c_{1-k}^{d_n-|\nei{k}{n}|}(1-c_k)^{d_m}(1-c_{1-k})^{|\nei{k}{n}|}.\notag
    \end{align}

    Subtracting $c_k^{|\nei{k}{n}|}(1-c_k)^{d_n+d_m-|\nei{k}{n}|}$ from both sides:
    \begin{multline}
        c_{1-k}^{d_m-|\nei{k}{n}|}(1-c_k)^{d_n}(1-c_{1-k})^{|\nei{k}{n}|} \\
        > c_{1-k}^{d_n-|\nei{k}{n}|}(1-c_k)^{d_m}(1-c_{1-k})^{|\nei{k}{n}|}.\notag
    \end{multline}

    Finally, we have:
    \begin{equation}
        (1-c_k)^{d_n-d_m} > c_{1-k}^{d_n-d_m}.\notag
    \end{equation}
    
    Since $c_k, c_{1-k} < 0.5$, $1-c_k > c_{1-k}$. And given $d_n > d_m$, the inequality holds.
\end{proof}

%% file: tex/appendix/4.dataset_statistics.tex
\section{Dataset Statistics}
\label[appendix]{appendix:dataset_statistics}

We provide detailed statistics and explanations about the dataset used for the experiments in \Cref{tab:appendix_dataset_statistics} and the paragraphs below.

\input{figure/table_dataset_statistics}

\paragraph{Cora, CiteSeer, and PubMed}
Each node represents a paper, and an edge indicates a reference relationship between two papers. The task is to predict the research subjects of the papers.

\paragraph{Computers and Photo}
Each node represents a product, and an edge indicates a high frequency of concurrent purchases of the two products. The task is to predict the product category.
\paragraph{Chameleon and Squirrel}
Each node represents a Wikipedia page, and an edge indicates a link between two pages. The task is to predict the monthly traffic for each page. We use the classification version of the dataset, where labels are converted by dividing monthly traffic into five bins.
\paragraph{Actor}
Each node represents an actor, and an edge indicates that two actors appear on the same Wikipedia page. The task is to predict the category of the actors.
\paragraph{Texas and Cornell}
Each node represents a web page from the computer science department of a university, and an edge indicates a link between two pages. The task is to predict the category of each web page as one of the following: student, project, course, staff, or faculty.

%% file: figure/table_dataset_statistics.tex
\begin{table}[ht]
\begin{center}
\caption{Statistics of the dataset utilized in the experiments.}
\begin{tabular}{crrrrr}
\toprule
Dataset & \# nodes & \# edges & \# features & \# classes \\ 
\midrule
Cora & 2,708 & 5,278 & 1,433 & 7\\
CiteSeer & 3,327 & 4,552 & 3,703 & 6 \\
PubMed & 19,717 & 44,324 & 500 & 3 \\
Computers & 13,752 & 245,861 & 767 & 10 \\
Photo & 7,650 & 119,081 & 745 & 8 \\
\hline
Chameleon& 2,277 & 31,396 & 2,325 & 5 \\
Actor& 7,600 & 30,019 & 932 & 5\\
Squirrel& 5,201 & 198,423 & 2,089 & 5 \\
Texas& 183 & 287 & 1,703 & 5\\
Cornell & 183 & 277 & 1,703 & 5\\
\bottomrule
\end{tabular}
\label{tab:appendix_dataset_statistics}
\end{center}

\end{table}

%% file: tex/appendix/5.experimental_setup.tex
\section{Detailed Experimental Setup}
\label[appendix]{appendix:experimental_setup}
In this section, we provide the computer resources and search space for hyperparameters. 
Our experiments are executed on AMD EPYC 7513 32-core Processor and a single NVIDIA RTX A6000 GPU with 48GB of memory. 

\subsection{Learning Hyperparameters}
We largely follow the search space outlined in \citet{he2021bernnet}:
\begin{itemize}
\item Learning Rate: \{0.001, 0.002, 0.01, 0.05\}
\item Weight Decay: \{0, 0.0005\}
\item Model Depth: All GNNs have 2 layers.
\item Linear Layer Dropout: Fixed at 0.5.
\end{itemize}


\subsection{Model-Specific Hyperparameters}

\begin{itemize}

\item \textbf{GCN} \citep{kipf2016semi}
  \begin{itemize}
    \item 2 layers
    \item Hidden dimension: 64
  \end{itemize}

\item \textbf{GAT} \citep{velivckovic2017graph}
  \begin{itemize}
    \item 2 layers
    \item 8 attention heads, each with hidden dimension 8
  \end{itemize}

\item \textbf{APPNP} \citep{gasteiger2018predict}
  \begin{itemize}
    \item 2-layer MLP for feature extraction
    \item Hidden dimension: 64
    \item Power iteration steps: 10
    \item Teleport probability: \(\{0.1, 0.2, 0.5, 0.9\}\)
  \end{itemize}

\item \textbf{MLP}
  \begin{itemize}
    \item 2 layers
    \item Hidden dimension: 64
  \end{itemize}

\item \textbf{ChebNet}
  \begin{itemize}
    \item 2 layers
    \item Hidden dimension: 32
    \item 2 propagation steps
  \end{itemize}

\item \textbf{GPR-GNN} \citep{chien2020adaptive}
  \begin{itemize}
    \item 2-layer MLP for feature extraction
    \item Hidden dimension: 64
    \item Random walk path length: 10
    \item PPR teleport probability: \(\{0.1, 0.2, 0.5, 0.9\}\)
    \item Dropout ratio (propagation layers): 
    $$\{0.0, 0.1, 0.2, 0.3, 0.4, 0.5, 0.6, 0.7, 0.8, 0.9\}$$
  \end{itemize}

\item \textbf{BernNet} \citep{he2021bernnet}
  \begin{itemize}
    \item 2-layer MLP for feature extraction
    \item Hidden dimension: 64
    \item Polynomial approximation order: 10
    \item Dropout ratio (propagation layers): 
    $$\{0.0, 0.1, 0.2, 0.3, 0.4, 0.5, 0.6, 0.7, 0.8, 0.9\}$$
  \end{itemize}

\end{itemize}


\subsection{\ours{} Hyperparameters}
\begin{itemize}
\item Posterior Label Ratio $\alpha$: \{0.1, 0.2, 0.3, 0.4, 0.5, 0.6, 0.7, 0.8, 0.9, 1.0\}
\item Uniform Noise Ratio $\beta$: \{0, 0.1, 0.2, 0.3, 0.4, 0.5, 0.6, 0.7, 0.8, 0.9\}
\end{itemize}

These two parameters control the interpolation weight between the posterior and one-hot labels ($\alpha$) and the magnitude of uniform noise added to the posterior ($\beta$).


%% file: tex/appendix/5.5.conditional_label_distribution_influence.tex
\section{Influence of Conditional Label Distribution}
\label[appendix]{appendix:neighborhood_label_distribution}
\input{figure/figure_neighborhood_label_distribution}

In this section, we analyze the conditional label distributions and relate these observations to the behavior of \ours{}. \Cref{fig:neighborhood_label_distribution} presents the conditional label distributions for three different datasets. On the PubMed and Cornell datasets, there are clear differences in conditional distributions depending on the target node's label. Specifically, the PubMed dataset is known to exhibit homophily, meaning nodes with identical labels tend to connect more frequently, and its conditional distributions clearly reflect this characteristic. In contrast, the conditional distributions in the Actor dataset exhibit minimal variation across different target node labels. In such cases, the prior distribution tends to dominate the posterior, thereby reducing the informativeness of posterior estimates regarding neighborhood labels, potentially limiting the effectiveness of our approach. These findings align with the results in \Cref{tab:main}, which indicate that the performance improvements on the Actor dataset are smaller compared to those observed on the PubMed and Cornell datasets.

%% file: figure/figure_neighborhood_label_distribution.tex
\begin{figure*}[!ht]
    \centering
    \begin{subfigure}[b]{0.9\textwidth}
        \centering
        \includegraphics[width=0.54\linewidth]{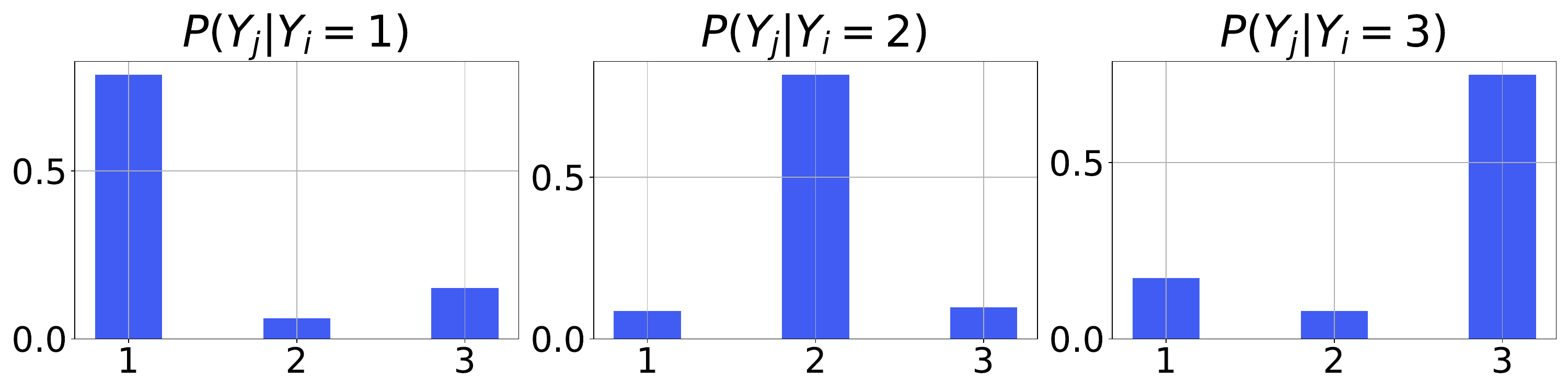 }
        \caption{PubMed}
        \label{fig:pubmed_nld}
    \end{subfigure}
    \begin{subfigure}[b]{0.9\textwidth}
        \centering
        \includegraphics[width=\linewidth]{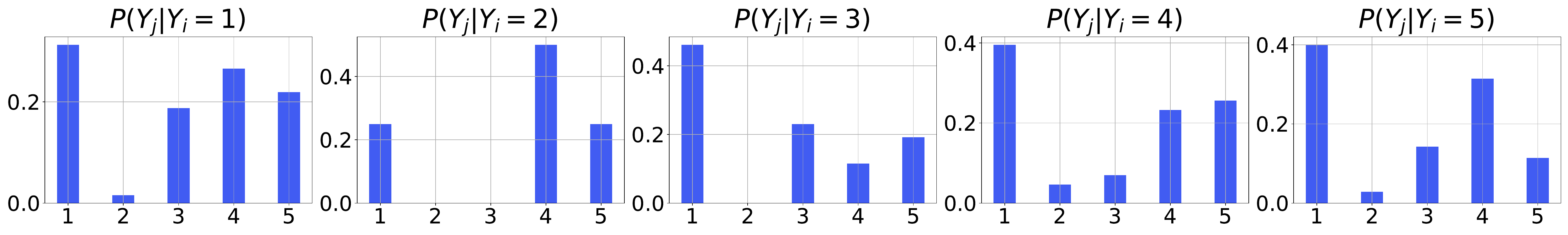}
        \caption{Cornell}
        \label{fig:texas_nld}
    \end{subfigure}
    \begin{subfigure}[b]{0.9\textwidth}
        \centering
        \includegraphics[width=\linewidth]{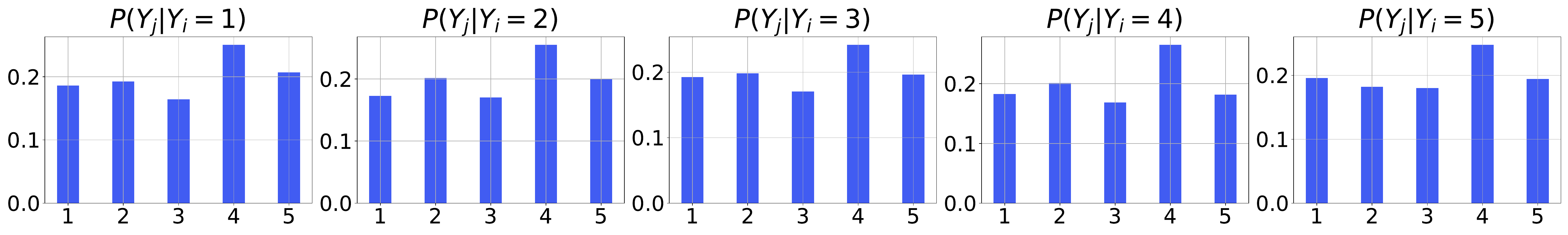}
        \caption{Actor}
        \label{fig:actor_nld}
    \end{subfigure}
    \caption{Empirical conditional distributions between two adjacent nodes. We omit the adjacent condition $(i,j)\in\set{E}$ from the figures for simplicity.}
    \label{fig:neighborhood_label_distribution}
\end{figure*}

%% file: tex/appendix/7.complexity_analysis.tex
\section{Complexity Analysis}
\label[appendix]{appendix:complexity}
In this section, we analyze the time complexity of \Cref{subsec:posterior_node_relabeling} in detail. Specifically, we first show the complexities of deriving the prior and likelihood distributions independently, and then combine these results to determine the overall complexity of computing the posterior distribution.

First, the prior distribution $P(\hat{Y}_i=m)$ can be obtained as follows:

\begin{equation}
\label{eqn:appendix_prior}
\hat{P}(Y_i=m) = \frac{\lvert \{ u \mid y_u = k\}\rvert}{\lvert \set{V} \rvert} = \frac{\sum_{u\in \set{V}}e_{um}}{\lvert\set{V}\rvert}.
\end{equation}
The time complexity of calculating \Cref{eqn:appendix_prior} is $O(\lvert\set{V}\rvert)$, so the time complexity of calculating the prior distribution for $K$ classes is $O(\lvert \set{V} \rvert K)$. 

Next, calculating the empirical conditional $P(Y_j = m| \hat{Y}_i = n)$ from \Cref{eqn:edgewise_likelihood} can be performed as follows:
\begin{equation}
\label{eqn:appendix_conditional}
P(Y_j=m|\hat{Y}_i=n)\propto\sum_{u:u\in\set{V},y_u=n}\sum_{v\in\mathcal{N}(u)}e_{vm}.
\end{equation}
The time complexity of calculating \Cref{eqn:appendix_conditional} for all possible pairs of $m$ and $n$ is $O(\sum_{u\in\set{V}}|\mathcal{N}(u)|K)$. Since $\sum_{u\in\set{V}}\mathcal{N}(u)=2\lvert\set{E}\rvert$, the time complexity for calculating empirical conditional is $O(\lvert \set{E} \rvert K)$.

The likelihood is approximated through the product of empirical conditional distributions, denoted as $P(\{Y_j = y_j\}_{j\in\set{N}(i)}|\hat{Y}_i=k) \approx \prod_{j \in \set{N}(i)} P(Y_j=y_j | \hat{Y}_i = k)$. Likelihood calculation for all training nodes operates in $O(\sum_{u\in\set{V}}\lvert\mathcal{N}(u)\rvert K)$ time complexity. So the overall computational complexity for likelihood calculation is $O(\lvert\mathcal{E}\rvert K)$.

After obtaining the prior distribution and likelihood, the posterior distribution is obtained by Bayes' rule in \Cref{eqn:bayes}. Applying Bayes' rule for $\lvert \set{V} \rvert$ nodes and $K$ classes can be done in $O(\lvert \set{V} \rvert K)$. So the overall time complexity is $O\left(\left(\lvert \set{E} \rvert+\lvert\set{V}\rvert\right)K\right)$. In most cases, $\lvert \set{V} \rvert < \lvert \set{E} \rvert$, so the time complexity of \ours{} is $O(\lvert\set{E}\rvert K)$.

In \Cref{subsec:ILR}, we introduce an iterative pseudo-labeling procedure that repeatedly refines the pseudo-labels of validation and test nodes to compute posterior labels. Because each iteration requires retraining the model from scratch, the number of iterations can become a significant bottleneck in terms of runtime. Consequently, we evaluate the iteration counts to assess this overhead. The average number of iterations for each backbone and dataset in \Cref{tab:main} is presented in \Cref{tab:num_iterations}. With an overall mean iteration count of 1.13, we argue that this level of additional time investment is justifiable for the sake of performance enhancement.

\input{figure/table_num_iterations}

\Cref{tab:overall_time} presents the average training times of \ours{} and other baselines across all datasets in \Cref{sec:exp}, using a GCN backbone. When iterative pseudo-labeling is employed, \ours{} is approximately four times slower than using the ground-truth labels, while requiring a training time comparable to knowledge distillation and ALS. If this overhead is excessive, \ours{} can be applied without iterative pseudo-labeling or with only a single pseudo-labeling iteration. In particular, \ours{} without pseudo-labeling trains approximately three times faster than knowledge distillation and ALS, and one pseudo-labeling iteration is still faster than those methods while achieving comparable accuracy. \Cref{tab:ablation} summarizes the accuracy results for each variant.

\input{figure/table_overall_time}

%% file: figure/table_num_iterations.tex
\begin{table*}[!ht]
    \caption{Average iteration counts of iterative pseudo-labeling for each backbone and dataset used to report \Cref{tab:main}.}
    \label{tab:num_iterations}
    \centering
    \resizebox{.99\linewidth}{!}{
    \begin{tabular}{ccccccccccc} 
         \toprule & Cora & CiteSeer & PubMed & Computers & Photo & Chameleon & Actor & Squirrel & Texas & Cornell \\
         \midrule
         GCN+\ours{} & 2.5 & 2.2 & 1.5 & 1 & 0.9 & 0.9 & 1.1 & 0.7 & 1.8 & 2.5 \\
         GAT+\ours{} & 1.6 & 1.8 & 1 & 1.2 & 0.7 & 0.8 & 2 & 1.1 & 3.1 & 2.4 \\
         APPNP+\ours{} & 1.9 & 2 & 1.1 & 0.8 & 1.1 & 1 & 1.1 & 0.9 & 1.4 & 2.9 \\
         MLP+\ours{} & 1.7 & 2.2 & 0.4 & 0.7 & 0.7 & 0.1 & 0.8 & 0.6 & 0.9 & 2.4 \\
         ChebNet+\ours{} & 1.6 & 2.1 & 1.2 & 0.6 & 0.6 & 1 & 0.7 & 0.7 & 2 & 2 \\
         GPR-GNN+\ours{} & 0.8 & 1.1 & 0.8 & 0.5 & 1.3 & 1 & 0.3 & 0.7 & 1.1 & 1 \\
         BernNet+\ours{} & 1.5 & 1.8 & 0.9 & 0.8 & 1 & 1.5 & 1.5 & 0.5 & 1.2 & 2.1 \\
         \bottomrule
    \end{tabular}
    }

\end{table*}

%% file: figure/table_overall_time.tex
\begin{table*}[ht!]
    \centering
    \caption{Overall training time for each smoothing method. \ours{} (w/o) refers to \ours{} without iterative pseudo-labeling, and \ours{} (1) refers to \ours{} with one iteration of pseudo-labeling.}
    \resizebox{.7\linewidth}{!}{
    \begin{tabular}{lcccccccc} 
         \toprule & GCN & +LS & +KD & +SALS & +ALS & +\ours{} & +\ours{} (w/o) & +\ours{} (1)\\
         \midrule
         time (s) & 0.83 & 0.84 & 3.54 & 0.90 & 2.87 & 3.38 & 1.08 & 1.94 \\
         \bottomrule
    \end{tabular}
    }
    \label{tab:overall_time}
\end{table*}

%% file: tex/appendix/8.additianal_experiments.tex
\newpage
\section{Additional Experiments}
\label[appendix]{appendix:additional_exp}

\paragraph{Hyperparameter Sensitivity Analysis}
\Cref{fig:appendix_hyperparameter} shows the performance with varying values of $\alpha$ and $\beta$ on GCN. The blue line indicates the performance with varying $\alpha$, and the green line shows the performance with varying $\beta$. The red dotted line represents the performance with the ground-truth label. Regardless of the values of $\alpha$ and $\beta$, the performance consistently outperforms the case using ground-truth labels, indicating that \ours{} is insensitive to $\alpha$ and $\beta$. We observe that $\alpha$ values greater than 0.8 may harm training, suggesting the necessity of interpolating ground-truth labels.

\input{figure/figure_appendix_hyperparameter}

\paragraph{Scalability to Large-scale Graphs}
We evaluated the runtime of \ours{} on the ogbn-products dataset~\citep{hu2020open}, which contains 2,449,029 nodes and 61,859,140 edges, to assess its computational efficiency on a large-scale graph. We measured the time excluding the training time for iterative pseudo-labeling. Generating soft labels with \ours{} takes 5.65 seconds, whereas a single training epoch requires 19.11 seconds, indicating that \ours{} can generate soft labels efficiently even on large-scale graphs.

\input{figure/table_ogbn_products}

\Cref{tab:ogbn-products} presents the performance of \ours{} on the ogbn-products dataset using GCN. Although the improvement is not statistically significant, \ours{} achieves the highest performance compared to other smoothing methods.

\paragraph{Design Choices of Likelihood Model}

We investigate alternative likelihood designs and introduce two \ours{} variants: \ours{} (normalized) and \ours{} (local-$H$).
In \Cref{eqn:edgewise_likelihood}, each edge equally contributes to the conditional probability, which can over-rely on high-degree nodes. To mitigate this, \ours{} (normalized) adjusts edge contributions based on node degrees:
\begin{multline}
    P^{\text{norm.}}(Y_j=m|\hat{Y}_i=n) \\
    \coloneq \frac{\sum_{y_u=n}\sum_{v\in\mathcal{N}(u)}\frac{1}{\lvert \set{N}(u) \rvert} \cdot\mathds{1}[y_v = m]}{\lvert \{y_u=n \mid u \in \set{V} \}\rvert}\;,
\end{multline}
where $\mathds{1}$ is an indicator function.

In \ours{} (local-$H$), likelihoods and priors are estimated from H-hop ego graphs, emphasizing local neighborhood statistics:
\begin{multline}
    P^{\text{local-$H$}}(Y_j=m|\hat{Y}_i=n) \\
    \coloneq \frac{\lvert \{(u,v) | y_v=m, y_u=n, (u,v)\in\set{E}_i^{(H)} \rvert}{\lvert \{(u,v) | y_u=n, (u,v)\in\set{E}_i^{(H)} \rvert}\;,
\end{multline}
where $\set{E}_i^{(H)}$ is the set of edges in the $H$-hop neighborhood of node $i$, denoted as $\set{N}^{(H)}(i)$. These variants allow us to assess the importance of global versus local statistics in the smoothing process.

\input{figure/table_local_likelihood}
\input{figure/table_ablations}

\Cref{tab:local} shows the comparison between these variants. The likelihood with global statistics, e.g., \ours{} and \ours{} (normalized), performs better than the local likelihood methods, e.g., \ours{} (local-1) and \ours{} (local-2) in general, highlighting the importance of simultaneously utilizing global statistics. Especially in the Cornell dataset, a significant performance gap between \ours{} and \ours{} (local) is observed. \ours{} (normalized) demonstrates similar performance to \ours{}.

\paragraph{Ablation Studies}

We conduct ablation studies on three components of \ours{}: posterior smoothing (PS), uniform noise (UN), and iterative pseudo-labeling (IPL), to evaluate their individual contributions. \Cref{tab:ablation} summarizes the results.

The best performance is achieved when all components are included, highlighting their collective importance. IPL consistently improves performance across most datasets, especially Cornell, although omitting IPL still yields competitive results. Adding uniform noise further enhances performance on several datasets. Notably, \ours{} surpasses using only uniform noise, a common label smoothing baseline.

\paragraph{Visualization of Node Embeddings} \Cref{fig:tsne} presents the t-SNE~\citep{van2008visualizing} plots of node embeddings from the GCN with the Chameleon and Squirrel datasets. The node color represents the label. For each dataset, the left plot visualizes the embeddings with the ground-truth labels, while the right plot visualizes the embeddings with \ours{} labels. The visualization shows that the embeddings from the soft labels form tighter clusters compared to those trained with the ground-truth labels. 
This visualization results coincide with the t-SNE visualization of the previous work of \citet{muller2019does}.

\input{figure/figure_tsne}

\paragraph{Extended Results of the Main Text}
We extend the experimental results presented in the main text to additional GNN backbones and datasets.
\Cref{tab:main} reports the classification accuracy for eight additional backbones, including APPNP~\citep{gasteiger2018predict}, ChebNet~\citep{defferrard2016convolutional}, MLP, GPR-GNN~\citep{chien2020adaptive}, and OrderedGNN~\citep{song2023ordered}.
\Cref{fig:appendix_baselinewise_loss_curve_homo,fig:appendix_baselinewise_loss_curve_hetero} show training/validation/test loss curves comparing ground-truth labels, SALS~\citep{wang2021structure} labels, and \ours{} labels on all 10 datasets used in our experiments.
\Cref{fig:conditional_estimation_full} shows the estimated conditional distributions based on (i) training labels only, (ii) training labels combined with pseudo-labels, and (iii) all ground-truth labels.

\input{figure/main_table}
\input{figure/figure_appendix_baselinewise_loss_curve_homo}
\input{figure/figure_appendix_baselinewise_loss_curve_hetero}
\input{figure/figure_conditional_estimation_full}

%% file: figure/figure_appendix_hyperparameter.tex
\begin{figure}[h!]
    \centering
    \begin{subfigure}{0.49\columnwidth}
        \centering
        \includegraphics[width=\linewidth]{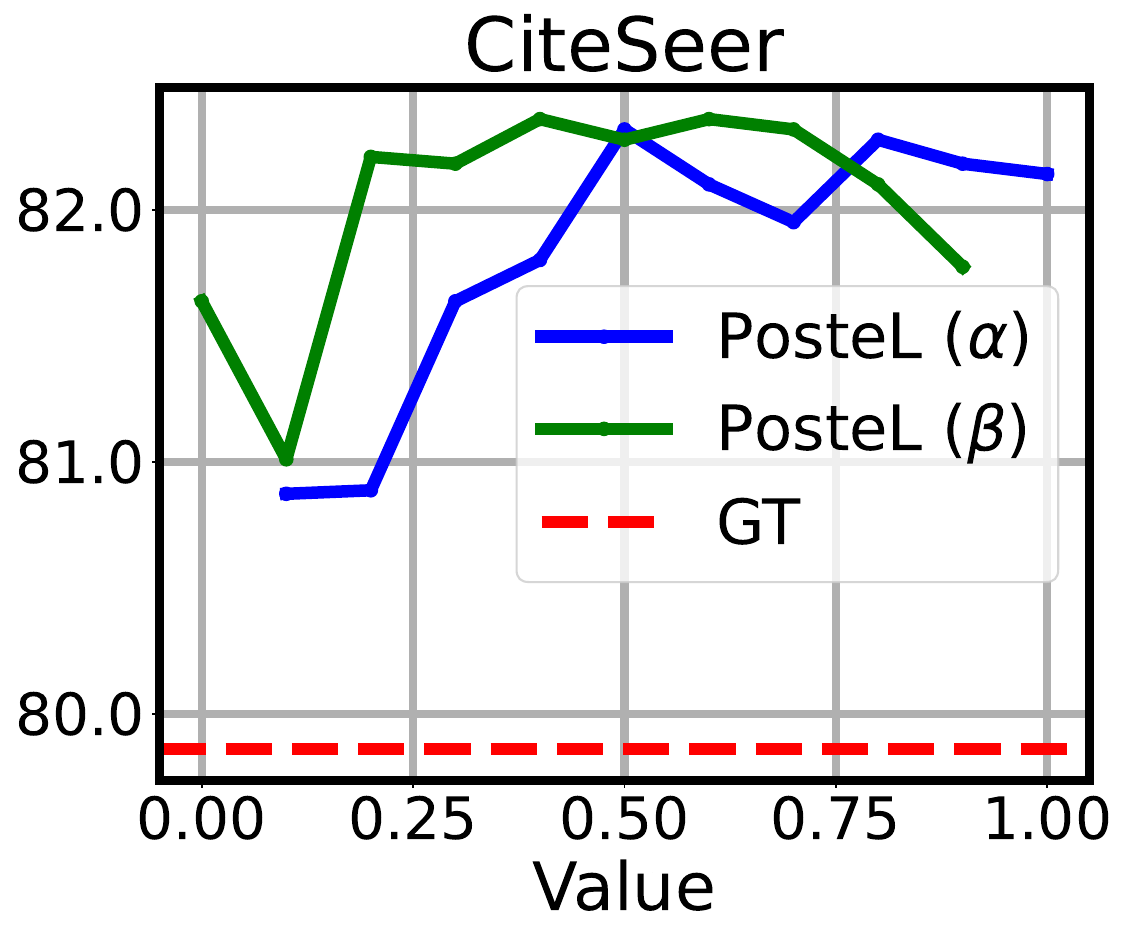}
    \end{subfigure}
    \hspace{-1mm}
    \begin{subfigure}{0.49\columnwidth}
        \centering
        \includegraphics[width=\linewidth]{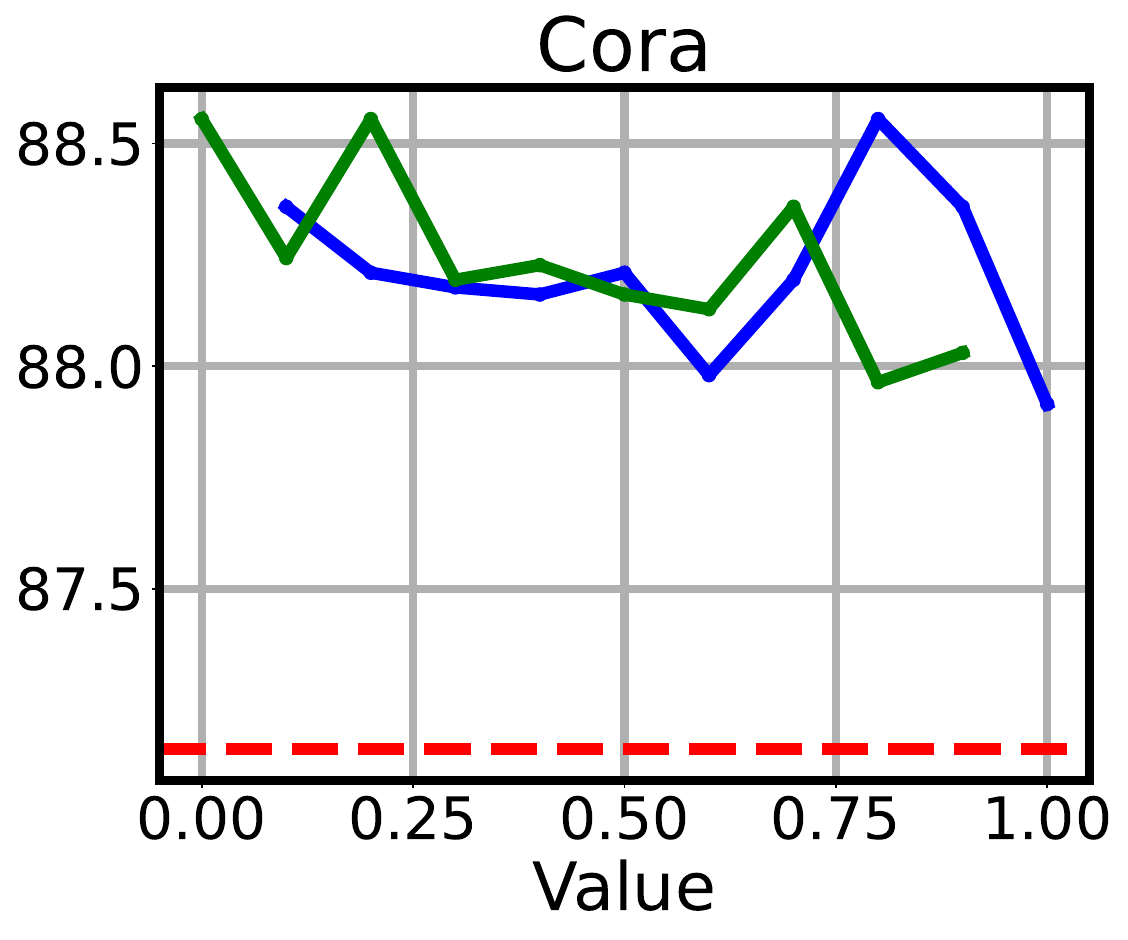}
    \end{subfigure}
    \hspace{-1mm}
    \begin{subfigure}{0.49\columnwidth}
        \centering
        \includegraphics[width=\linewidth]{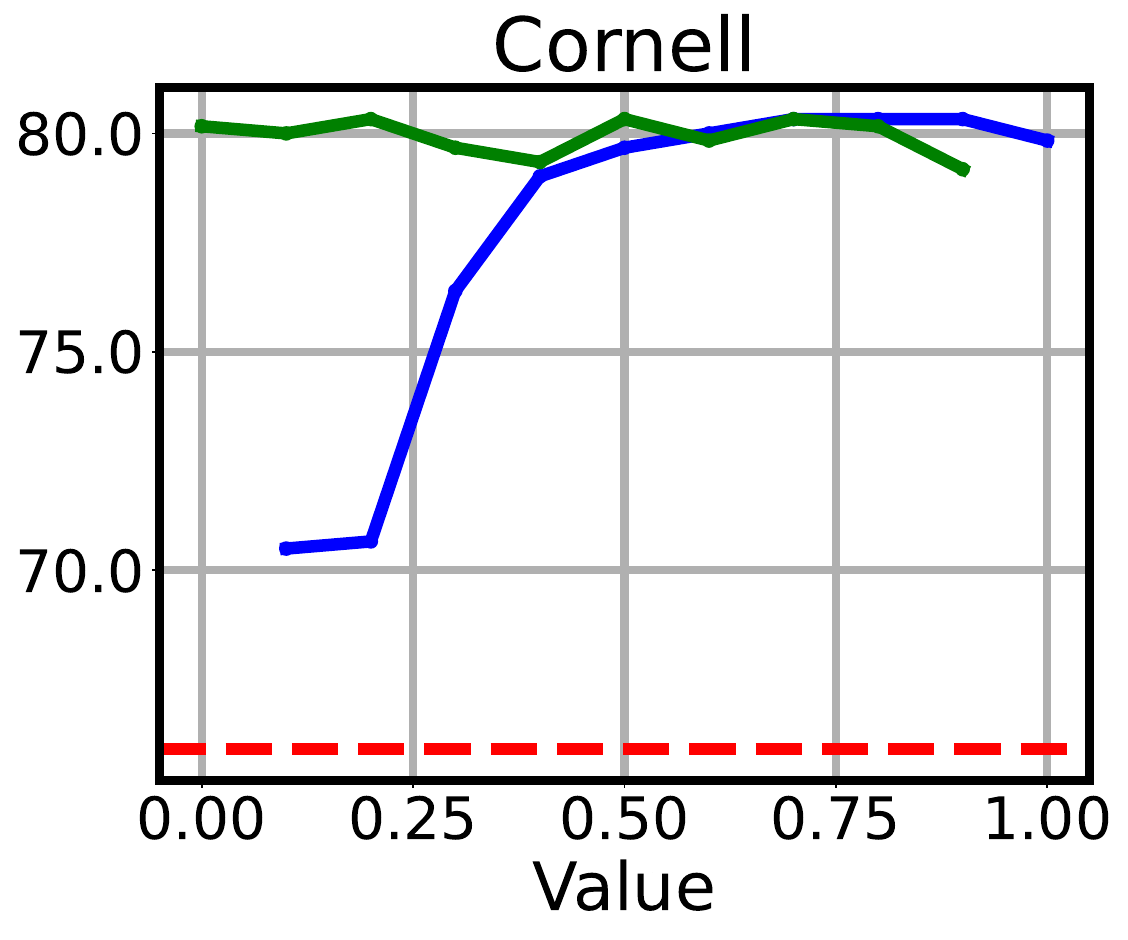}
    \end{subfigure}
    \begin{subfigure}{0.49\columnwidth}
        \centering
        \includegraphics[width=\linewidth]{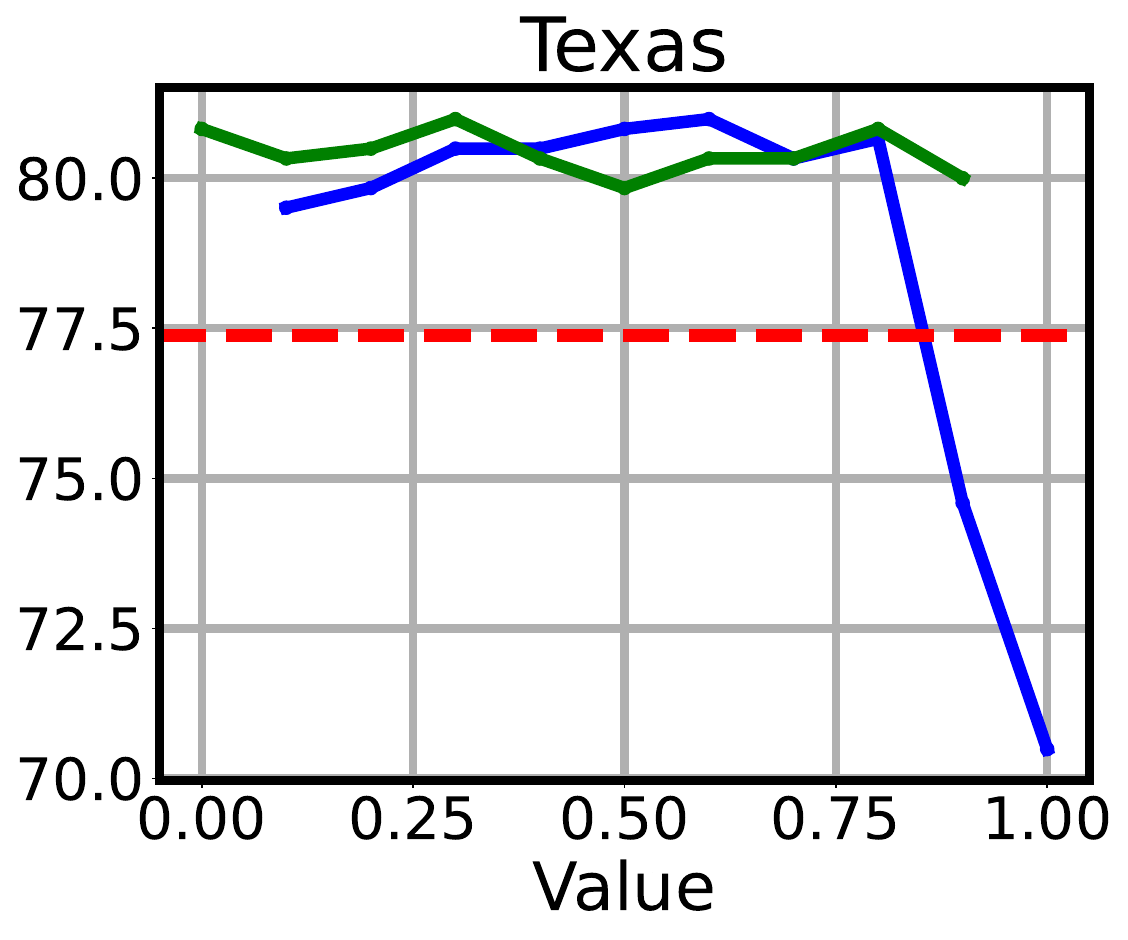}
    \end{subfigure}
    \caption{Hyperparameter sensitivity analysis on GCN.}
    \label{fig:appendix_hyperparameter}
\end{figure}

%% file: figure/table_ogbn_products.tex
\begin{table}[ht!]
    \centering
    \caption{The accuracy of label smoothing methods on the ogbn-products dataset using GCN.}
    \resizebox{\linewidth}{!}{
    \begin{tabular}{lccccc} 
         \toprule & GCN & +LS & +SALS & +ALS & +\ours{} \\
         \midrule
         Products & \tabnum{80.62}{0.68} & \tabnum{80.99}{0.50} & \tabnum{81.12}{0.13} & \tabnum{80.46}{0.38} & \besttabnum{81.20}{0.68}\\
         \bottomrule
    \end{tabular}
    }
    \label{tab:ogbn-products}
\end{table}

%% file: figure/table_local_likelihood.tex
\begin{table*}[!ht]
    \caption{Classification accuracy with various choices of likelihood model. \ours{} (local-1) and (local-2) indicate that the likelihood is estimated within one- and two-hop neighbors of a target node, respectively. \ours{} (norm.), shortened from \ours{} (normalized), indicates that the likelihood is normalized based on the degree of a node.}
    \centering
    \resizebox{.9\linewidth}{!}{
    \begin{tabular}{lcccccccc} 
         \toprule & Cora & CiteSeer & Computers & Photo & Chameleon & Actor & Texas & Cornell \\
         \midrule
         GCN & \tabnum{87.14}{1.01} & \tabnum{79.86}{0.67} &  \tabnum{83.32}{0.33}& \tabnum{88.26}{0.73} & \tabnum{59.61}{2.21} & \tabnum{33.23}{1.16} &  \tabnum{77.38}{3.28}& \tabnum{65.90}{4.43} \\
         +\ours{} (local-1) & \tabnum{88.26}{1.07} & \tabnum{81.42}{0.46} & \tabnum{89.08}{0.31}& \tabnum{93.61}{0.40} & \tabnum{65.36}{1.25} & \tabnum{33.48}{1.03} & \tabnum{79.02}{3.11}& \tabnum{71.97}{4.10} \\
         +\ours{} (local-2) & \tabnum{88.62}{0.97} & \tabnum{81.92}{0.42} & \tabnum{88.62}{0.48} & \tabnum{93.95}{0.37} & \tabnum{65.10}{1.55} & \tabnum{34.63}{0.46}  & \tabnum{78.20}{2.79} & \tabnum{73.28}{4.10} \\
         +\ours{} (norm.) & \besttabnum{89.00}{0.99} & \tabnum{81.86}{0.70} & \besttabnum{89.30}{0.39}& \besttabnum{94.13}{0.39} & \besttabnum{66.00}{1.14} & \tabnum{34.90}{0.63} & \tabnum{80.33}{2.95}& \tabnum{80.00}{1.97} \\
         +\ours & \tabnum{88.56}{0.90} & \besttabnum{82.10}{0.50} & \besttabnum{89.30}{0.23}& \tabnum{94.08}{0.35} & \tabnum{65.80}{1.23} & \besttabnum{35.16}{0.43} & \besttabnum{80.82}{2.79}& \besttabnum{80.33}{1.80} \\
         \bottomrule
    \end{tabular}
    }
    \label{tab:local}
\end{table*}

%% file: figure/table_ablations.tex
\begin{table*}[!ht]
    \caption{
    Ablation studies on three main components of \ours{} on GCN. PS stands for posterior label smoothing, UN stands for uniform noise, and IPL stands for iterative pseudo-labeling. We use \checkmark to indicate the presence of the corresponding component in training and \ding{55} to indicate its absence. IPL with one indicates the performance with a single pseudo-labeling step.
    }
    \centering
    \resizebox{.9\linewidth}{!}{
    \begin{tabular}{ccccccccccc} 
         \toprule 
         PS & UN & IPL & Cora & CiteSeer & Computers & Photo & Chameleon & Actor & Texas & Cornell \\
         
         \midrule
         \ding{55} & \ding{55} & \ding{55} & \tabnum{87.14}{1.01} & \tabnum{79.86}{0.67} &  \tabnum{83.32}{0.33}& \tabnum{88.26}{0.73} & \tabnum{59.61}{2.21} & \tabnum{33.23}{1.16} &  \tabnum{77.38}{3.28}& \tabnum{65.90}{4.43} \\ [1pt]
         \checkmark & \ding{55} & \ding{55} & \tabnum{88.11}{1.22} & \tabnum{80.95}{0.52} &  \tabnum{88.86}{0.40}& \tabnum{93.55}{0.30} & \tabnum{64.53}{1.23} & \tabnum{33.48}{0.62} &  \tabnum{78.52}{2.46}& \tabnum{68.52}{4.43} \\[1pt]
         \ding{55} & \checkmark & \ding{55} & \tabnum{87.77}{0.97} & \tabnum{81.06}{0.59} &  \tabnum{89.08}{0.30}& \tabnum{94.05}{0.26} & \tabnum{64.81}{1.53} & \tabnum{33.81}{0.75} &  \tabnum{77.87}{3.11}& \tabnum{67.87}{3.77} \\
         [1pt]
          \checkmark & \ding{55} & \checkmark &  \besttabnum{88.56}{0.90} & \tabnum{81.64}{0.57} &  \tabnum{88.70}{0.27} & \tabnum{93.70}{0.37} & \tabnum{64.25}{1.93} & \tabnum{34.71}{0.76} &  \besttabnum{80.82}{2.79} & \tabnum{80.16}{1.97} \\
          [1pt]
          \checkmark & \checkmark & \ding{55} &  \tabnum{87.83}{0.92} & \tabnum{82.09}{0.44} &  \tabnum{89.17}{0.31} & \tabnum{93.98}{0.34} & \besttabnum{66.19}{1.60} & \tabnum{34.91}{0.48} &  \tabnum{79.51}{3.61} & \tabnum{71.97}{5.25} \\
          [1pt]
         \checkmark & \checkmark & 1 &  \tabnum{87.96}{0.90} & \besttabnum{82.33}{0.52} &  \tabnum{89.16}{0.30} & \tabnum{94.06}{0.27} & \tabnum{65.89}{1.51} & \tabnum{34.96}{0.48} &  \tabnum{80.16}{2.79} & \besttabnum{80.33}{1.97} \\
         [1pt]
          \checkmark & \checkmark & \checkmark &  \besttabnum{88.56}{0.90} &\tabnum{82.10}{0.50} &  \besttabnum{89.30}{0.23}& \besttabnum{94.08}{0.35} & \tabnum{65.80}{1.23} & \besttabnum{35.16}{0.43} & \besttabnum{80.82}{2.79}& \besttabnum{80.33}{1.80} \\
         \bottomrule
    \end{tabular}
    }
    \label{tab:ablation}
\end{table*}

%% file: figure/figure_tsne.tex
\begin{figure}[!ht]
    \centering
    \begin{subfigure}{0.49\textwidth}
        \begin{subfigure}{0.49\textwidth}
            \centering
            \includegraphics[width=\linewidth]{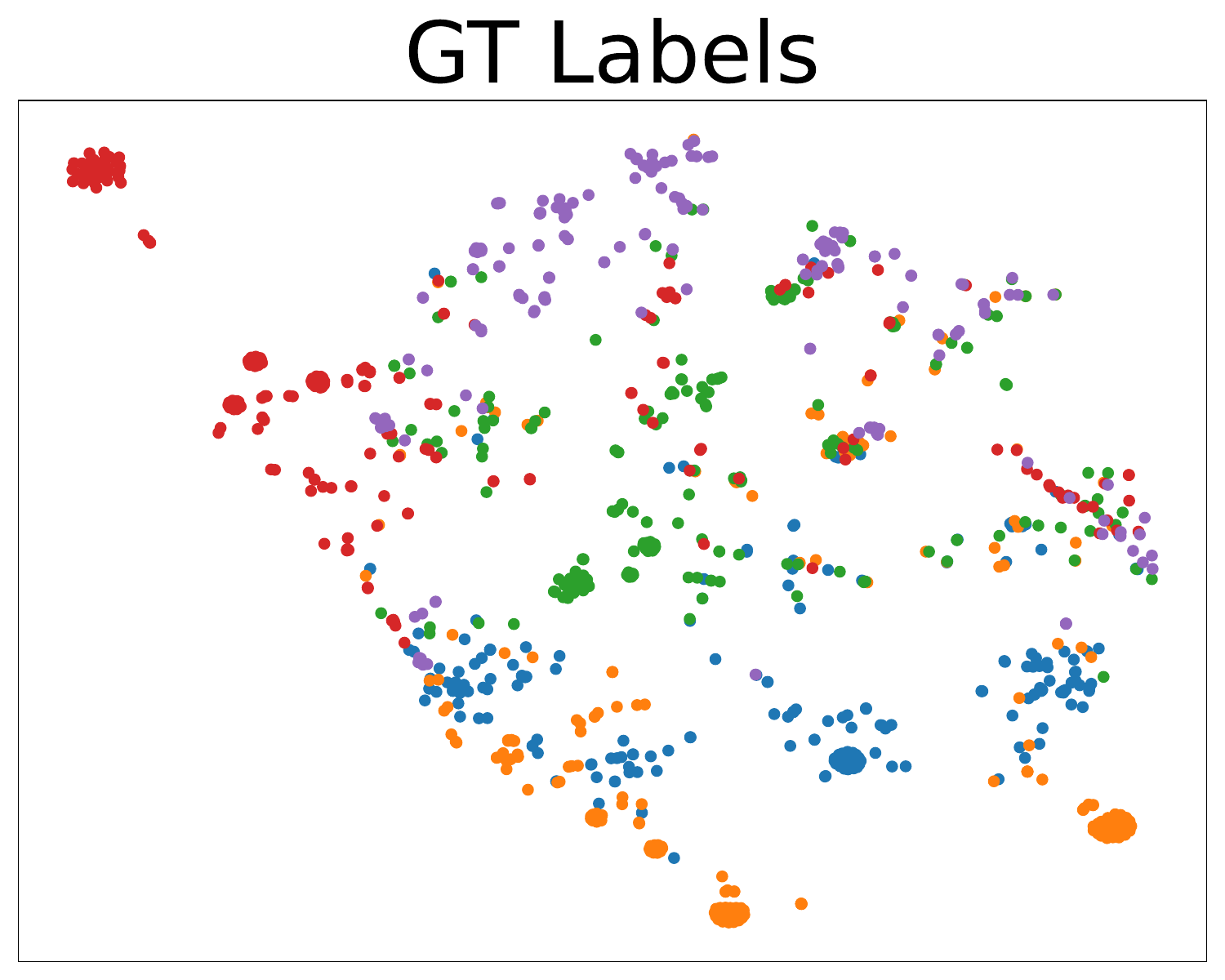}
        \end{subfigure}
        \begin{subfigure}{0.49\textwidth}
            \centering
            \includegraphics[width=\linewidth]{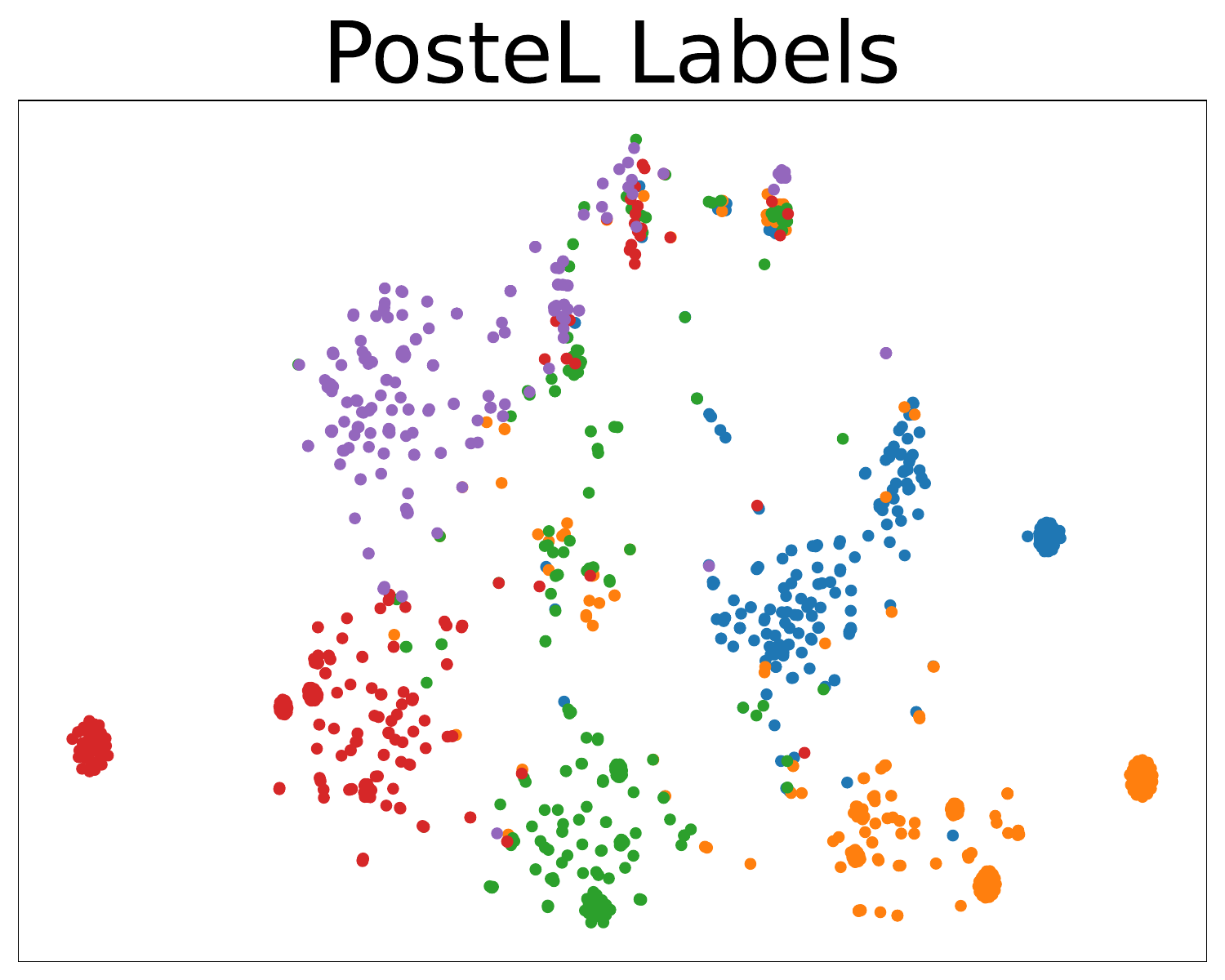}
        \end{subfigure}
        \caption{Chameleon}
    \end{subfigure}
    \begin{subfigure}{0.49\textwidth}
        \begin{subfigure}{0.49\textwidth}
            \centering
            \includegraphics[width=\linewidth]{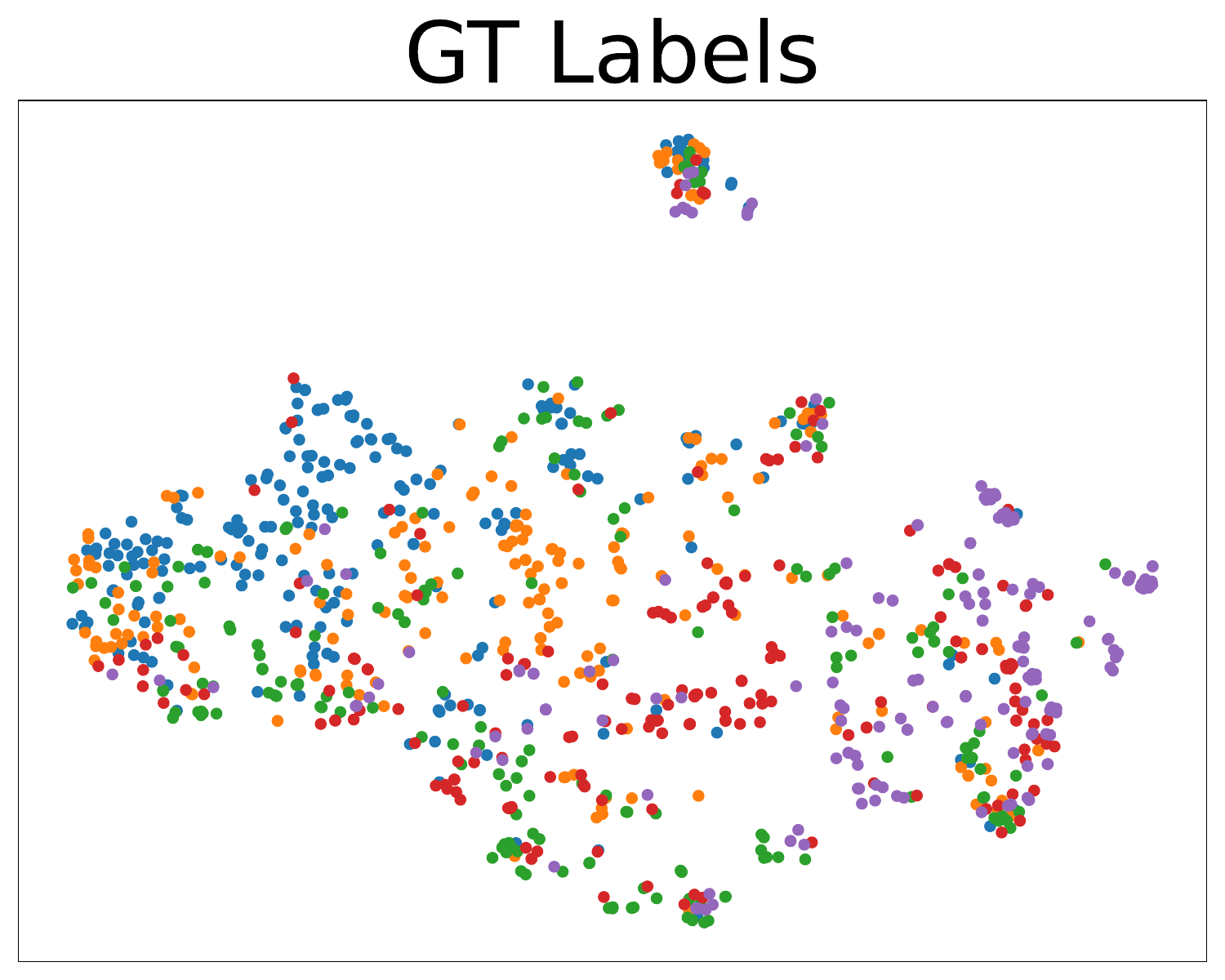}
        \end{subfigure}
        \begin{subfigure}{0.49\textwidth}
            \centering
            \includegraphics[width=\linewidth]
            {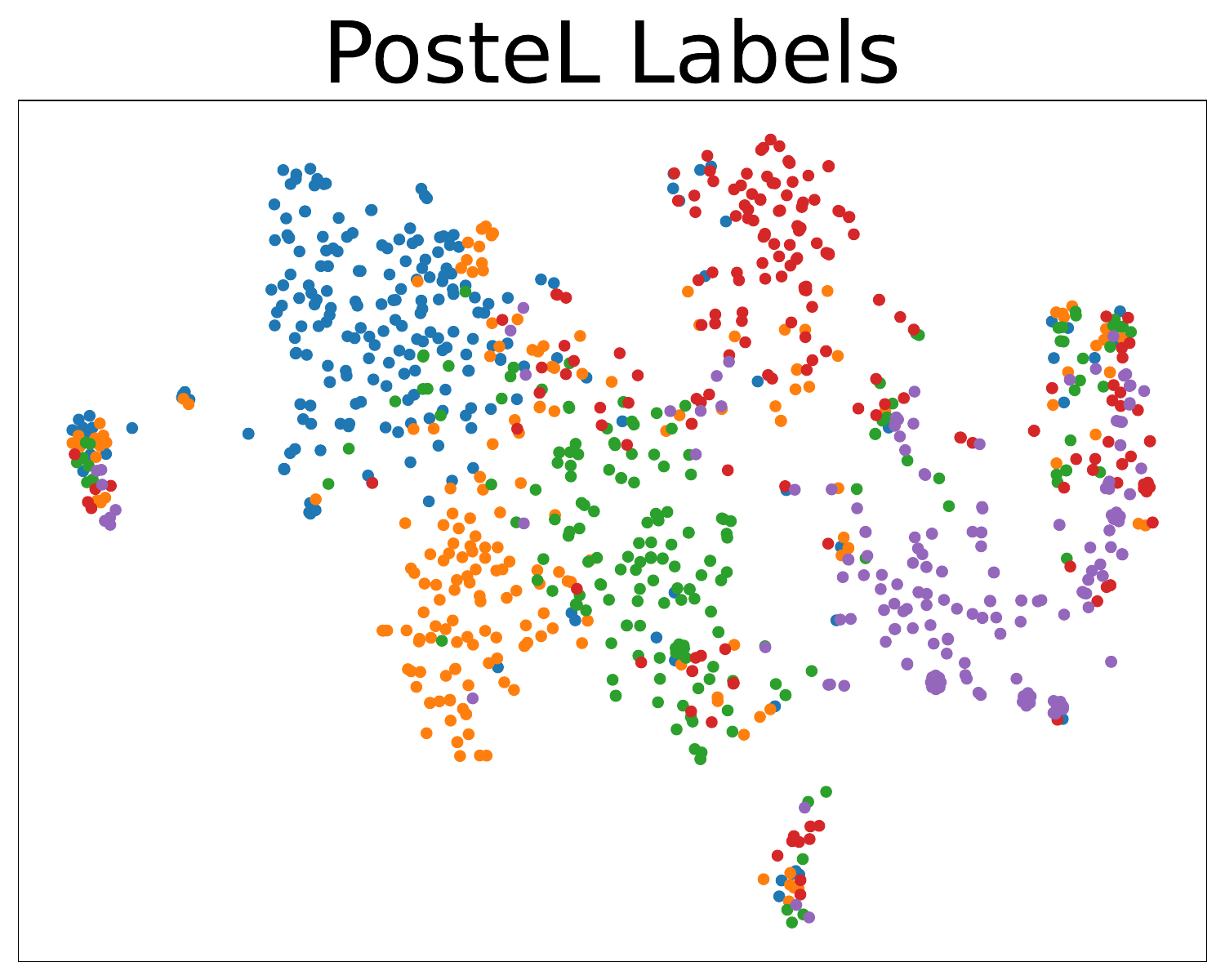}
        \end{subfigure}
        \caption{Squirrel}
    \end{subfigure}
    \caption{t-SNE plots of the final layer representation of the Chameleon and Squirrel datasets. For each dataset, the left figure displays the representations trained on the ground-truth labels, while the right figure displays the representations trained on the \ours{} labels.}
    \label{fig:tsne}
\end{figure}

%% file: figure/main_table.tex
\begin{table*}[ht!]
    \centering
    \resizebox{\linewidth}{!}{
    \begin{tabular}{lcccccccccc} 
         \toprule &\multicolumn{5}{c}{Homophilic} & \multicolumn{5}{c}{Heterophilic} \\
         \cmidrule(lr){2-6}\cmidrule(lr){7-11}
         & Cora & CiteSeer & PubMed & Computers & Photo & Chameleon & Actor & Squirrel & Texas & Cornell \\
         \midrule
         GCN & \tabnum{87.14}{1.01} & \tabnum{79.86}{0.67} & \tabnum{86.74}{0.27} & \tabnum{83.32}{0.33}& \tabnum{88.26}{0.73} & \tabnum{59.61}{2.21} & \tabnum{33.23}{1.16} & \tabnum{46.78}{0.87} & \tabnum{77.38}{3.28}& \tabnum{65.90}{4.43} \\
         +LS & \tabnum{87.77}{0.97} & \tabnum{81.06}{0.59} & \tabnum{87.73}{0.24} & \tabnum{89.08}{0.30} & \tabnum{94.05}{0.26} & \tabnum{64.81}{1.53} & \tabnum{33.81}{0.75} & \tabnum{49.53}{1.10} & \tabnum{77.87}{3.11} & \tabnum{67.87}{3.77} \\
         +KD & \tabnum{87.90}{0.90} & \tabnum{80.97}{0.56} & \tabnum{87.03}{0.29} & \tabnum{88.56}{0.36} & \tabnum{93.64}{0.31} & \tabnum{64.49}{1.38} & \tabnum{33.33}{0.78} & \tabnum{49.38}{0.64} & \tabnum{78.03}{2.62} & \tabnum{63.61}{5.57} \\
         +SALS & \tabnum{88.10}{1.08} & \tabnum{80.52}{0.85} & \tabnum{87.23}{0.13} & \tabnum{88.88}{0.54} & \tabnum{93.80}{0.31} & \tabnum{63.00}{1.75} & \tabnum{33.24}{0.92} & \tabnum{49.16}{0.77} & \tabnum{70.00}{3.93} & \tabnum{58.36}{7.54} \\
         +ALS & \tabnum{88.10}{0.85} & \tabnum{81.02}{0.52} & \tabnum{87.30}{0.30} & \tabnum{89.18}{0.36} & \tabnum{93.88}{0.27} & \tabnum{64.11}{1.29} & \tabnum{34.05}{0.49} & \tabnum{47.44}{0.76} & \tabnum{77.38}{2.13} & \tabnum{71.64}{3.28} \\
         +\ours{} & \besttabnum{88.56}{0.90} & \besttabnum{82.10}{0.50} & \besttabnum{88.00}{0.25} & \besttabnum{89.30}{0.23}& \besttabnum{94.08}{0.35} & \besttabnum{65.80}{1.23} & \besttabnum{35.16}{0.43} & \besttabnum{52.76}{0.64}  & \besttabnum{80.82}{2.79}& \besttabnum{80.33}{1.80} \\
         $\Delta$ & $+1.42(\textcolor{red}{\uparrow})$ & $+2.24(\textcolor{red}{\uparrow})$ & $+1.26(\textcolor{red}{\uparrow})$ & $+5.98(\textcolor{red}{\uparrow})$ & $+5.82(\textcolor{red}{\uparrow})$ & $+6.19(\textcolor{red}{\uparrow})$ & $+1.93(\textcolor{red}{\uparrow})$ & $+5.98(\textcolor{red}{\uparrow})$ & $+3.44(\textcolor{red}{\uparrow})$ & $+14.43(\textcolor{red}{\uparrow})$\\ 
         \midrule
         GAT & \tabnum{88.03}{0.79} & \tabnum{80.52}{0.71} & \tabnum{87.04}{0.24} & \tabnum{83.32}{0.39}& \tabnum{90.94}{0.68} & \tabnum{63.13}{1.93} & \tabnum{33.93}{2.47} & \tabnum{44.49}{0.88} & \besttabnum{80.82}{2.13}& \tabnum{78.21}{2.95}\\
         +LS & \tabnum{88.69}{0.99} & \tabnum{81.27}{0.86} & \tabnum{86.33}{0.32} & \tabnum{88.95}{0.31} & \tabnum{94.06}{0.39} & \tabnum{65.16}{1.49} & \tabnum{34.55}{1.15} & \tabnum{45.94}{1.60} & \tabnum{78.69}{4.10} & \tabnum{74.10}{4.10} \\
         +KD & \tabnum{87.47}{0.94} & \tabnum{80.79}{0.60} & \tabnum{86.54}{0.31} & \tabnum{88.99}{0.46} & \tabnum{93.76}{0.31} & \tabnum{65.14}{1.47} & \tabnum{35.13}{1.36} & \tabnum{43.86}{0.85} & \tabnum{79.02}{2.46} & \tabnum{73.44}{2.46} \\
         +SALS & \tabnum{88.64}{0.94} & \tabnum{81.23}{0.59} & \tabnum{86.49}{0.25} & \tabnum{88.75}{0.36} & \tabnum{93.74}{0.37} & \tabnum{62.76}{1.42} & \tabnum{33.91}{1.41} & \tabnum{42.29}{0.94} & \tabnum{74.92}{4.43} & \tabnum{65.57}{10.00} \\
         +ALS & \tabnum{88.60}{0.92} & \tabnum{81.09}{0.68} & \tabnum{87.06}{0.24} & \tabnum{89.57}{0.35} & \tabnum{94.16}{0.36} & \tabnum{66.15}{1.25} & \tabnum{34.05}{0.52} & \tabnum{46.85}{1.45} & \tabnum{78.03}{3.11} & \tabnum{75.08}{3.77} \\
         +\ours{} & \besttabnum{89.21}{1.08} & \besttabnum{82.13}{0.64} & \besttabnum{87.08}{0.19} & \besttabnum{89.60}{0.29}& \besttabnum{94.31}{0.31} & \besttabnum{66.28}{1.14} & \besttabnum{35.92}{0.72} & \besttabnum{49.38}{1.05} & \tabnum{80.33}{2.62}& \besttabnum{80.33}{1.81} \\
         $\Delta$ & $+1.18(\textcolor{red}{\uparrow})$ & $+1.61(\textcolor{red}{\uparrow})$ & $+0.04(\textcolor{red}{\uparrow})$ & $+6.28(\textcolor{red}{\uparrow})$ & $+3.37(\textcolor{red}{\uparrow})$ & $+3.15(\textcolor{red}{\uparrow})$ & $+1.99(\textcolor{red}{\uparrow})$ & $+4.89(\textcolor{red}{\uparrow})$ & $-0.49(\textcolor{blue}{\downarrow})$ & $+2.12(\textcolor{red}{\uparrow})$\\ 
         \midrule
         APPNP & \tabnum{88.14}{0.73} & \tabnum{80.47}{0.74} & \tabnum{88.12}{0.31} & \tabnum{85.32}{0.37}& \tabnum{88.51}{0.31} & \tabnum{51.84}{1.82} & \tabnum{39.66}{0.55} & \tabnum{34.71}{0.57} & \tabnum{90.98}{1.64}& \tabnum{91.81}{1.96} \\
         +LS & \tabnum{89.01}{0.64} & \tabnum{81.58}{0.61} & \tabnum{88.90}{0.32} & \tabnum{87.28}{0.27} & \tabnum{94.34}{0.23} & \besttabnum{53.98}{1.47} & \tabnum{39.44}{0.78} & \besttabnum{36.81}{0.98} & \tabnum{91.31}{1.48} & \tabnum{89.51}{1.81} \\
         +KD & \tabnum{89.16}{0.74} & \tabnum{81.88}{0.61} & \tabnum{88.04}{0.39} & \tabnum{86.28}{0.44} & \tabnum{93.85}{0.26} & \tabnum{52.17}{1.23} & \besttabnum{41.43}{0.95} & \tabnum{35.28}{1.10} & \tabnum{90.33}{1.64} & \tabnum{91.48}{1.97} \\
         +SALS & \tabnum{88.97}{0.90} & \tabnum{81.53}{0.56} & \tabnum{88.50}{0.31} & \tabnum{86.49}{0.50} & \tabnum{93.74}{0.38} & \tabnum{52.82}{1.95} & \tabnum{39.66}{0.64} & \tabnum{36.34}{0.65} & \tabnum{83.44}{3.93} & \tabnum{89.51}{3.77} \\
         +ALS & \tabnum{88.93}{0.94} & \tabnum{81.75}{0.59} & \besttabnum{89.30}{0.30} & \tabnum{87.32}{0.23} & \tabnum{94.33}{0.24} & \tabnum{53.44}{1.99} & \tabnum{39.89}{0.67} & \tabnum{36.11}{0.81} & \tabnum{90.82}{2.62} & \tabnum{92.13}{1.48} \\
         +\ours{} & \besttabnum{89.62}{0.84} & \besttabnum{82.47}{0.66} & \tabnum{89.17}{0.26} & \besttabnum{87.46}{0.29}& \besttabnum{94.42}{0.24} & \tabnum{53.83}{1.66} & \tabnum{40.18}{0.70} & \tabnum{36.71}{0.60}  & \besttabnum{92.13}{1.48}& \besttabnum{93.44}{1.64} \\
         $\Delta$ & $+1.48(\textcolor{red}{\uparrow})$ & $+2.00(\textcolor{red}{\uparrow})$ & $+1.05(\textcolor{red}{\uparrow})$ & $+2.14(\textcolor{red}{\uparrow})$ & $+5.91(\textcolor{red}{\uparrow})$ & $+1.99(\textcolor{red}{\uparrow})$ & $+0.52(\textcolor{red}{\uparrow})$ & $+2.00(\textcolor{red}{\uparrow})$ & $+1.15(\textcolor{red}{\uparrow})$ & $+1.63(\textcolor{red}{\uparrow})$\\ 
         \midrule
         MLP & \tabnum{76.96}{0.95} & \tabnum{76.58}{0.88} & \tabnum{85.94}{0.22} & \tabnum{82.85}{0.38}& \tabnum{84.72}{0.34} & \tabnum{46.85}{1.51} & \tabnum{40.19}{0.56} & \tabnum{31.03}{1.18} & \tabnum{91.45}{1.14}& \tabnum{90.82}{1.63} \\
         +LS & \tabnum{77.21}{0.97} & \tabnum{76.82}{0.66} & \tabnum{86.14}{0.35} & \tabnum{83.62}{0.88} & \tabnum{89.46}{0.44} & \tabnum{48.23}{1.23} & \tabnum{39.75}{0.63} & \tabnum{31.10}{0.80} & \tabnum{90.98}{1.64} & \tabnum{90.98}{1.31} \\
         +KD & \tabnum{76.32}{0.94} & \tabnum{77.75}{0.75} & \tabnum{85.10}{0.29} & \tabnum{83.89}{0.53} & \tabnum{88.23}{0.38} & \tabnum{47.40}{1.75} & \besttabnum{41.32}{0.75} & \tabnum{32.58}{0.83} & \tabnum{89.34}{1.97} & \tabnum{91.80}{1.15} \\
         +SALS & \tabnum{77.29}{1.05} & \tabnum{77.00}{0.90} & \tabnum{85.78}{0.33} & \tabnum{82.55}{0.51} & \tabnum{89.11}{0.52} & \tabnum{43.68}{1.69} & \tabnum{39.47}{0.73} & \tabnum{30.88}{0.68} & \tabnum{86.39}{5.09} & \tabnum{89.11}{0.52} \\
         +ALS & \tabnum{77.59}{0.69} & \tabnum{77.24}{0.82} & \tabnum{86.43}{0.43} & \besttabnum{84.26}{0.66} & \tabnum{89.86}{0.43} & \tabnum{48.03}{1.38} & \tabnum{39.98}{0.94} & \tabnum{31.33}{0.89} & \tabnum{91.64}{3.44} & \tabnum{91.64}{1.31} \\
         +\ours{} & \besttabnum{78.39}{0.94} & \besttabnum{78.40}{0.71} & \besttabnum{86.51}{0.33} & \tabnum{84.20}{0.55}& \besttabnum{89.90}{0.27} & \besttabnum{48.51}{1.66} & \tabnum{40.15}{0.46} & \besttabnum{33.11}{0.60}  & \besttabnum{92.95}{1.31}& \besttabnum{93.61}{1.80} \\
         $\Delta$ & $+1.43(\textcolor{red}{\uparrow})$ & $+1.82(\textcolor{red}{\uparrow})$ & $+0.57(\textcolor{red}{\uparrow})$ & $+1.35(\textcolor{red}{\uparrow})$ & $+5.18(\textcolor{red}{\uparrow})$ & $+1.66(\textcolor{red}{\uparrow})$ & $-0.04(\textcolor{blue}{\downarrow})$ & $+2.08(\textcolor{red}{\uparrow})$ & $+1.50(\textcolor{red}{\uparrow})$ & $+2.79(\textcolor{red}{\uparrow})$ \\ 
         \midrule
         ChebNet & \tabnum{86.67}{0.82} & \tabnum{79.11}{0.75} & \tabnum{87.95}{0.28} & \tabnum{87.54}{0.43}& \tabnum{93.77}{0.32} & \tabnum{59.28}{1.25} & \tabnum{37.61}{0.89} & \tabnum{40.55}{0.42} & \tabnum{86.22}{2.45}& \tabnum{83.93}{2.13} \\
         +LS & \tabnum{87.22}{0.99} & \tabnum{79.70}{0.63} & \tabnum{88.48}{0.29} & \tabnum{89.55}{0.38} & \tabnum{94.53}{0.37} & \tabnum{66.41}{1.16} & \tabnum{39.39}{0.73} & \tabnum{42.55}{1.11} & \besttabnum{87.21}{2.62} & \tabnum{84.59}{2.30} \\
         +KD & \tabnum{87.36}{0.95} & \tabnum{80.80}{0.72} & \tabnum{88.41}{0.20} & \tabnum{89.81}{0.30} & \tabnum{94.76}{0.30} & \tabnum{61.47}{1.23} & \besttabnum{40.68}{0.50} & \tabnum{43.88}{1.97} & \tabnum{84.75}{3.61} & \tabnum{83.61}{2.30} \\
         +SALS & \tabnum{87.31}{0.94} & \tabnum{79.71}{0.83} & \tabnum{88.46}{0.30} & \tabnum{89.52}{0.35} & \tabnum{94.19}{0.27} & \tabnum{56.94}{2.52} & \tabnum{39.25}{0.67} & \tabnum{41.61}{0.93} & \tabnum{74.26}{3.61} & \tabnum{73.44}{6.89} \\
         +ALS & \tabnum{87.39}{0.97} & \tabnum{79.81}{0.81} & \tabnum{88.80}{0.33} & \tabnum{89.88}{0.36} & \besttabnum{95.21}{0.23} & \tabnum{61.09}{0.63} & \tabnum{39.61}{1.12} & \tabnum{41.98}{0.85} & \tabnum{85.57}{3.28} & \tabnum{86.39}{2.30} \\
         +\ours{} & \besttabnum{88.57}{0.92} & \besttabnum{82.48}{0.52} & \besttabnum{89.20}{0.31} & \besttabnum{89.95}{0.40}& \tabnum{94.87}{0.25} & \besttabnum{66.83}{0.77} & \tabnum{39.56}{0.51} & \besttabnum{50.87}{0.90}  & \tabnum{86.39}{2.46}& \besttabnum{88.52}{2.63} \\
         $\Delta$ & $+1.90(\textcolor{red}{\uparrow})$ & $+3.37(\textcolor{red}{\uparrow})$ & $+1.25(\textcolor{red}{\uparrow})$ & $+2.41(\textcolor{red}{\uparrow})$ & $+1.10(\textcolor{red}{\uparrow})$ & $+7.55(\textcolor{red}{\uparrow})$ & $+1.95(\textcolor{red}{\uparrow})$ & $+10.32(\textcolor{red}{\uparrow})$ & $+0.17(\textcolor{red}{\uparrow})$ & $+4.59(\textcolor{red}{\uparrow})$\\ 
         \midrule
         GPR-GNN & \tabnum{88.57}{0.69} & \tabnum{80.12}{0.83} & \tabnum{88.46}{0.33} & \tabnum{86.85}{0.25}& \tabnum{93.85}{0.28} & \tabnum{67.28}{1.09} & \tabnum{39.92}{0.67} & \tabnum{50.15}{1.92} & \tabnum{92.95}{1.31}& \tabnum{91.37}{1.81} \\
         +LS & \tabnum{88.82}{0.99} & \tabnum{79.78}{1.06} & \tabnum{88.24}{0.42} & \tabnum{88.39}{0.48} & \tabnum{93.97}{0.33} & \tabnum{67.90}{1.01} & \tabnum{39.72}{0.70} & \tabnum{53.39}{1.80} & \tabnum{92.79}{1.15} & \tabnum{90.49}{2.46} \\
         +KD & \besttabnum{89.33}{1.03} & \besttabnum{81.24}{0.85} & \tabnum{89.85}{0.56} & \tabnum{87.88}{1.11} & \tabnum{94.23}{0.51} & \tabnum{66.76}{1.31} & \besttabnum{42.00}{0.63} & \tabnum{53.26}{1.07} & \besttabnum{94.26}{1.48} & \tabnum{88.52}{1.97} \\
         +SALS & \tabnum{88.78}{0.90} & \tabnum{80.71}{0.91} & \tabnum{90.12}{0.46} & \tabnum{88.63}{0.35} & \tabnum{94.23}{0.65} & \tabnum{65.16}{1.49} & \tabnum{39.67}{0.73} & \tabnum{44.75}{1.45} & \tabnum{73.61}{3.44} & \tabnum{82.46}{2.95} \\
         +ALS & \tabnum{88.93}{1.31} & \tabnum{80.31}{0.71} & \tabnum{90.23}{0.50} & \tabnum{89.14}{0.48} & \tabnum{94.55}{0.53} & \tabnum{67.79}{1.07} & \tabnum{40.09}{0.72} & \tabnum{51.34}{1.00} & \tabnum{92.95}{1.31} & \tabnum{89.18}{2.13} \\
         +\ours{} & \tabnum{89.20}{1.07} & \tabnum{81.21}{0.64} & \besttabnum{90.57}{0.31} & \besttabnum{89.84}{0.43}& \besttabnum{94.76}{0.38} & \besttabnum{68.38}{1.12} & \tabnum{40.08}{0.69} & \besttabnum{53.54}{0.79}  & \tabnum{93.28}{1.31}& \besttabnum{92.46}{0.99} \\
         $\Delta$ & $+0.63(\textcolor{red}{\uparrow})$ & $+1.09(\textcolor{red}{\uparrow})$ & $+2.11(\textcolor{red}{\uparrow})$ & $+2.99(\textcolor{red}{\uparrow})$ & $+0.91(\textcolor{red}{\uparrow})$ & $+1.10(\textcolor{red}{\uparrow})$ & $+0.16(\textcolor{red}{\uparrow})$ & $+3.39(\textcolor{red}{\uparrow})$ & $+0.33(\textcolor{red}{\uparrow})$ & $+1.09(\textcolor{red}{\uparrow})$\\ 
         \midrule
         BernNet & \tabnum{88.52}{0.95} & \tabnum{80.09}{0.79} & \tabnum{88.48}{0.41} & \tabnum{87.64}{0.44}& \tabnum{93.63}{0.35} & \tabnum{68.29}{1.58} & \besttabnum{41.79}{1.01} & \tabnum{51.35}{0.73} & \tabnum{93.12}{0.65}& \tabnum{92.13}{1.64} \\
         +LS & \tabnum{88.80}{0.92} & \tabnum{80.37}{1.05} & \tabnum{87.40}{0.27} & \tabnum{88.32}{0.38} & \tabnum{93.70}{0.21} & \tabnum{69.58}{0.94} & \tabnum{39.60}{0.53} & \tabnum{52.39}{0.60} & \tabnum{91.80}{1.80} & \tabnum{90.49}{1.48} \\
         +KD & \tabnum{87.78}{0.99} & \tabnum{81.20}{0.86} & \tabnum{87.59}{0.41} & \tabnum{87.35}{0.40} & \tabnum{93.96}{0.40} & \tabnum{67.75}{1.42} & \tabnum{41.04}{0.89} & \tabnum{51.25}{0.83} & \tabnum{93.61}{1.31} & \tabnum{90.33}{2.30} \\
         +SALS & \tabnum{88.77}{0.85} & \tabnum{81.20}{0.61} & \tabnum{88.61}{0.35} & \tabnum{88.87}{0.33} & \tabnum{94.22}{0.43} & \tabnum{64.62}{0.85} & \tabnum{40.15}{1.07} & \tabnum{46.19}{0.78} & \tabnum{85.90}{4.10} & \tabnum{88.03}{3.12} \\
         +ALS & \tabnum{89.13}{0.79} & \tabnum{81.17}{0.67} & \besttabnum{89.19}{0.46} & \tabnum{89.52}{0.30} & \besttabnum{94.54}{0.32} & \tabnum{67.92}{1.07} & \tabnum{40.51}{0.61} & \tabnum{51.83}{1.31} & \tabnum{93.77}{1.31} & \tabnum{92.79}{1.48} \\
         +\ours{} & \besttabnum{89.39}{0.92} & \besttabnum{82.46}{0.67} & \tabnum{89.07}{0.29} & \besttabnum{89.56}{0.35}& \besttabnum{94.54}{0.36} & \besttabnum{69.65}{0.83} & \tabnum{40.40}{0.67} & \besttabnum{53.11}{0.87} & \besttabnum{93.93}{1.15}& \besttabnum{92.95}{1.80} \\
         $\Delta$ & $+0.87(\textcolor{red}{\uparrow})$ & $+2.37(\textcolor{red}{\uparrow})$ & $+0.59(\textcolor{red}{\uparrow})$ & $+1.92(\textcolor{red}{\uparrow})$ & $+0.91(\textcolor{red}{\uparrow})$ & $+1.36(\textcolor{red}{\uparrow})$ & $-1.39(\textcolor{blue}{\downarrow})$ & $+1.76(\textcolor{red}{\uparrow})$ & $+0.81(\textcolor{red}{\uparrow})$ & $+0.82(\textcolor{red}{\uparrow})$\\ 
         \midrule
         OrderedGNN &  \tabnum{88.62}{1.05} & \tabnum{80.11}{0.86} & \tabnum{88.74}{0.56} & \tabnum{89.72}{0.50} & \tabnum{94.76}{0.36}  & \tabnum{58.27}{1.33} & \tabnum{39.73}{1.15} & \tabnum{38.70}{1.10} & \tabnum{90.16}{2.63} & \tabnum{90.33}{2.46}\\
         +LS & \tabnum{88.52}{0.94} & \tabnum{80.23}{0.80} & \tabnum{88.16}{0.33} & \tabnum{89.59}{0.47} & \tabnum{94.49}{0.45}  & \tabnum{58.86}{1.62} & \tabnum{40.01}{0.66} & \tabnum{40.12}{0.82} & \tabnum{88.20}{3.61} & \tabnum{91.15}{1.31}\\
         +KD & \tabnum{88.26}{1.07} & \tabnum{80.52}{0.83} & \tabnum{88.23}{0.21} & \tabnum{89.35}{0.34} & \tabnum{94.40}{0.23} & \tabnum{58.21}{1.18} & \tabnum{40.17}{0.45} & \tabnum{40.92}{0.87} & \besttabnum{90.49}{1.48} & \tabnum{91.31}{1.80}\\
         +SALS & \tabnum{88.44}{0.97} & \tabnum{80.93}{0.72} & \tabnum{88.08}{0.62} & \tabnum{88.94}{0.51} & \tabnum{93.87}{0.35}  & \tabnum{59.30}{1.25} & \tabnum{39.52}{0.41} & \tabnum{40.85}{0.86} & \tabnum{77.70}{4.75} & \tabnum{84.75}{4.10}\\
         +ALS &\tabnum{87.96}{0.74} & \tabnum{80.60}{0.57} & \tabnum{88.69}{0.57} & \tabnum{89.84}{0.48} & \tabnum{94.76}{0.36}  & \tabnum{59.39}{1.23} & \besttabnum{40.28}{0.79} & \tabnum{40.37}{1.05} & \tabnum{90.00}{2.62} & \tabnum{89.84}{2.95}\\
         +\ours{} & \besttabnum{88.97}{1.15} & \besttabnum{82.54}{0.64} & \besttabnum{88.85}{0.61} & \besttabnum{90.13}{0.29} & \besttabnum{94.96}{0.34}  & \besttabnum{60.15}{1.20} & \tabnum{39.99}{1.00} & \besttabnum{43.72}{0.85} & \tabnum{87.70}{5.25} & \besttabnum{91.97}{1.15}\\
         $\Delta$ & $+0.35(\textcolor{red}{\uparrow})$ & $+2.43(\textcolor{red}{\uparrow})$ & $+0.11(\textcolor{red}{\uparrow})$ & $+0.41(\textcolor{red}{\uparrow})$ & $+0.20(\textcolor{red}{\uparrow})$ & $+1.88(\textcolor{red}{\uparrow})$ & $+0.26(\textcolor{red}{\uparrow})$ & $+5.02(\textcolor{red}{\uparrow})$ & $-2.46(\textcolor{blue}{\downarrow})$ & $+1.64(\textcolor{red}{\uparrow})$\\ 
         \bottomrule
    \end{tabular}
    }
    \caption{Classification accuracy on 10 node classification datasets. $\Delta$ represents the performance improvement achieved by \ours{} compared to the backbone model trained with the ground-truth label. All results of the backbone model trained with the ground-truth label are sourced from \citet{he2021bernnet}.}
    \label{tab:main}
\end{table*}

%% file: figure/figure_appendix_baselinewise_loss_curve_homo.tex
\begin{figure*}[ht!]
    \centering
    \begin{subfigure}{.9\textwidth}
    \centering
    \begin{subfigure}{0.28\textwidth}
        \centering
        \includegraphics[width=\linewidth]{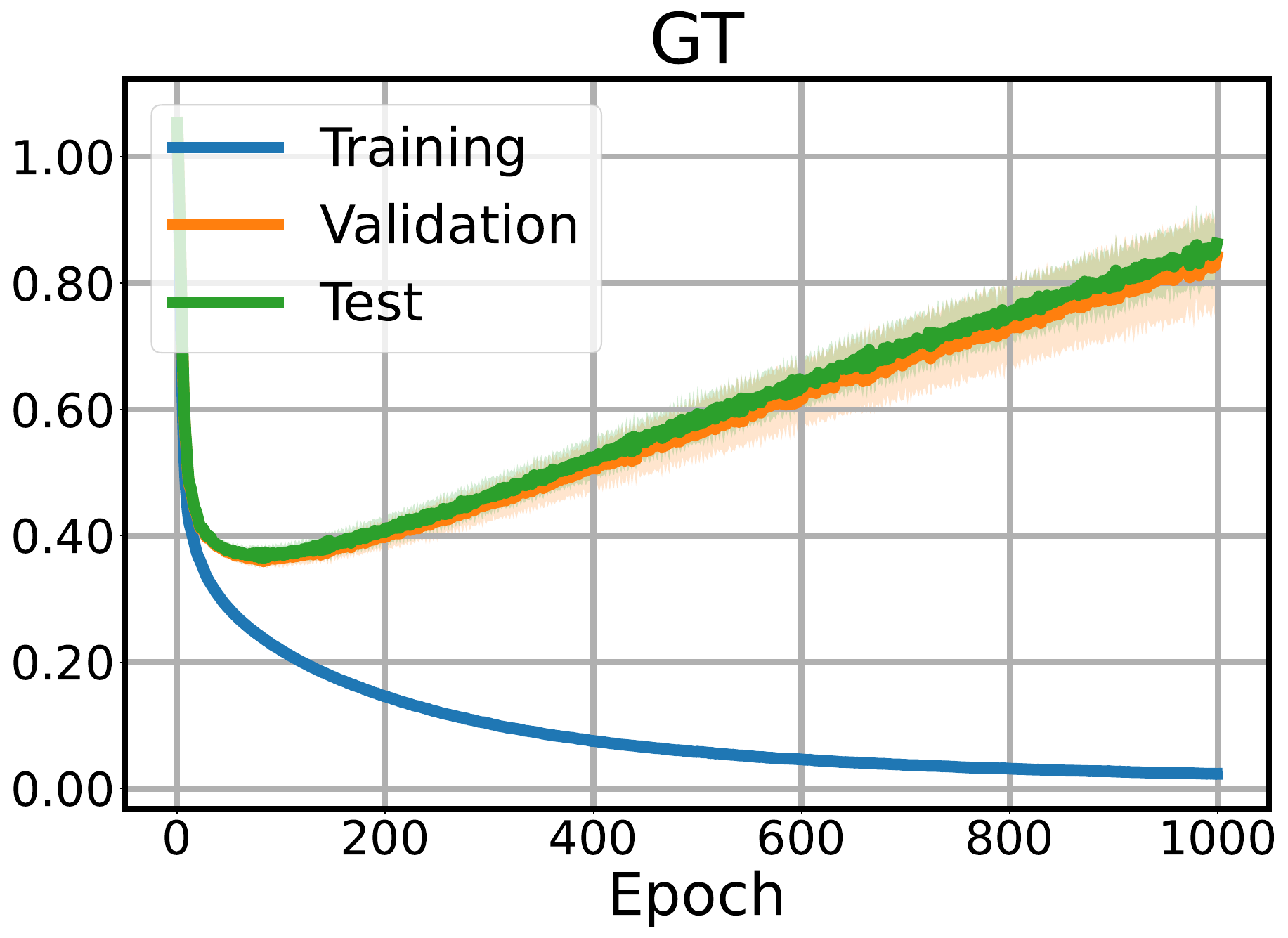}
    \end{subfigure}
    \hspace{-1mm}
    \begin{subfigure}{0.28\textwidth}
        \centering
        \includegraphics[width=\linewidth]{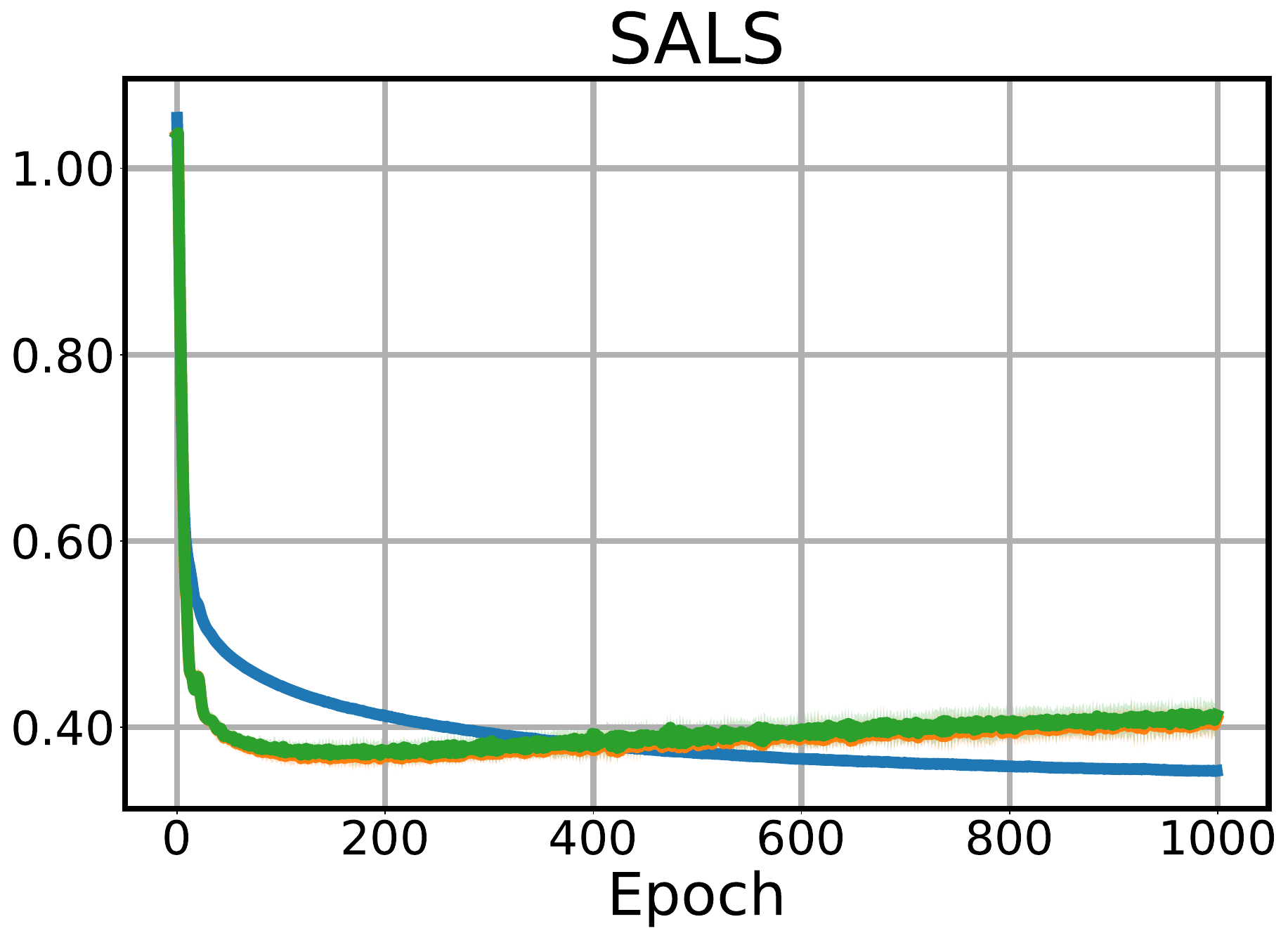}
    \end{subfigure}
    \hspace{-1mm}
    \begin{subfigure}{0.28\textwidth}
        \centering
        \includegraphics[width=\linewidth]{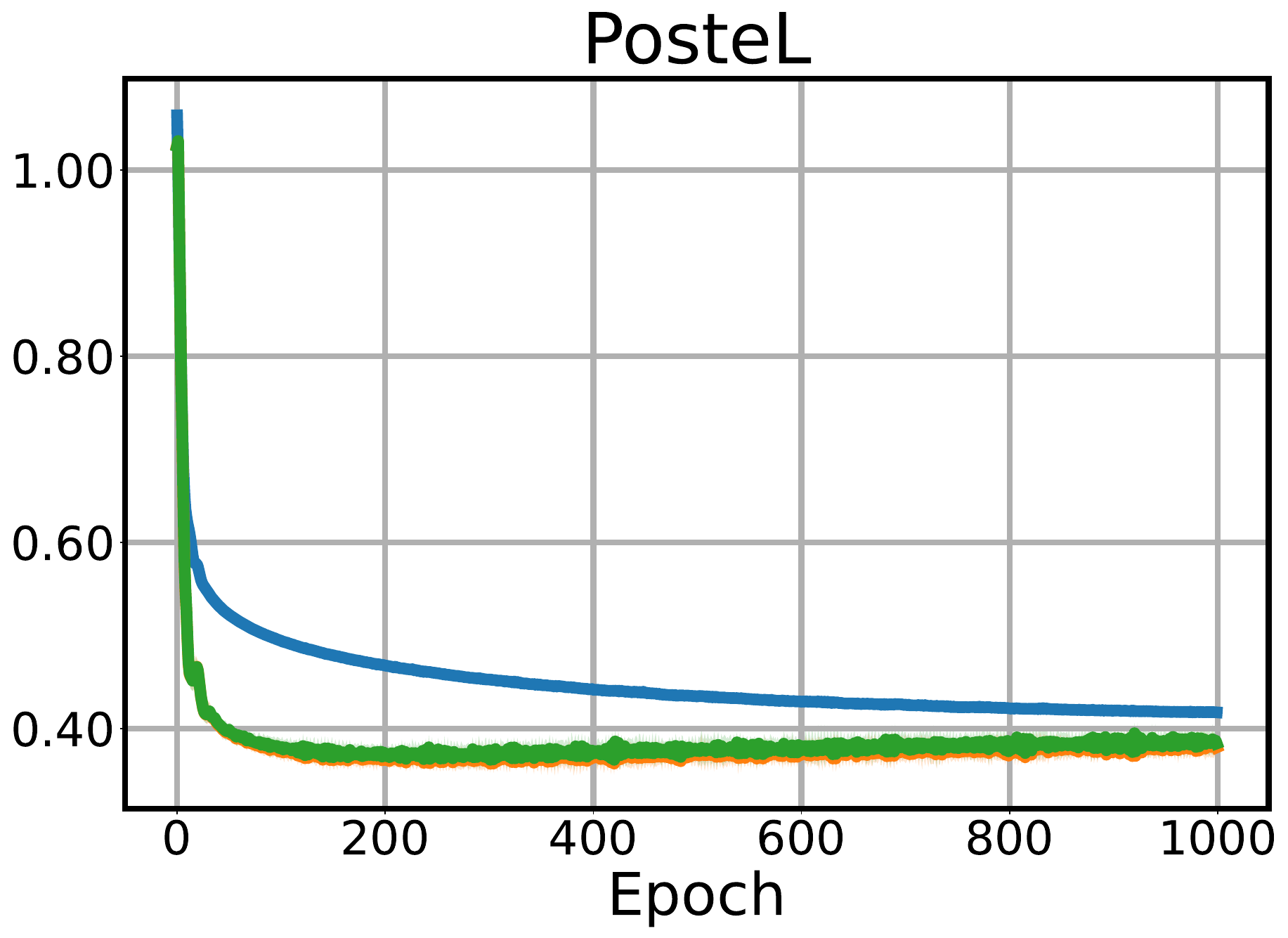}
    \end{subfigure}
    \caption{PubMed}
    \end{subfigure}

    \begin{subfigure}{.9\textwidth}
    \centering
    \begin{subfigure}{0.28\textwidth}
        \centering
        \includegraphics[width=\linewidth]{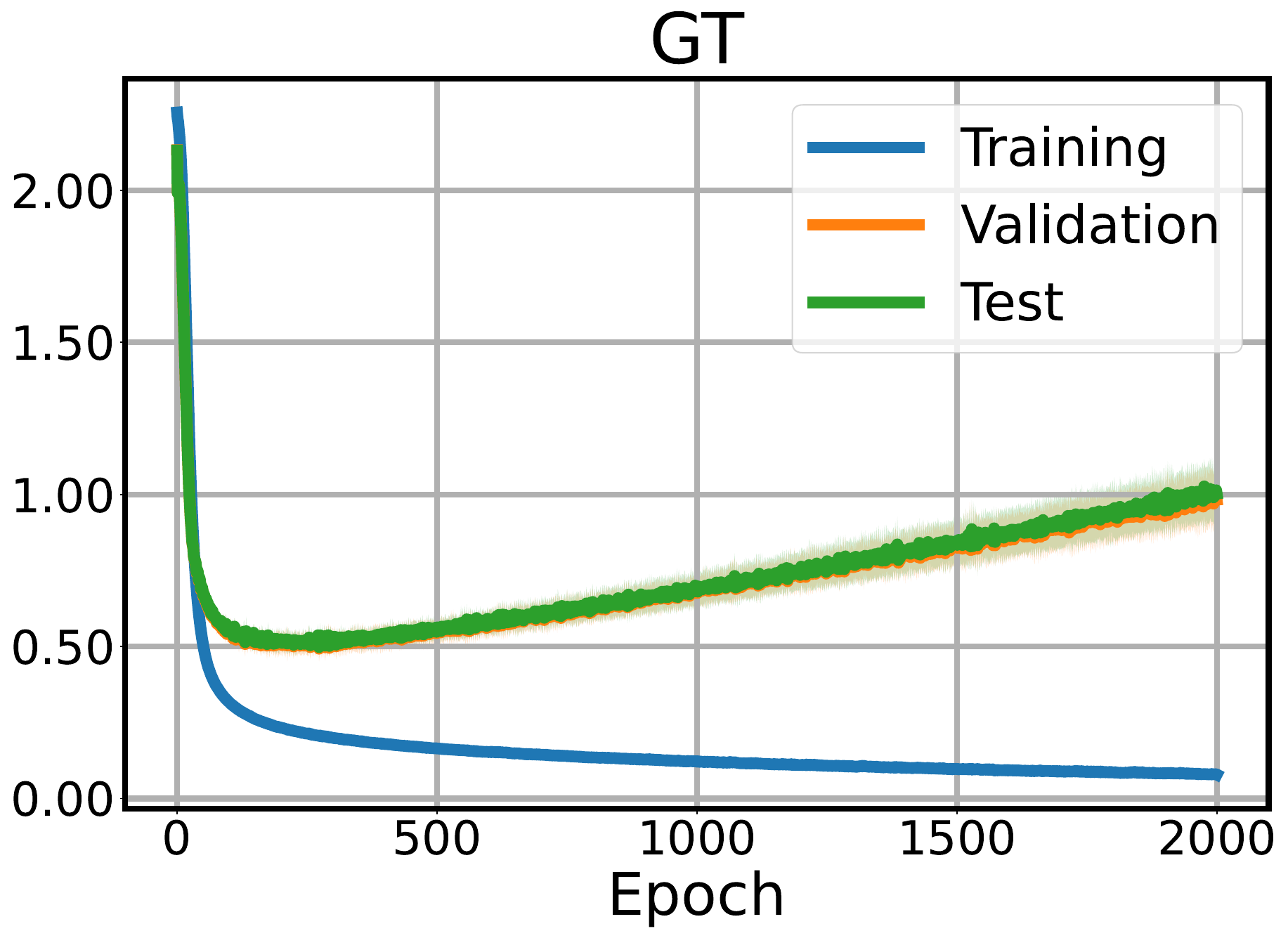}
    \end{subfigure}
    \hspace{-1mm}
    \begin{subfigure}{0.28\textwidth}
        \centering
        \includegraphics[width=\linewidth]{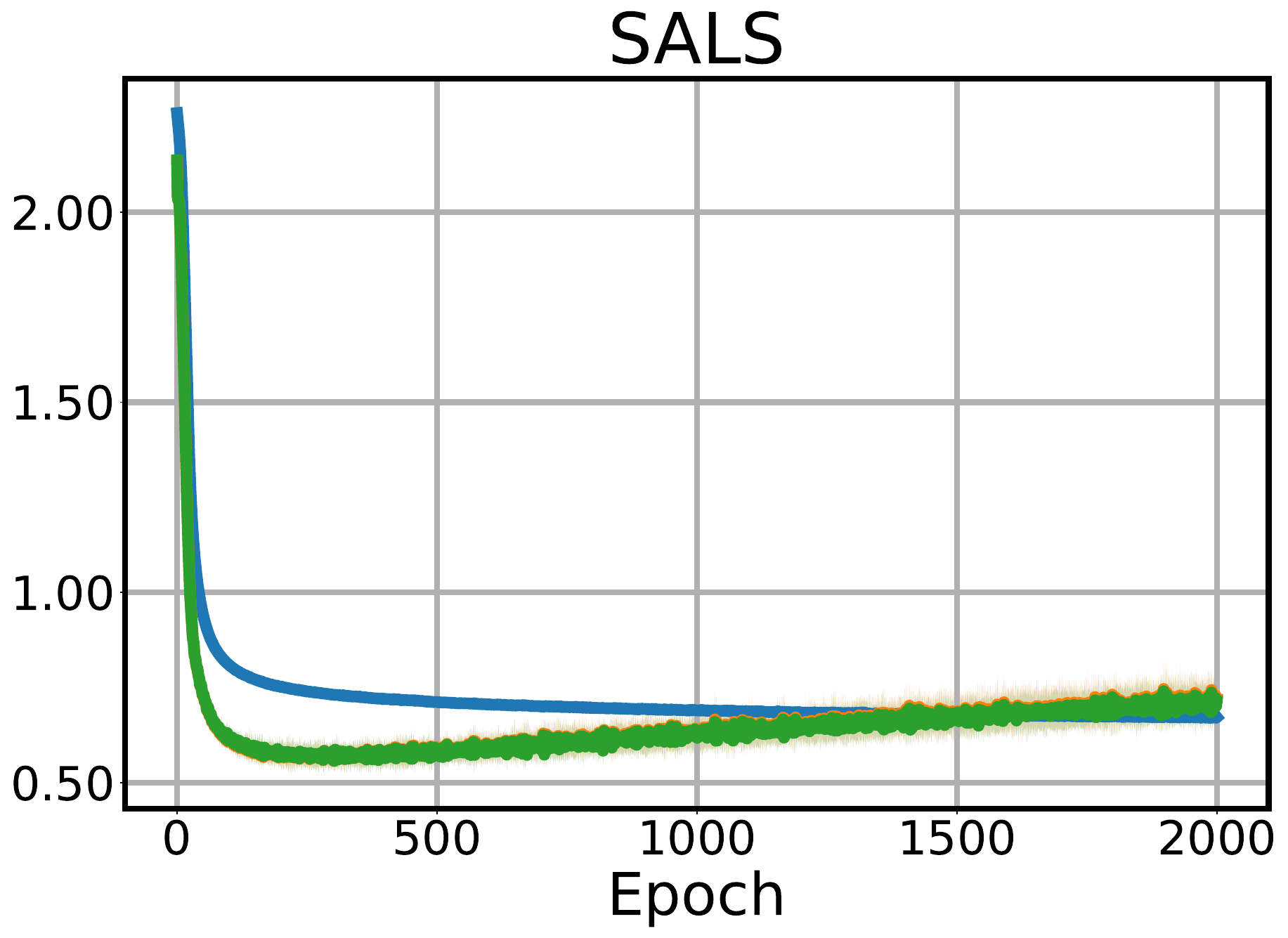}
    \end{subfigure}
    \hspace{-1mm}
    \begin{subfigure}{0.28\textwidth}
        \centering
        \includegraphics[width=\linewidth]{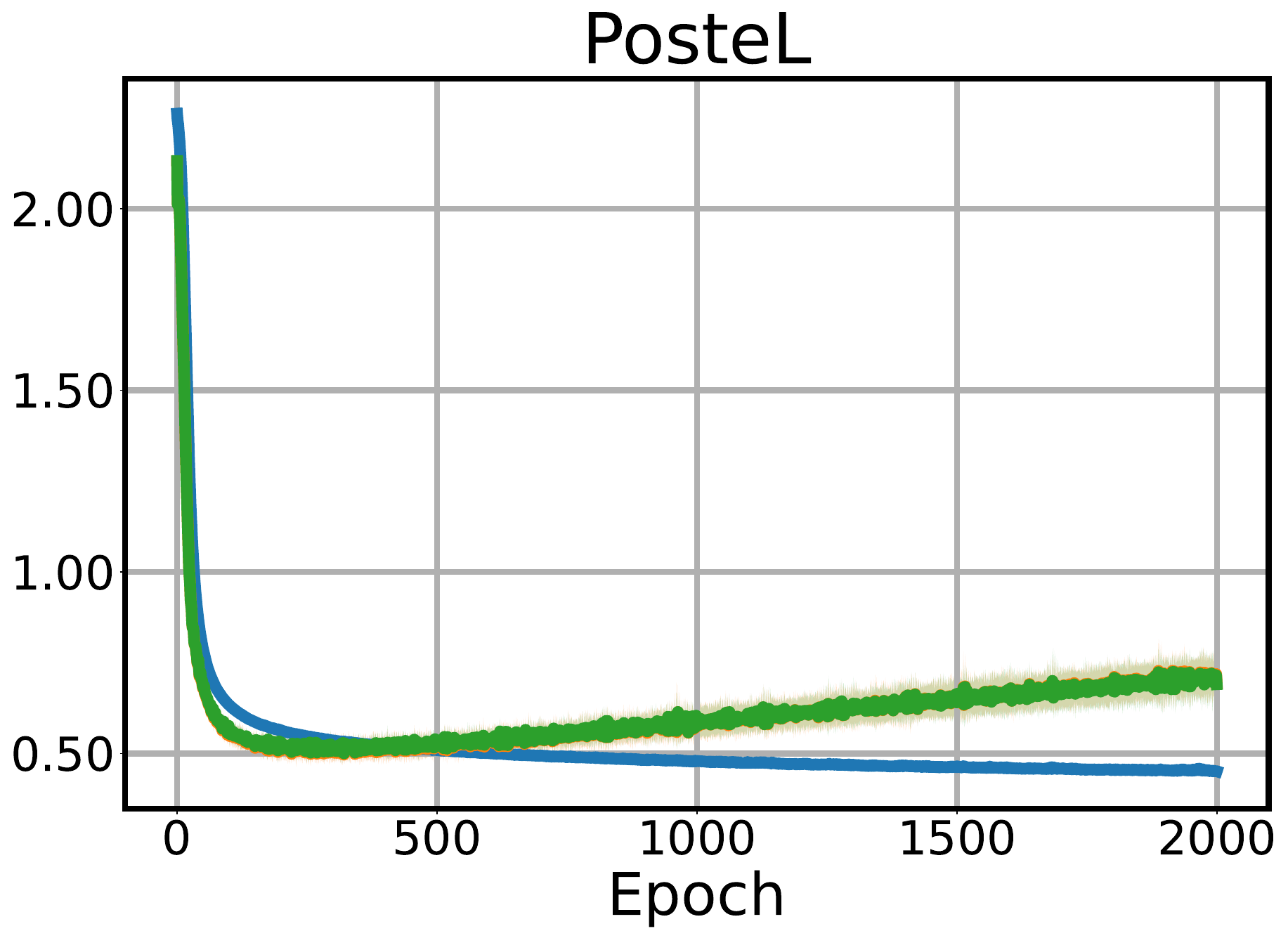}
    \end{subfigure}
    \caption{Computers}
    \end{subfigure}

    \begin{subfigure}{.9\textwidth}
    \centering
    \begin{subfigure}{0.28\textwidth}
        \centering
        \includegraphics[width=\linewidth]{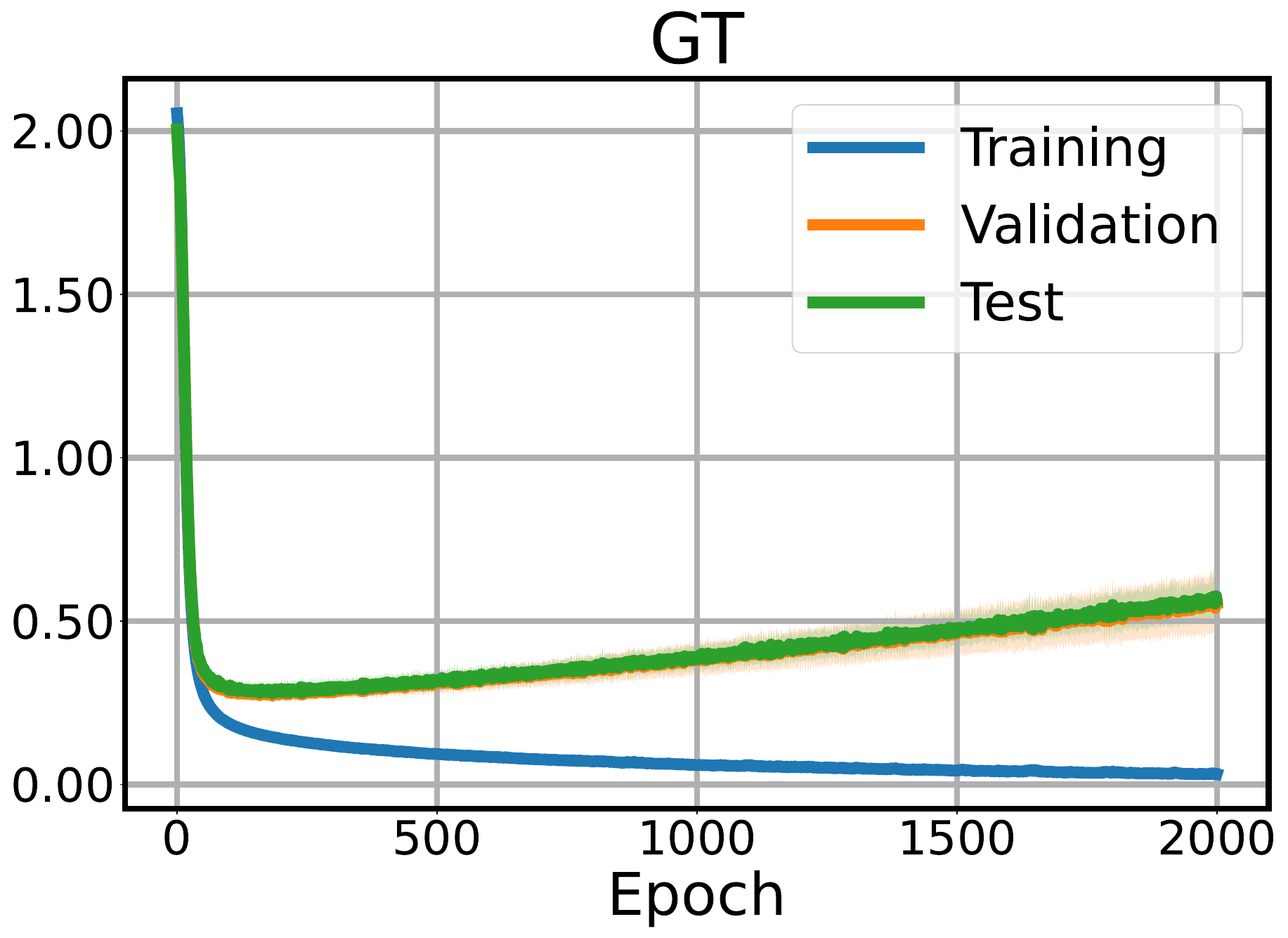}
    \end{subfigure}
    \hspace{-1mm}
    \begin{subfigure}{0.28\textwidth}
        \centering
        \includegraphics[width=\linewidth]{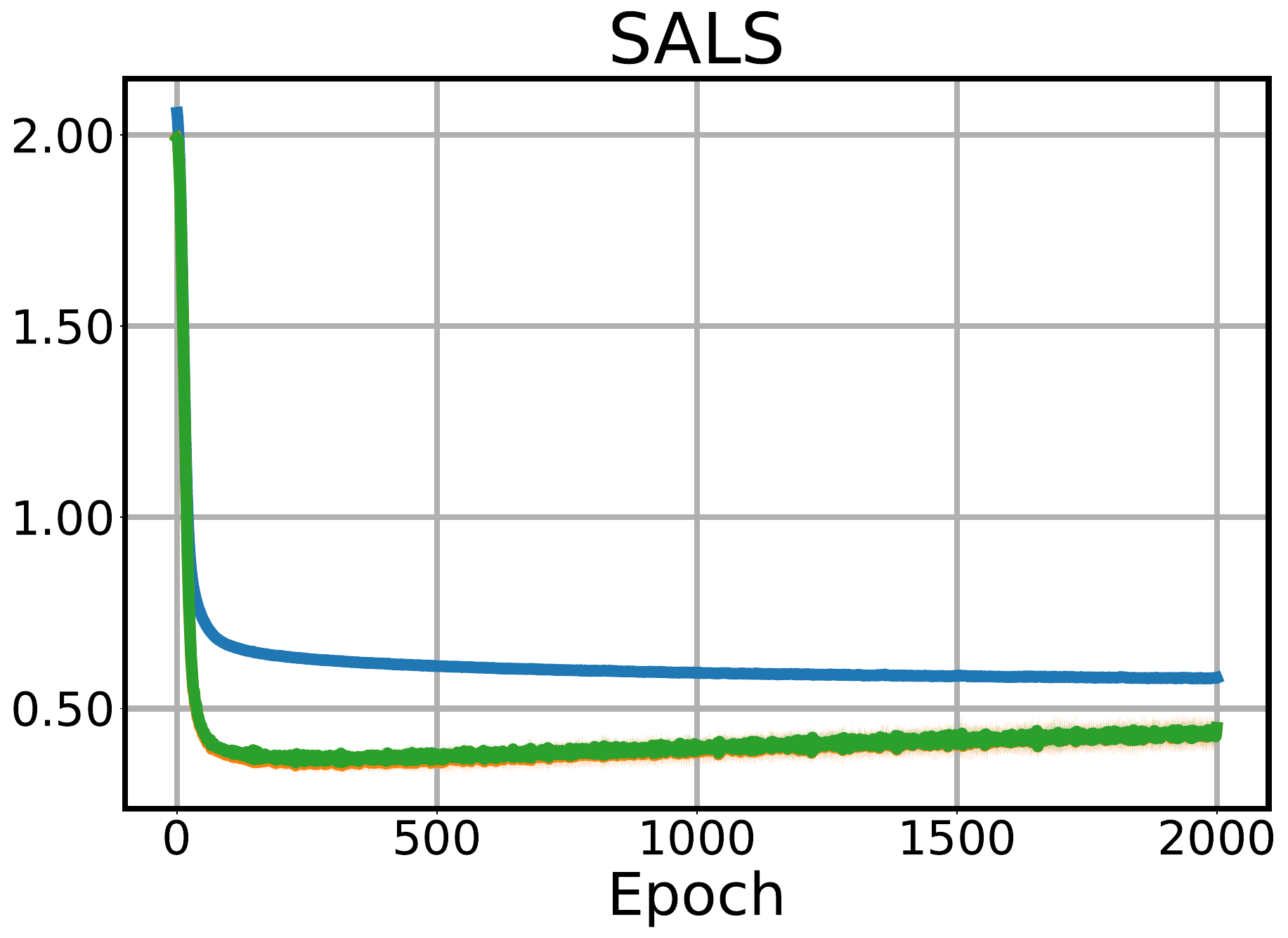}
    \end{subfigure}
    \hspace{-1mm}
    \begin{subfigure}{0.28\textwidth}
        \centering
        \includegraphics[width=\linewidth]{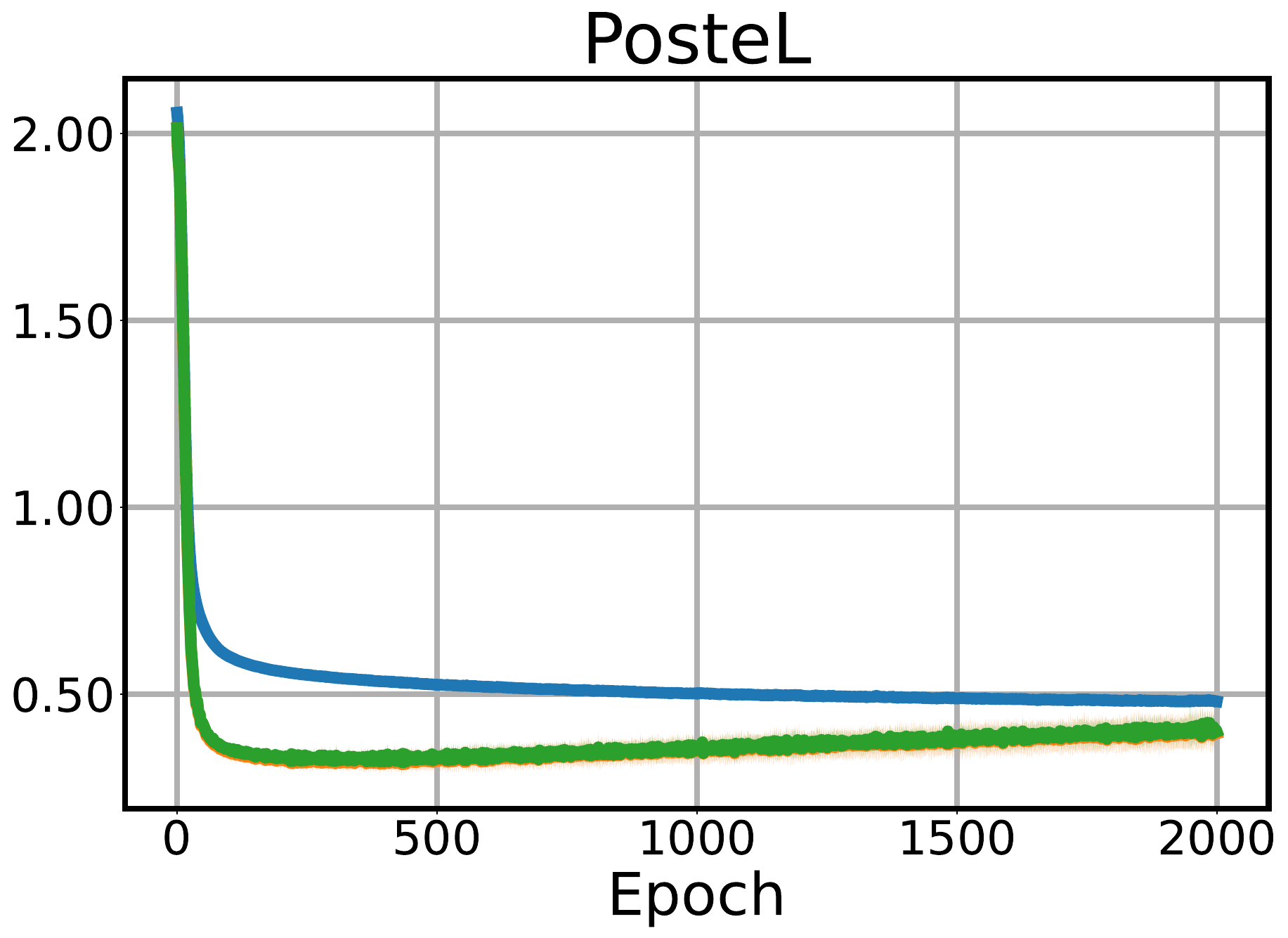}
    \end{subfigure}
    \caption{Photo}
    \end{subfigure}

    \begin{subfigure}{.9\textwidth}
    \centering
    \begin{subfigure}{0.28\textwidth}
        \centering
        \includegraphics[width=\linewidth]{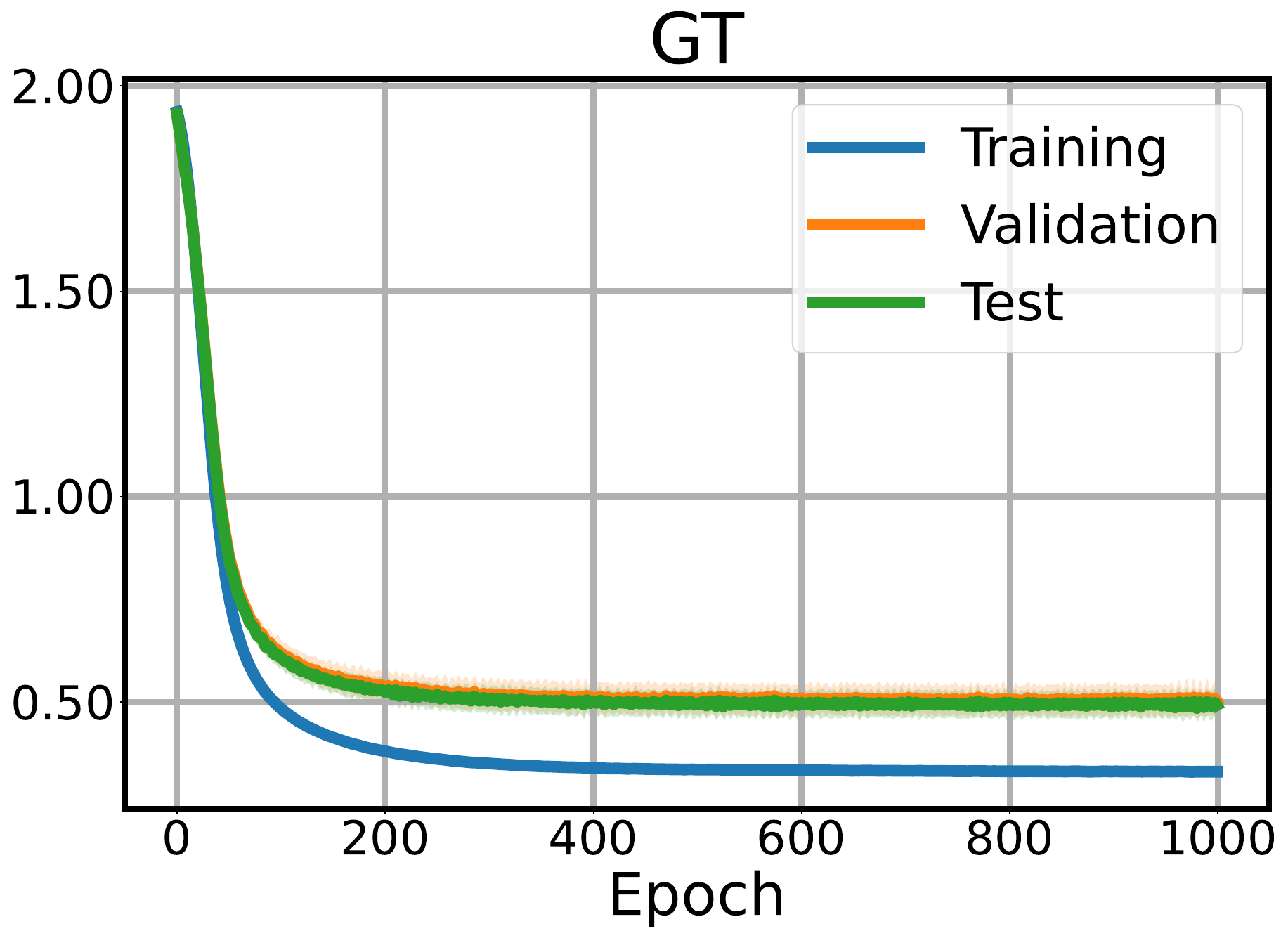}
    \end{subfigure}
    \hspace{-1mm}
    \begin{subfigure}{0.28\textwidth}
        \centering
        \includegraphics[width=\linewidth]{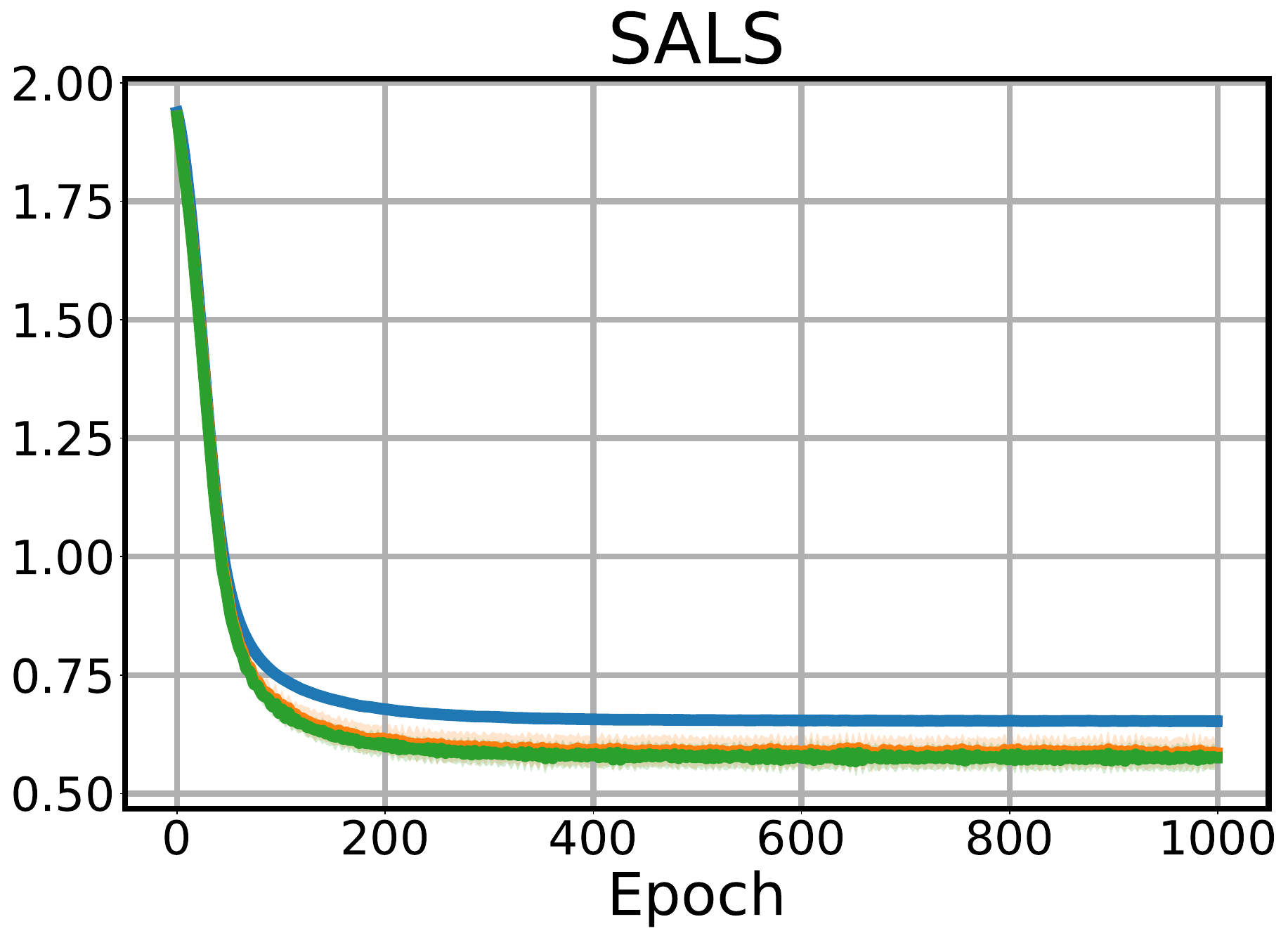}
    \end{subfigure}
    \hspace{-1mm}
    \begin{subfigure}{0.28\textwidth}
        \centering
        \includegraphics[width=\linewidth]{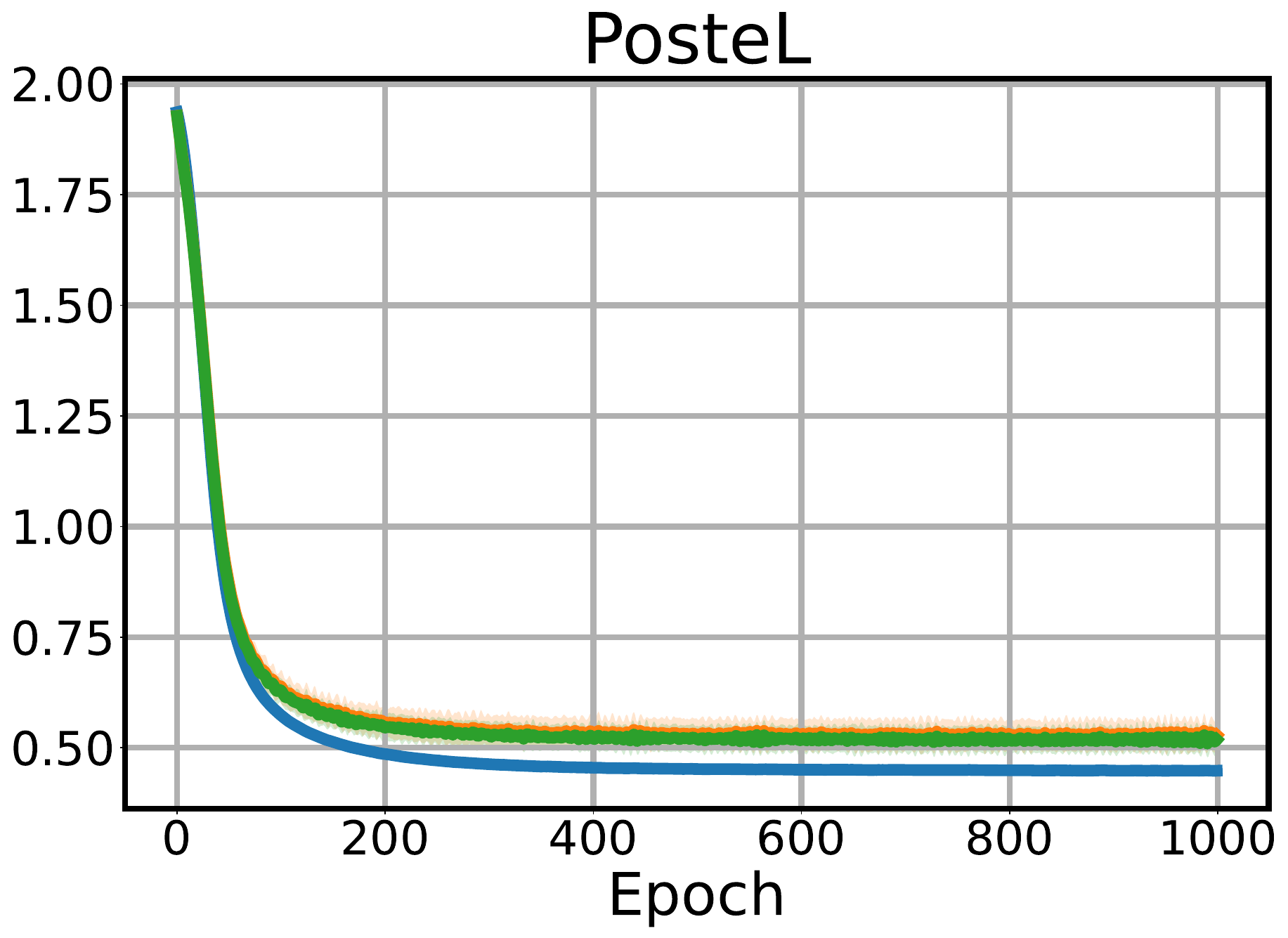}
    \end{subfigure}
    \caption{Cora}
    \end{subfigure}
    
    \begin{subfigure}{.9\textwidth}
    \centering
    \begin{subfigure}{0.28\textwidth}
        \centering
        \includegraphics[width=\linewidth]{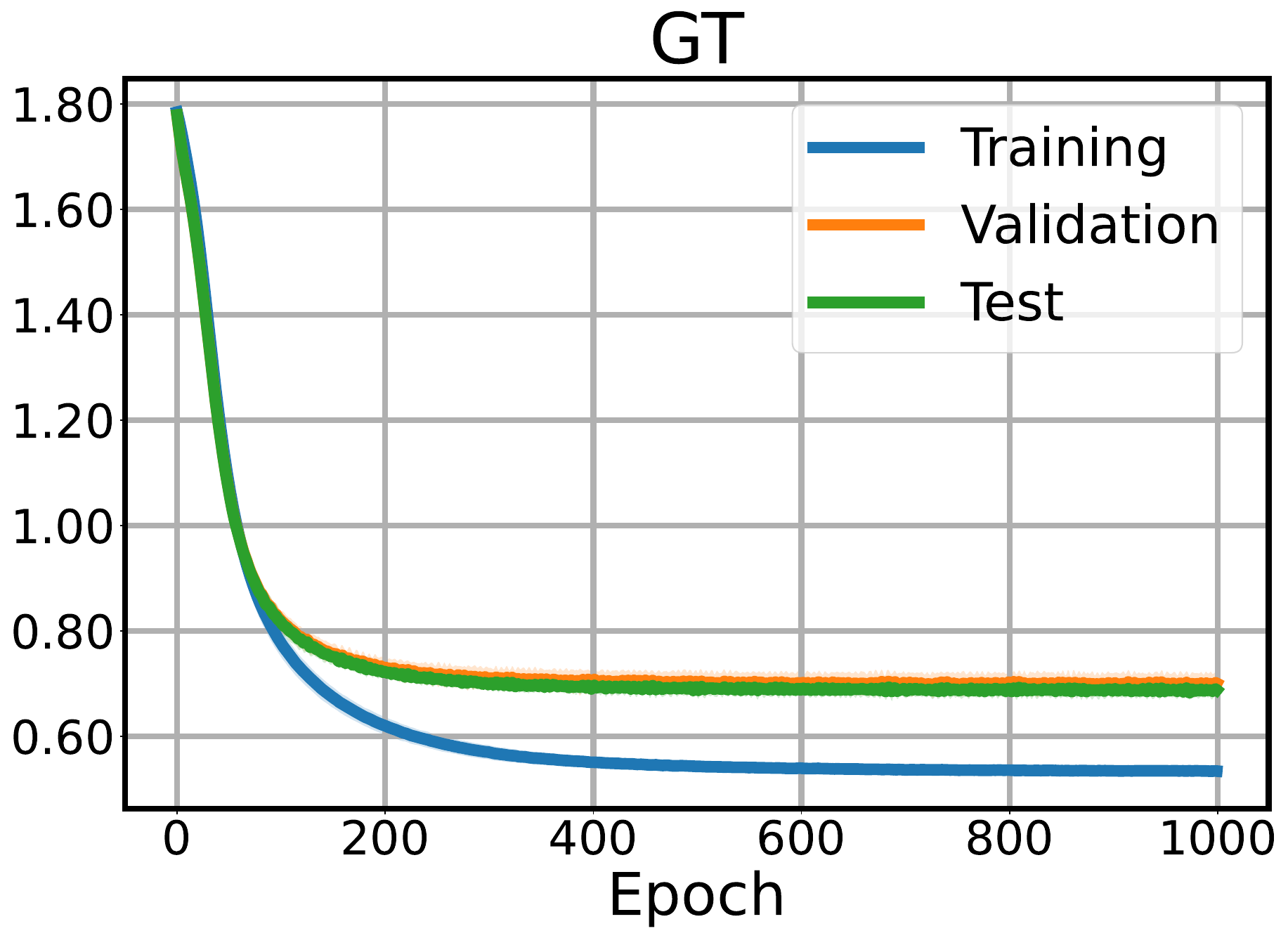}
    \end{subfigure}
    \hspace{-1mm}
    \begin{subfigure}{0.28\textwidth}
        \centering
        \includegraphics[width=\linewidth]{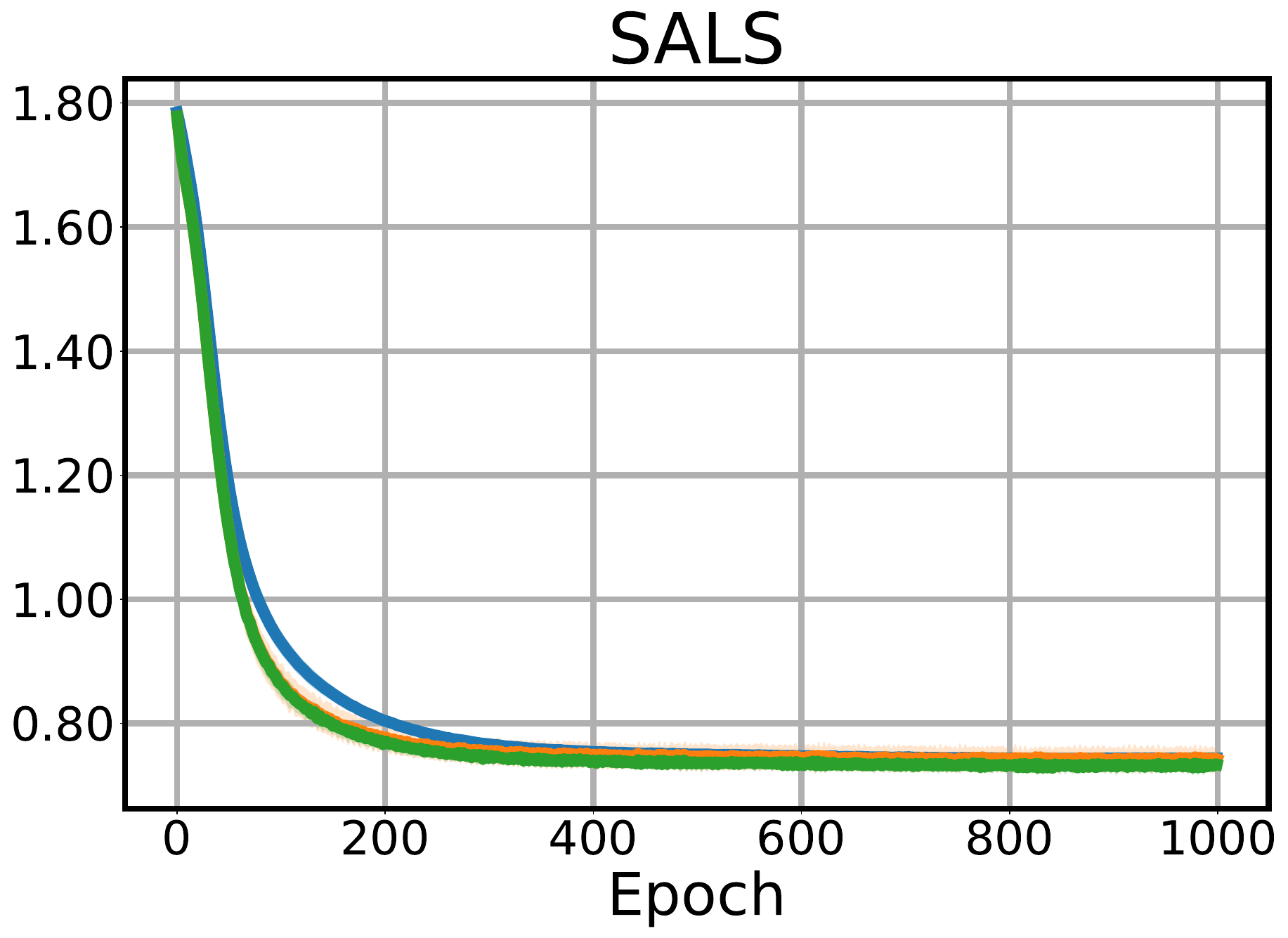}
    \end{subfigure}
    \hspace{-1mm}
    \begin{subfigure}{0.28\textwidth}
        \centering
        \includegraphics[width=\linewidth]{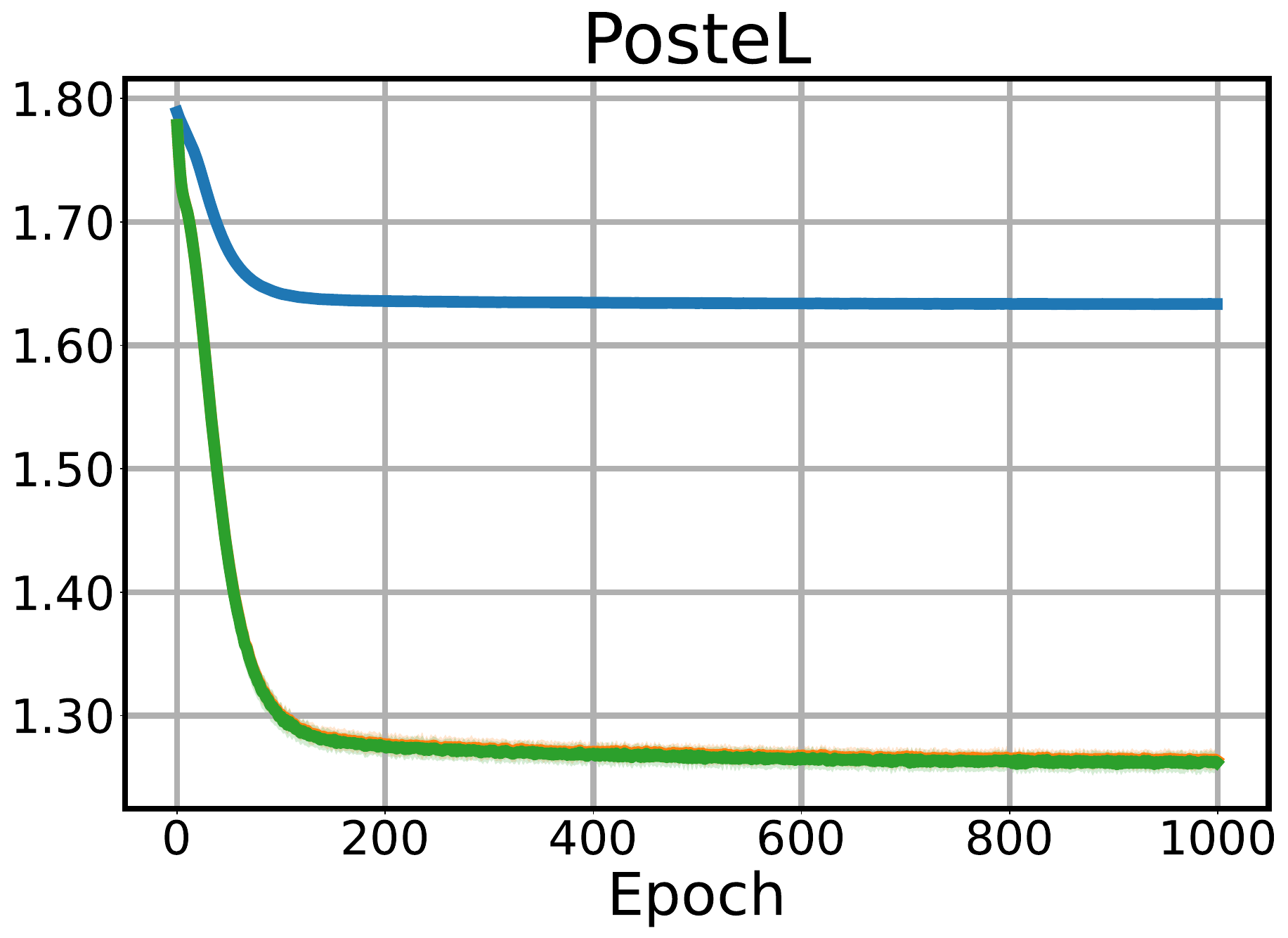}
    \end{subfigure}
    \caption{CiteSeer}
    \end{subfigure}
    \caption{Loss curve of GCN trained on \ours{} labels, SALS labels, and ground truth labels on homophilic datasets.}
    \label{fig:appendix_baselinewise_loss_curve_homo}
\end{figure*}

%% file: figure/figure_appendix_baselinewise_loss_curve_hetero.tex
\begin{figure*}[t!]
    \centering
    \begin{subfigure}{\textwidth}
    \centering
    \begin{subfigure}{0.28\textwidth}
        \centering
        \includegraphics[width=\linewidth]{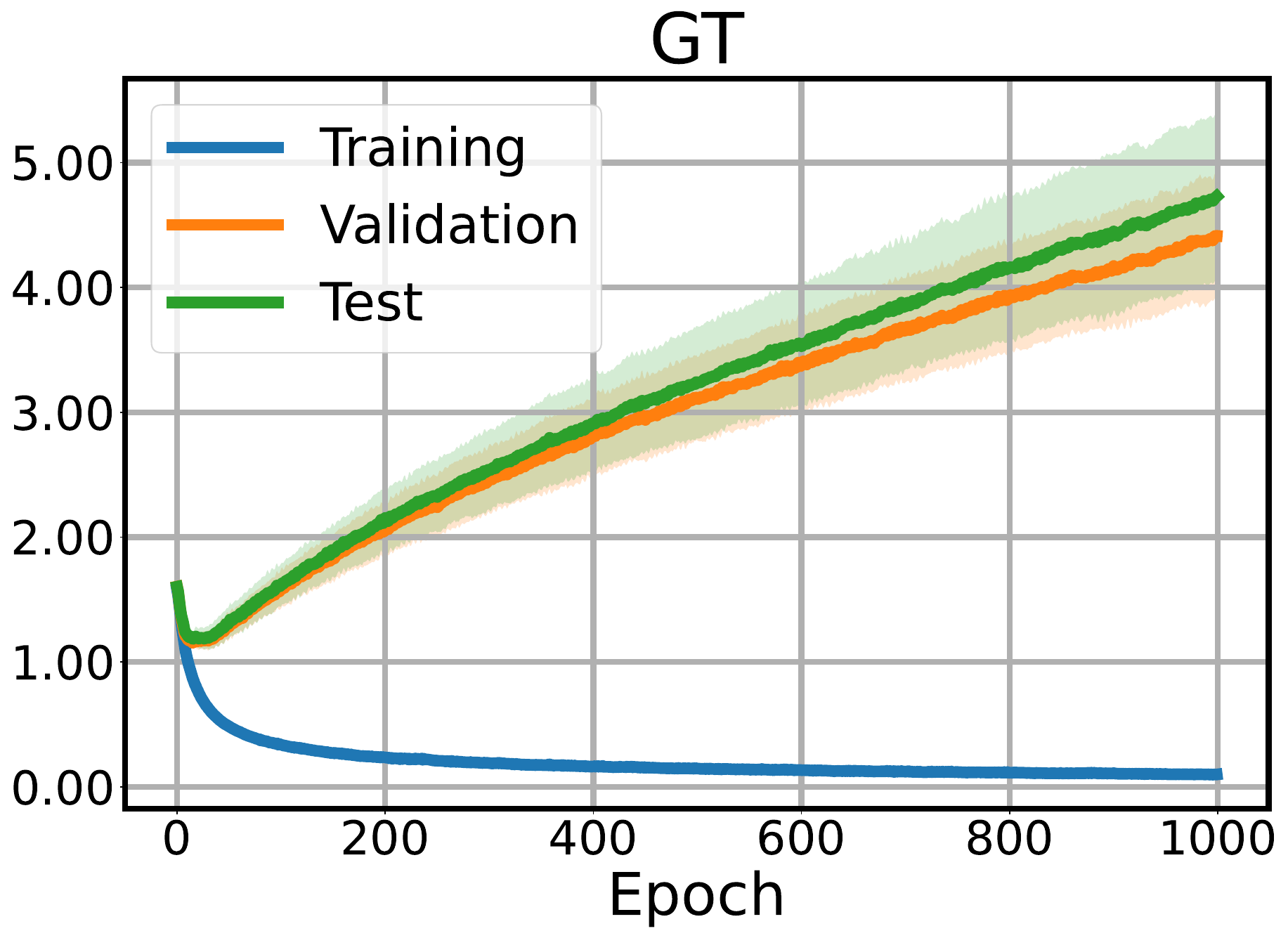}
    \end{subfigure}
    \hspace{-1mm}
    \begin{subfigure}{0.28\textwidth}
        \centering
        \includegraphics[width=\linewidth]{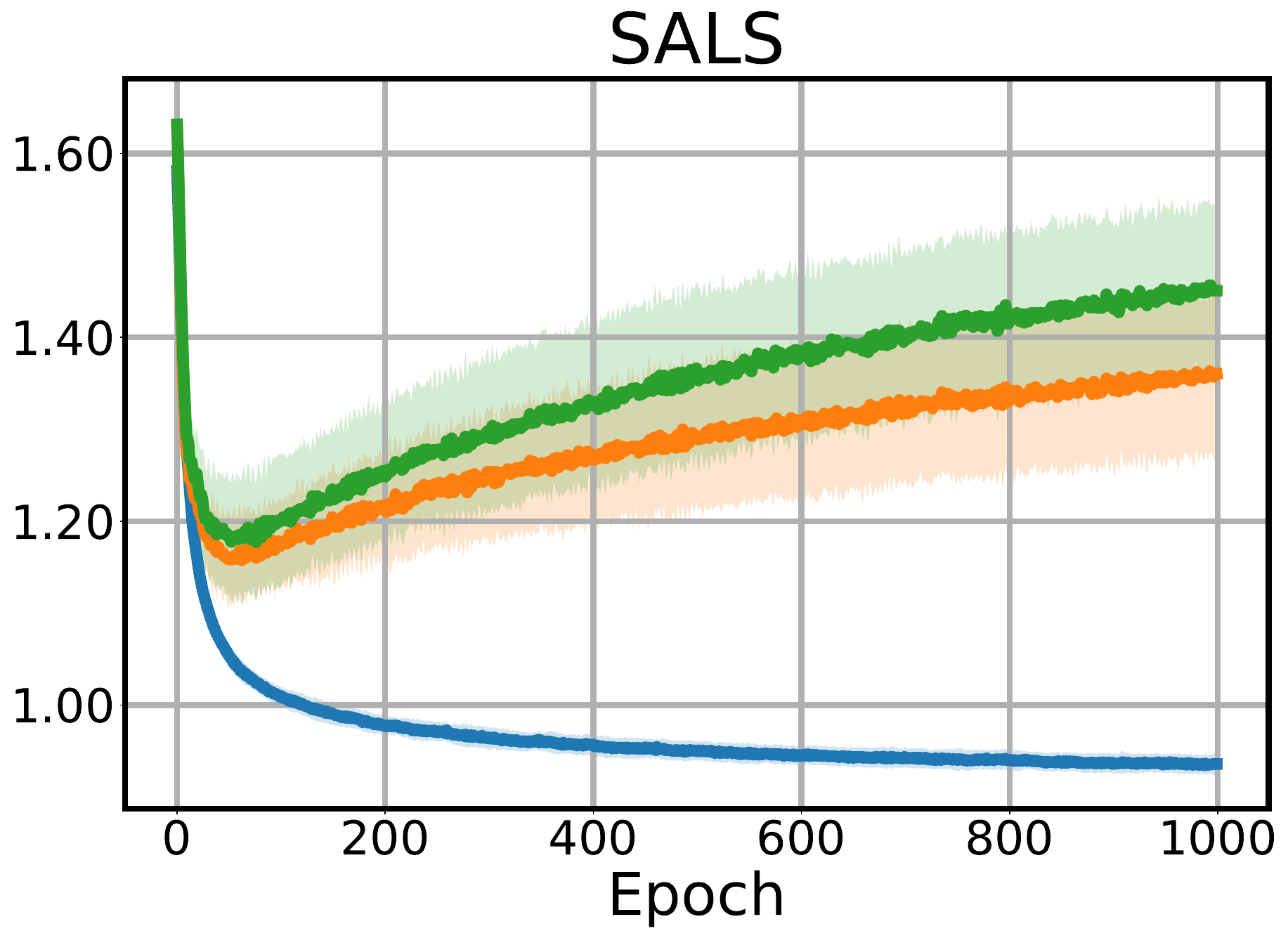}
    \end{subfigure}
    \hspace{-1mm}
    \begin{subfigure}{0.28\textwidth}
        \centering
        \includegraphics[width=\linewidth]{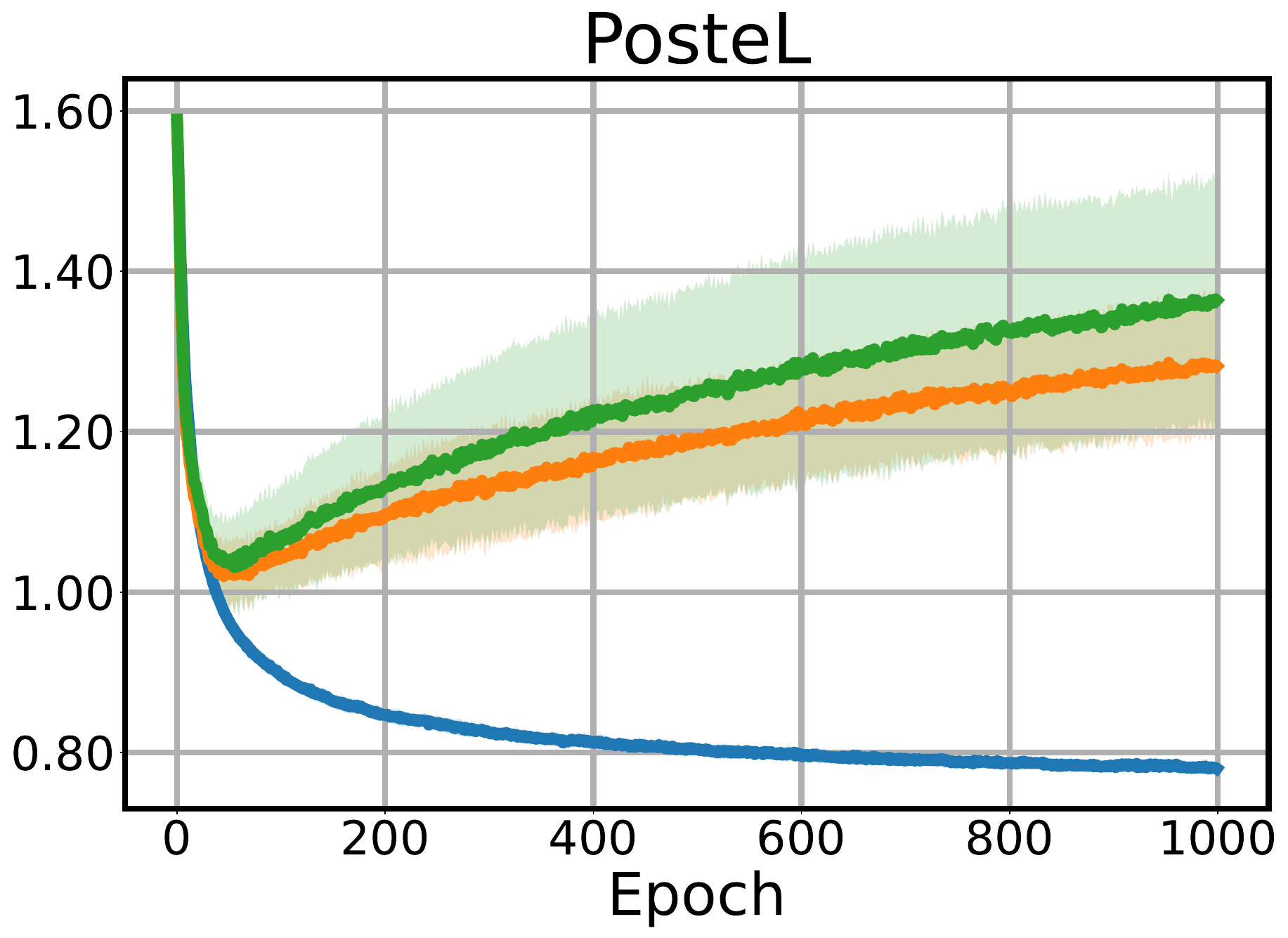}
    \end{subfigure}
    \caption{Chameleon}
    \end{subfigure}

    \begin{subfigure}{\textwidth}
    \centering
    \begin{subfigure}{0.28\textwidth}
        \centering
        \includegraphics[width=\linewidth]{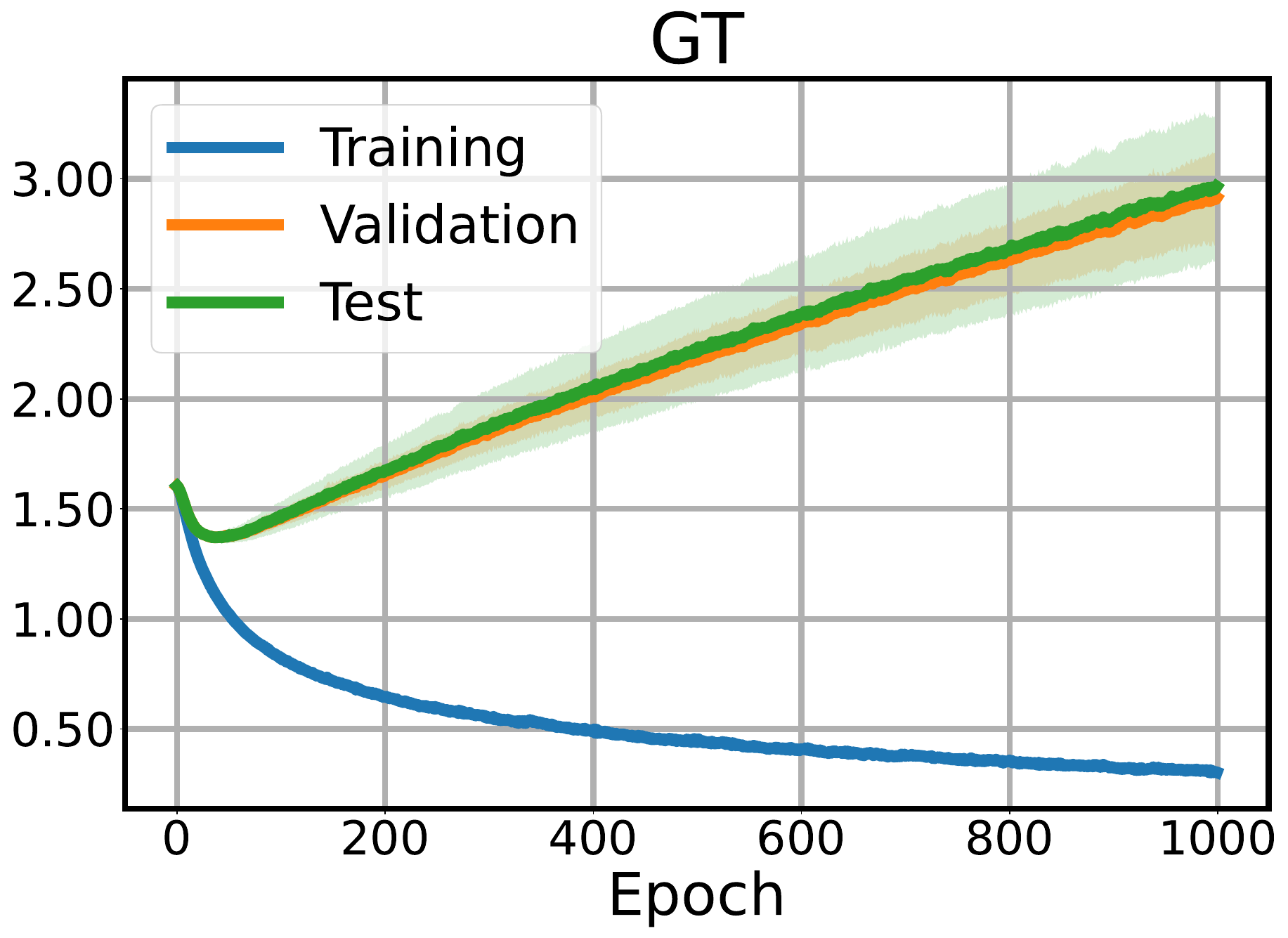}
    \end{subfigure}
    \hspace{-1mm}
    \begin{subfigure}{0.28\textwidth}
        \centering
        \includegraphics[width=\linewidth]{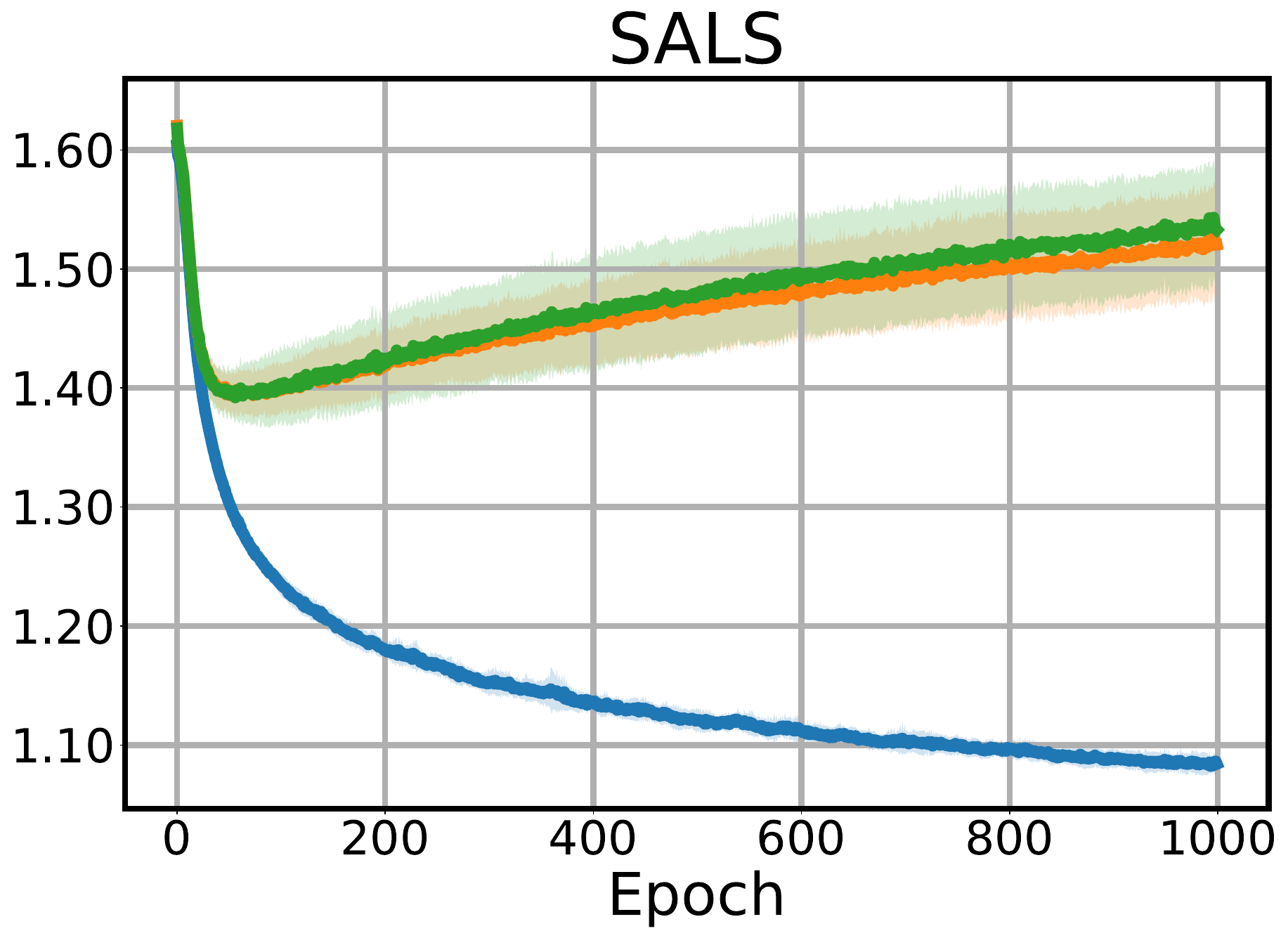}
    \end{subfigure}
    \hspace{-1mm}
    \begin{subfigure}{0.28\textwidth}
        \centering
        \includegraphics[width=\linewidth]{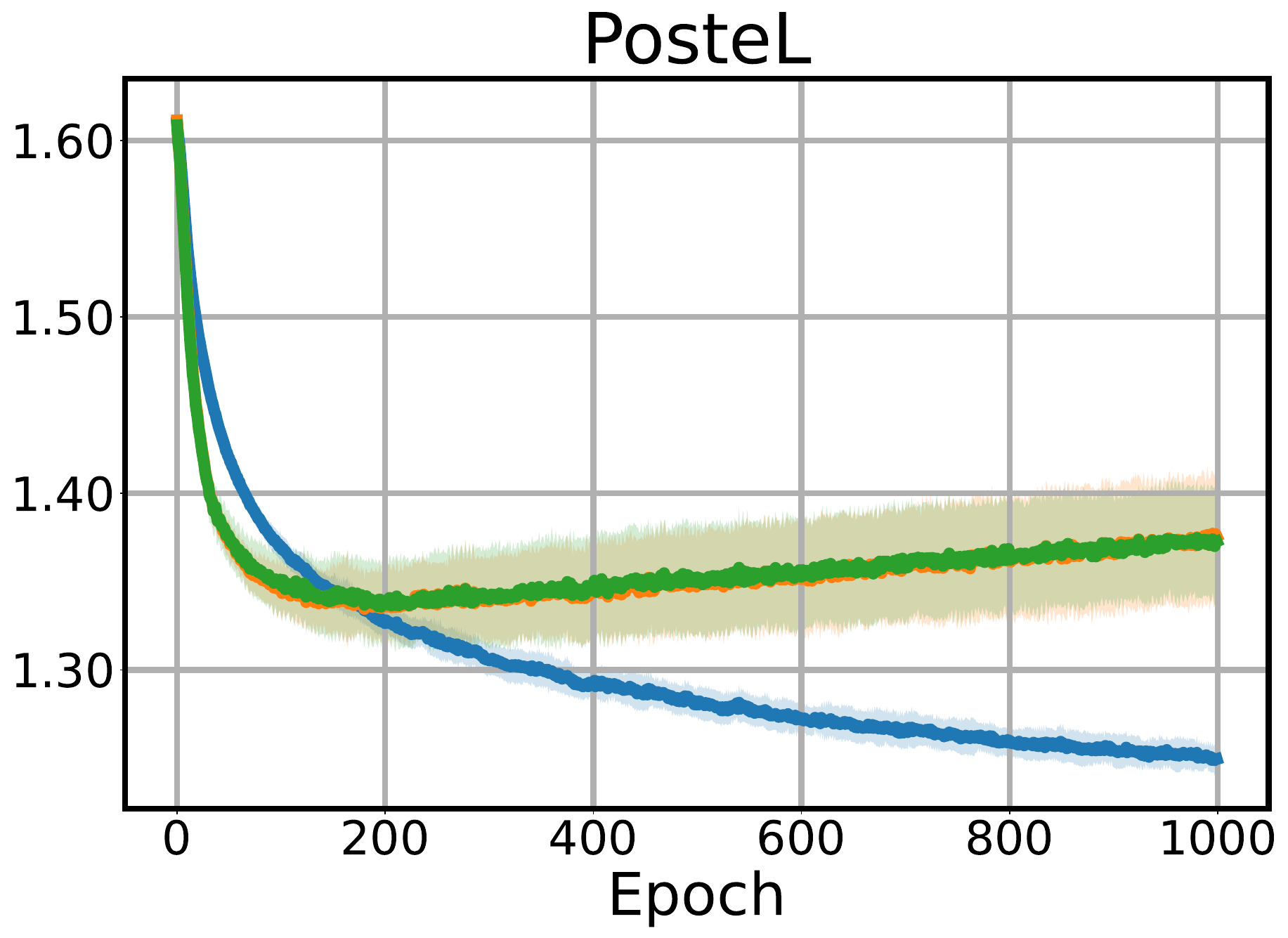}
    \end{subfigure}
    \caption{Squirrel}
    \end{subfigure}

    \begin{subfigure}{\textwidth}
    \centering
    \begin{subfigure}{0.28\textwidth}
        \centering
        \includegraphics[width=\linewidth]{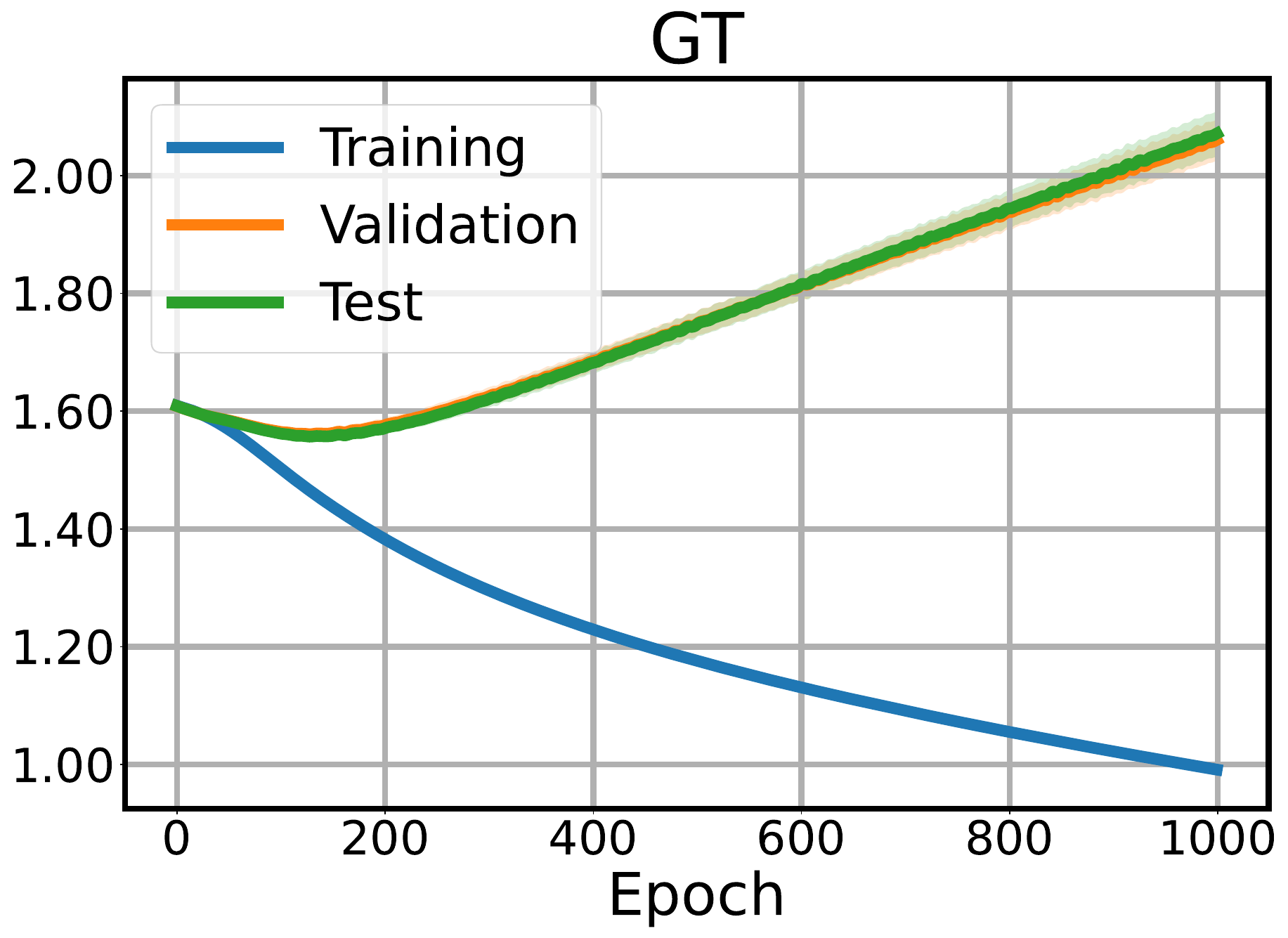}
    \end{subfigure}
    \hspace{-1mm}
    \begin{subfigure}{0.28\textwidth}
        \centering
        \includegraphics[width=\linewidth]{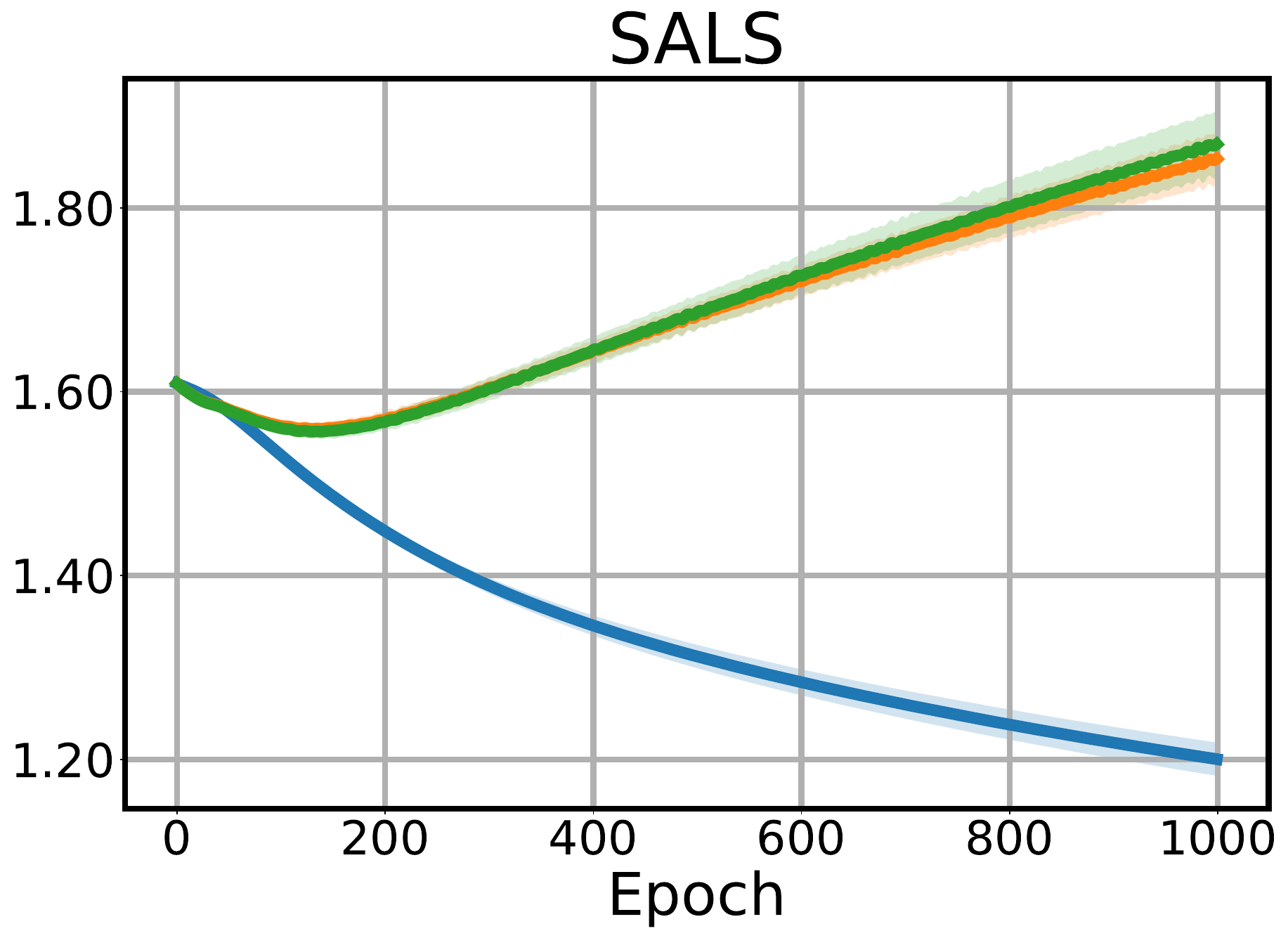}
    \end{subfigure}
    \hspace{-1mm}
    \begin{subfigure}{0.28\textwidth}
        \centering
        \includegraphics[width=\linewidth]{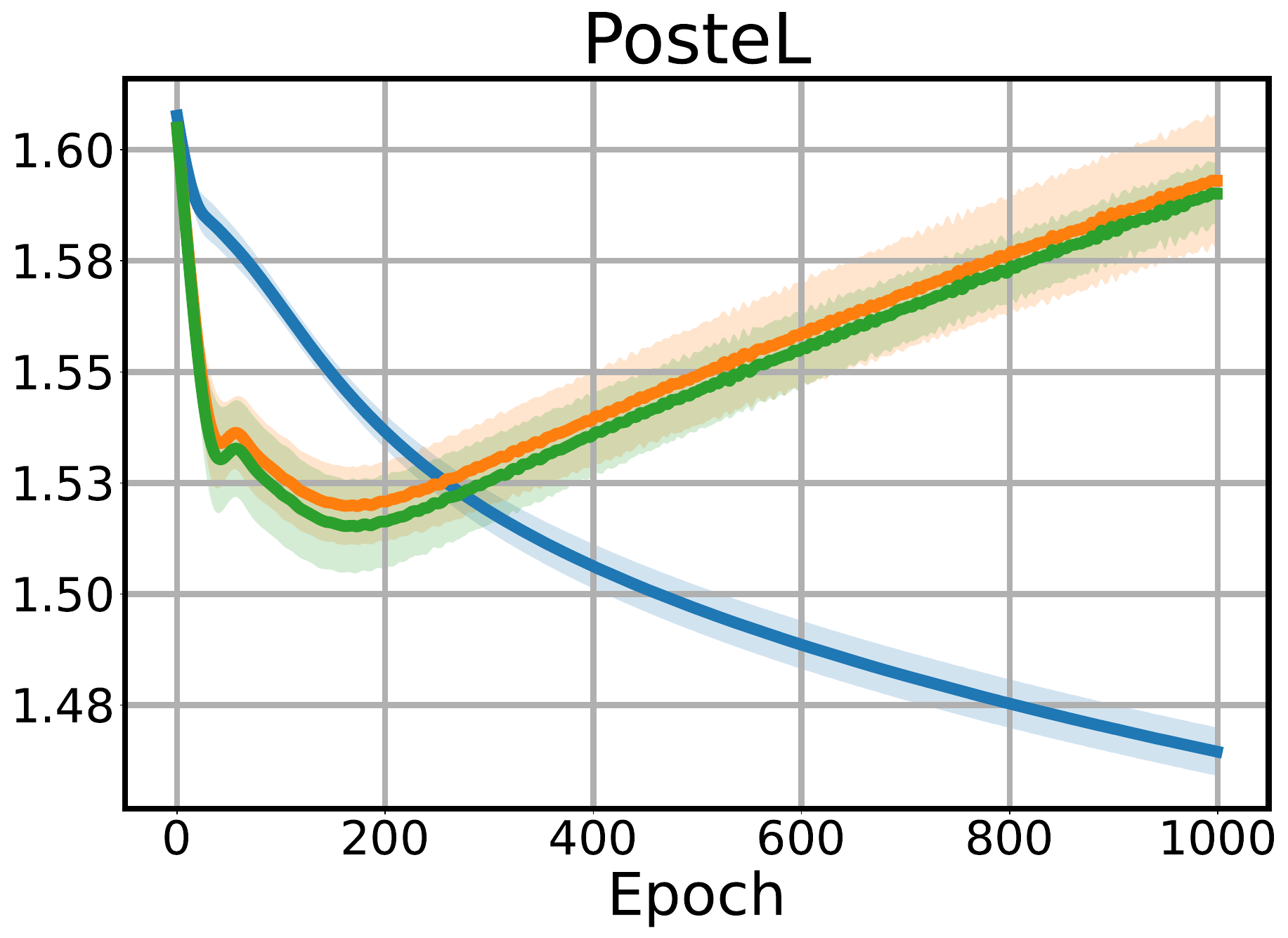}
    \end{subfigure}
    \caption{Actor}
    \end{subfigure}

    \begin{subfigure}{\textwidth}
    \centering
    \begin{subfigure}{0.28\textwidth}
        \centering
        \includegraphics[width=\linewidth]{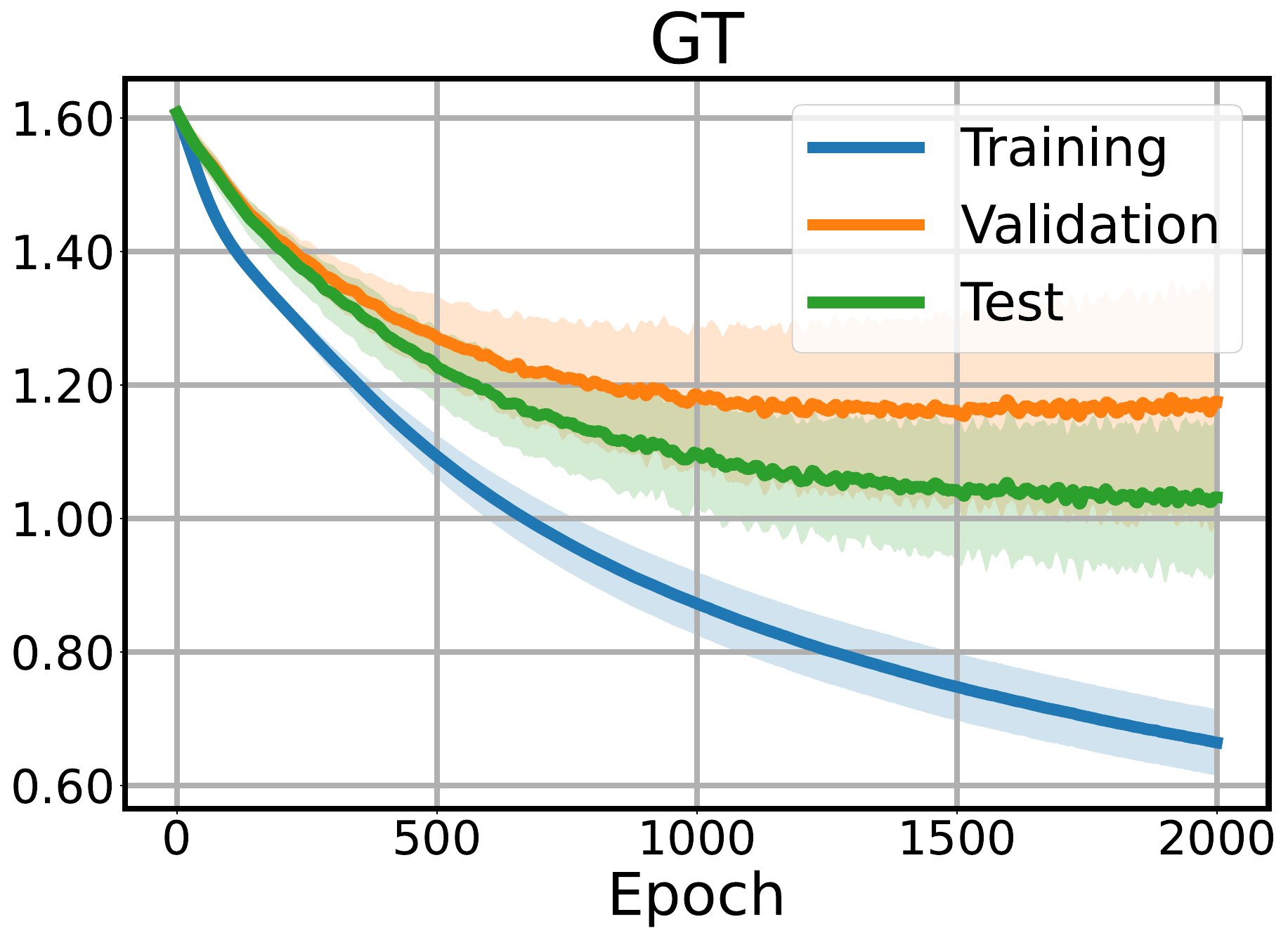}
    \end{subfigure}
    \hspace{-1mm}
    \begin{subfigure}{0.28\textwidth}
        \centering
        \includegraphics[width=\linewidth]{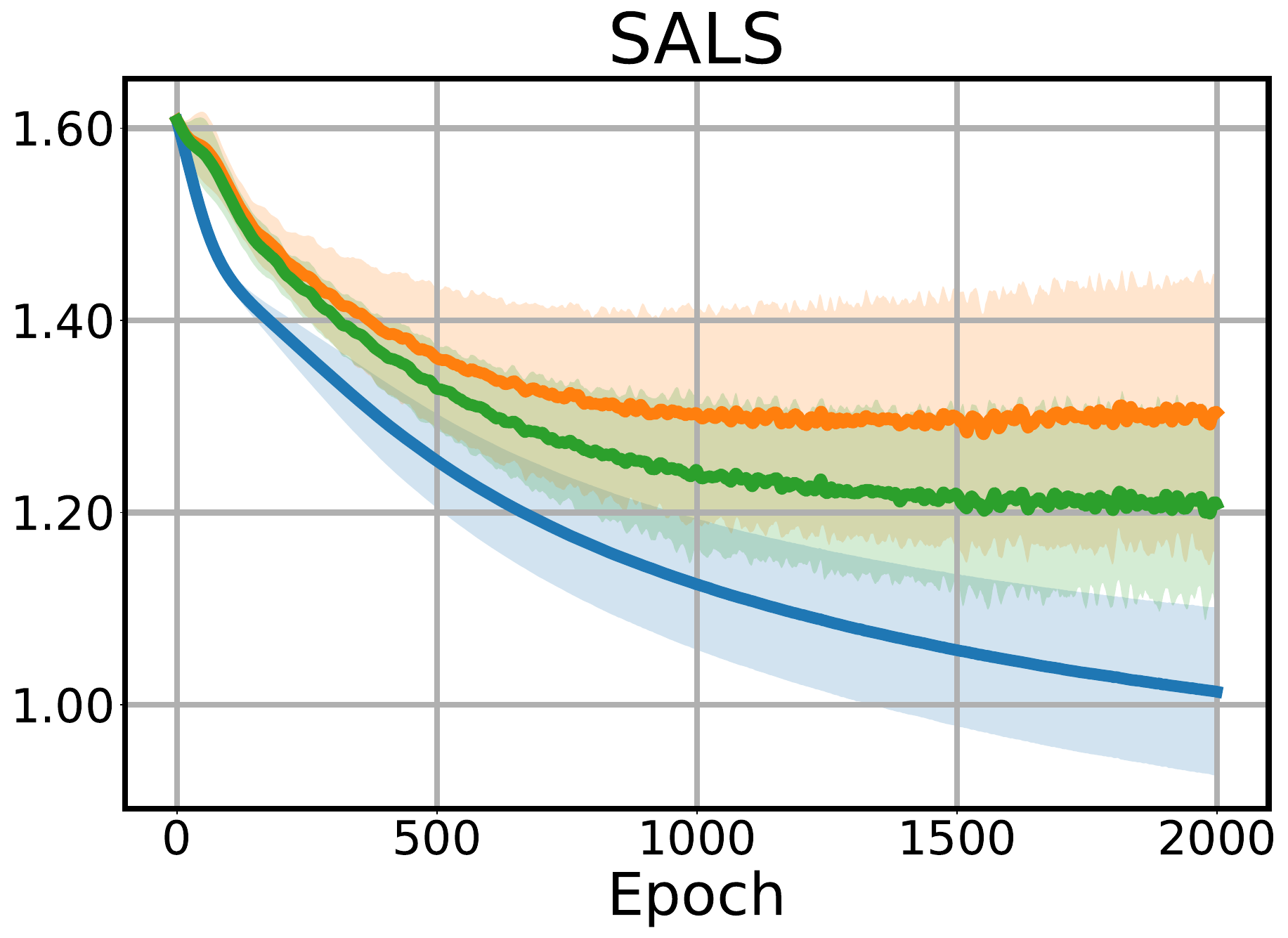}
    \end{subfigure}
    \hspace{-1mm}
    \begin{subfigure}{0.28\textwidth}
        \centering
        \includegraphics[width=\linewidth]{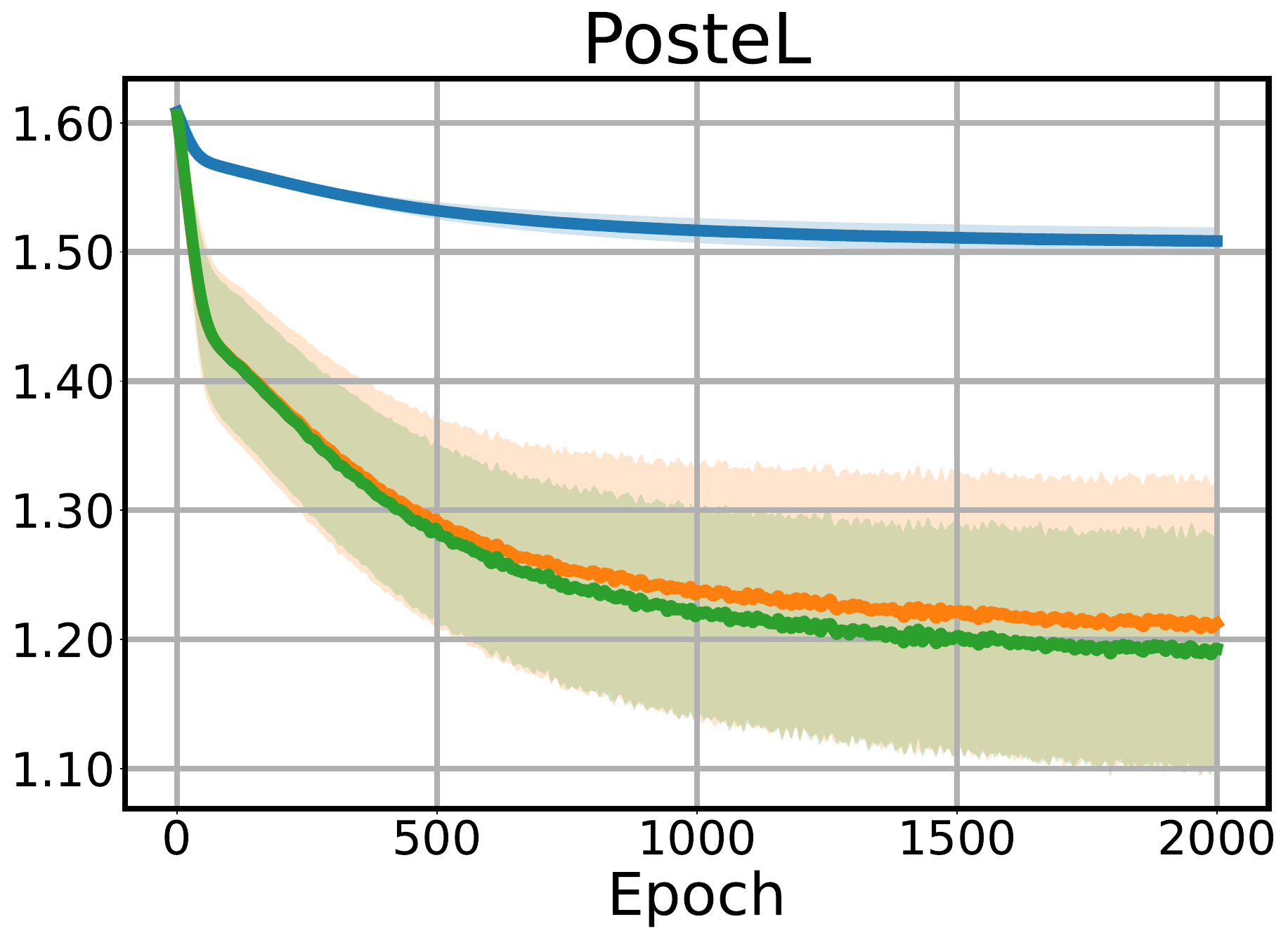}
    \end{subfigure}
    \caption{Texas}
    \end{subfigure}

    \begin{subfigure}{\textwidth}
    \centering
    \begin{subfigure}{0.28\textwidth}
        \centering
        \includegraphics[width=\linewidth]{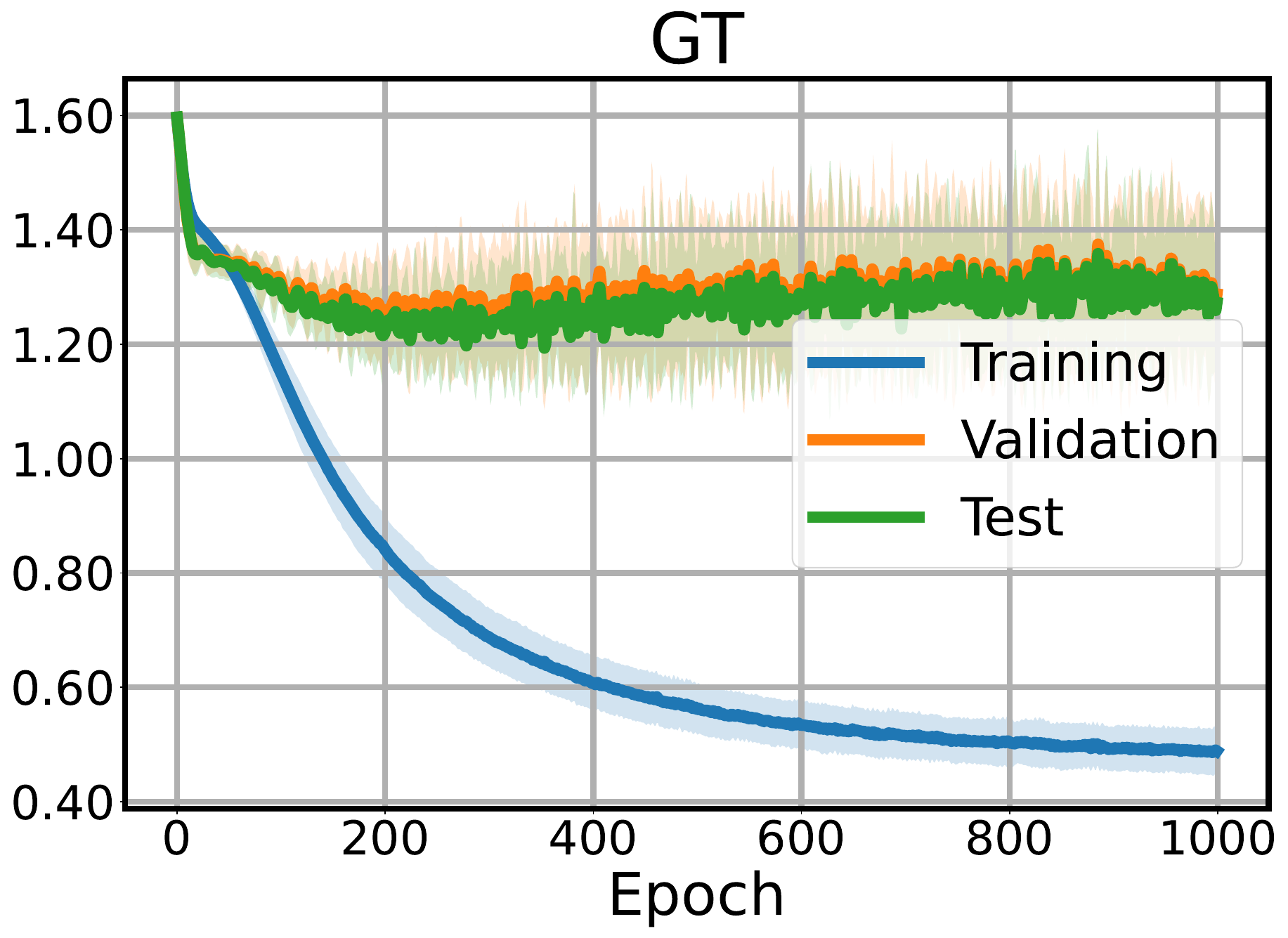}
    \end{subfigure}
    \hspace{-1mm}
    \begin{subfigure}{0.28\textwidth}
        \centering
        \includegraphics[width=\linewidth]{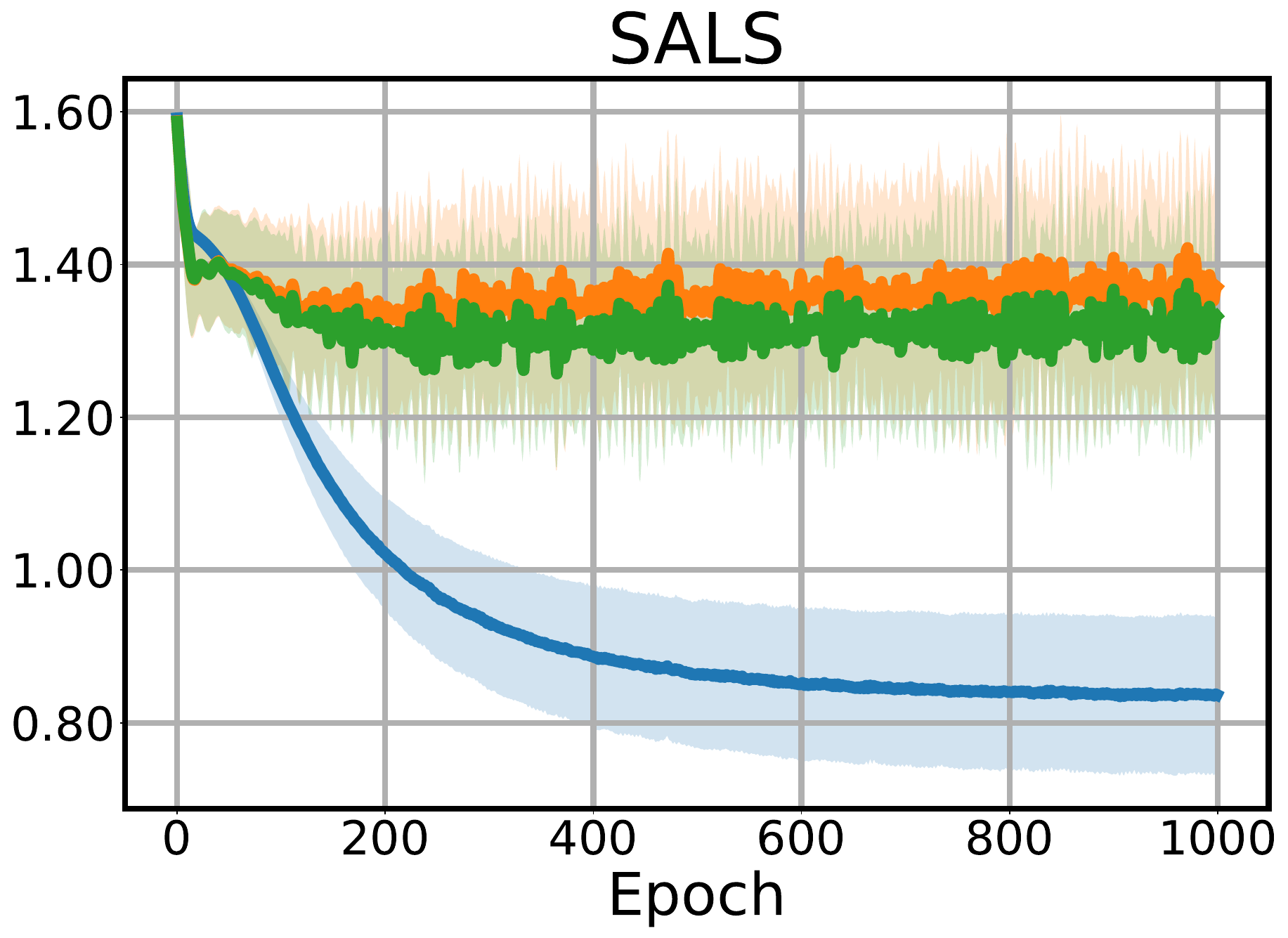}
    \end{subfigure}
    \hspace{-1mm}
    \begin{subfigure}{0.28\textwidth}
        \centering
        \includegraphics[width=\linewidth]{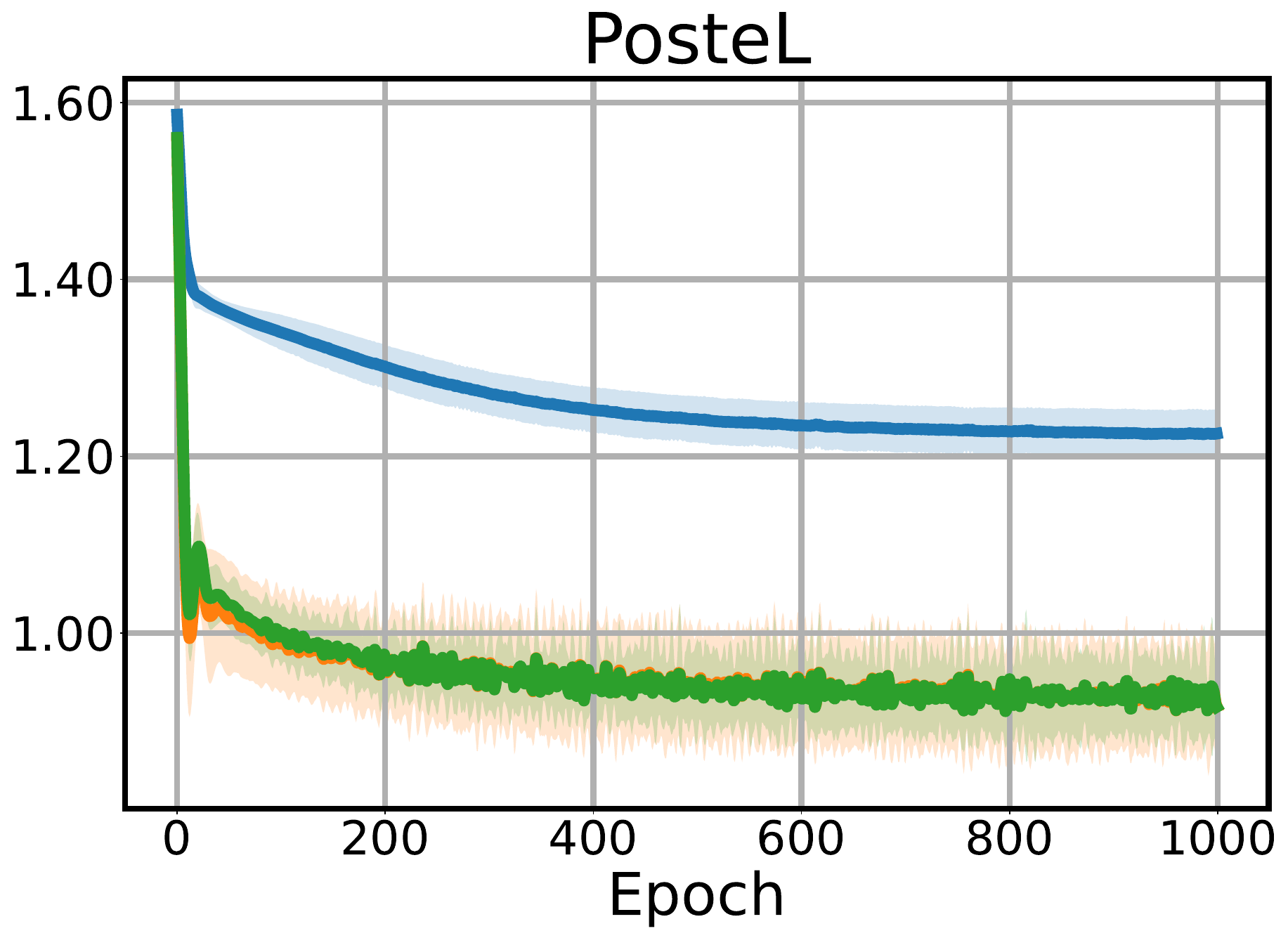}
    \end{subfigure}
    \caption{Cornell}
    \end{subfigure} 
    \caption{Loss curve of GCN trained on \ours{} labels, SALS labels, and ground truth labels on heterophilic datasets.}
    \label{fig:appendix_baselinewise_loss_curve_hetero}
\end{figure*}
    

%% file: figure/figure_conditional_estimation_full.tex
\begin{figure*}[ht]
    \centering
    
    \begin{subfigure}{0.8\textwidth}
        \includegraphics[width=\linewidth]{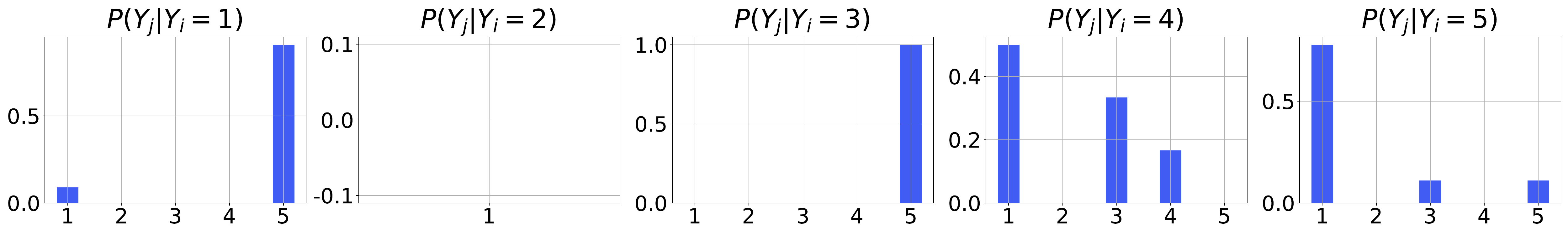}
    \end{subfigure}

    \begin{subfigure}{0.8\textwidth}
        \includegraphics[width=\linewidth]{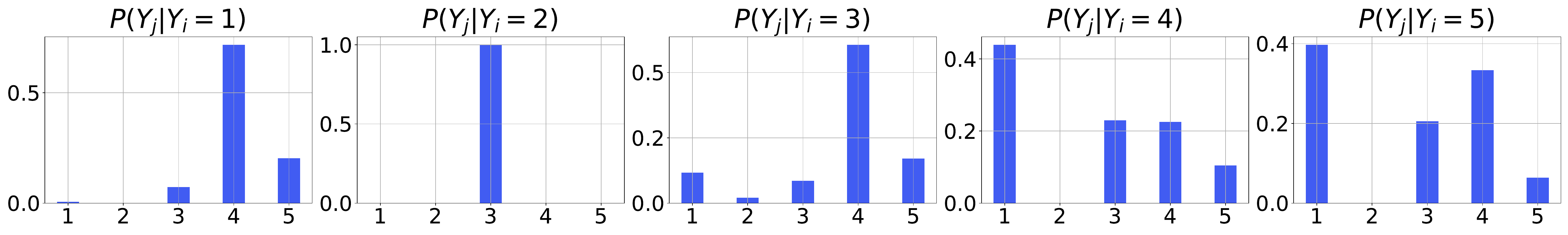}
    \end{subfigure}
    
    \begin{subfigure}{0.8\textwidth}
        \includegraphics[width=\linewidth]{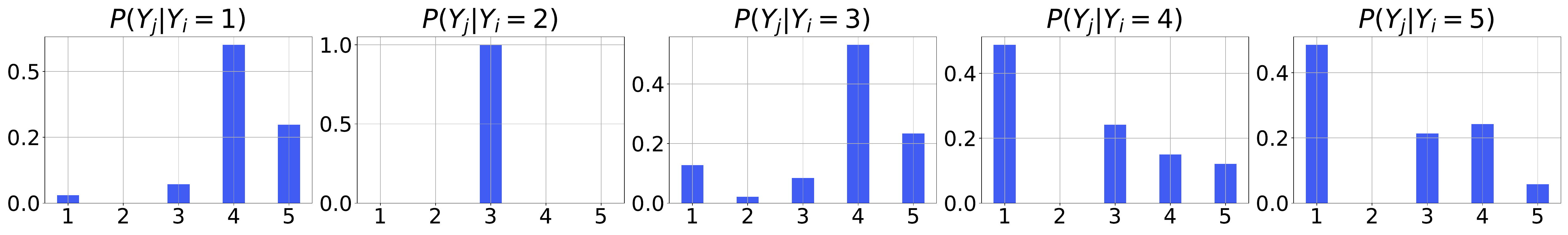}
        \caption{Texas}
    \end{subfigure}

    \begin{subfigure}{0.64\textwidth}
        \includegraphics[width=\linewidth]{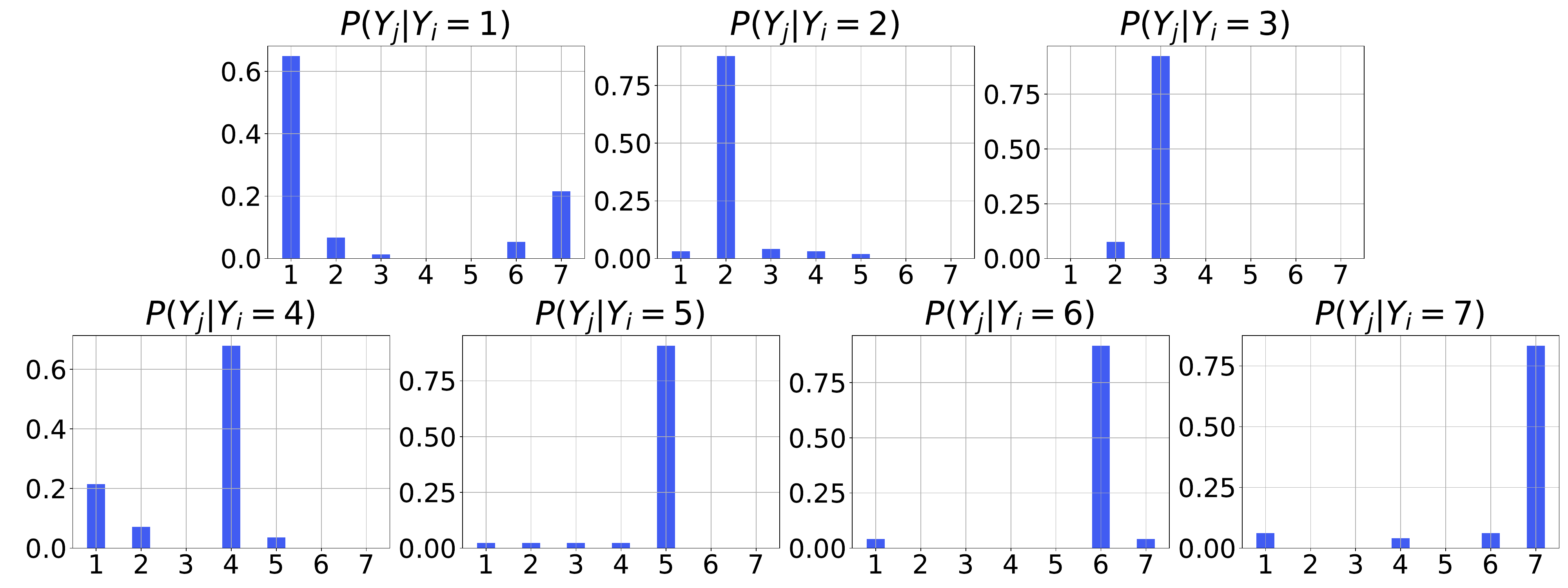}
    \end{subfigure}
    
    \begin{subfigure}{0.64\textwidth}
        \includegraphics[width=\linewidth]{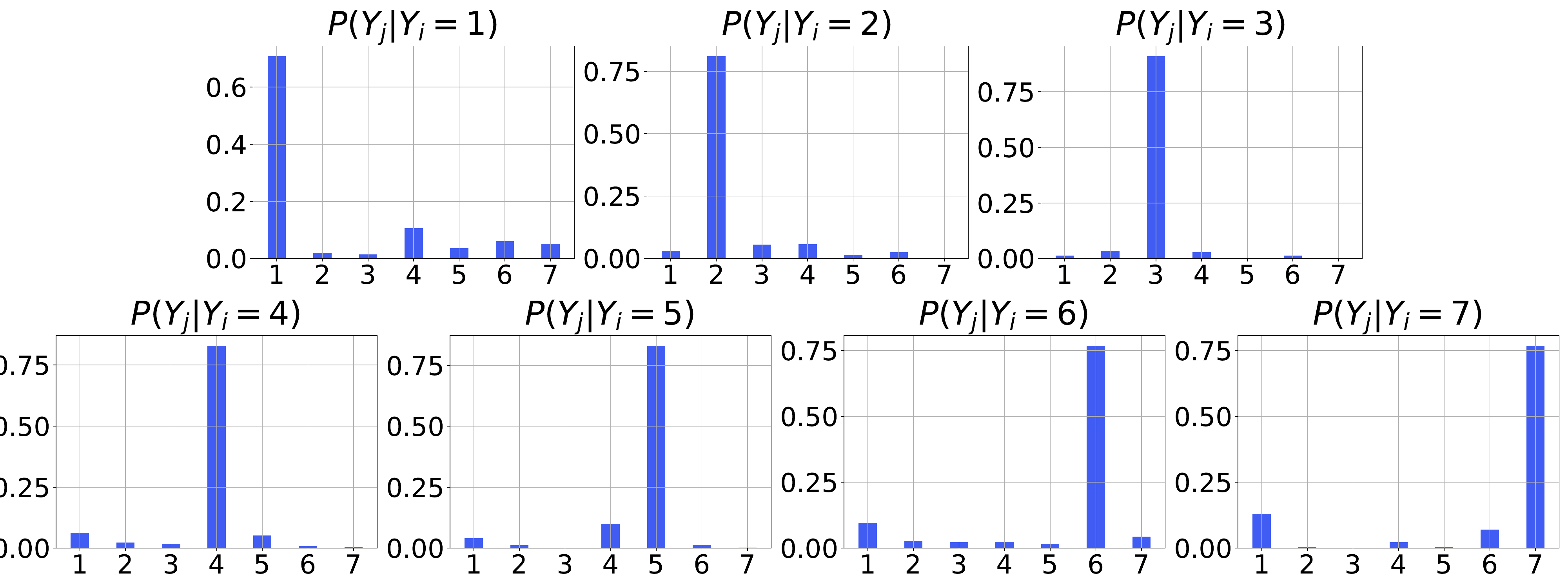}
    \end{subfigure}
    
    \begin{subfigure}{0.64\textwidth}
        \includegraphics[width=\linewidth]{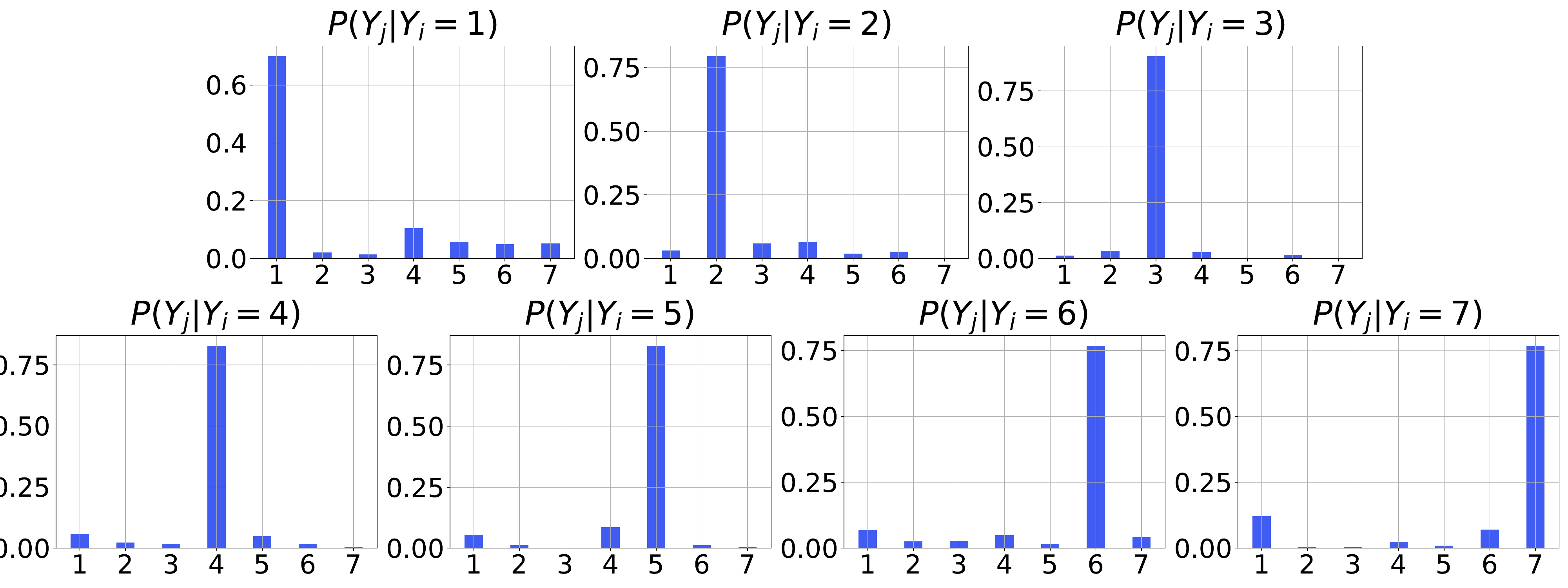}
        \caption{Cora}
    \end{subfigure}
    
    \caption{Estimated conditional distributions based on training labels only (top), training labels with pseudo-labels (middle), and all ground-truth labels (bottom).}
    \label{fig:conditional_estimation_full}
\end{figure*}